 \documentclass[pdflatex,sn-nature]{sn-jnl}

\usepackage{graphicx}%
\usepackage{multirow}%
\usepackage{amsmath,amssymb,amsfonts}%
\usepackage{amsthm}%
\usepackage{mathrsfs}%
\usepackage[title]{appendix}%
\usepackage{xcolor}%
\usepackage{textcomp}%
\usepackage{manyfoot}%
\usepackage{booktabs}%
\usepackage{algorithm}%
\usepackage{algorithmicx}%
\usepackage{algpseudocode}%
\usepackage{listings}%
\usepackage{booktabs}%
\usepackage{graphicx}
\usepackage{xr}

\theoremstyle{thmstyleone}%
\theoremstyle{thmstyletwo}%

\theoremstyle{thmstylethree}%

\begin{document}

\title[Article Title]{Timesynth: A Temporal Fidelity Framework for Health Signal Digital Twins}


\author[1,2]{\fnm{Md Rakibul} \sur{Haque}}\email{rakibul.haque@utah.edu} 
\author[1,2]{\fnm{Shireen} \sur{Elhabian}}\email{shireen@sci.utah.edu} 
\author*[3]{\fnm{Warren Woodrich} \sur{Pettine}}\email{warren.pettine@hsc.utah.edu} 
\affil[1]{\orgdiv{Scientific Computing and Imaging Institute}, \orgname{University of Utah}, \orgaddress{\city{Salt Lake City}, \postcode{84112}, \state{UT}, \country{USA}}}
\affil[2]{\orgdiv{Kahlert School of Computing}, \orgname{University of Utah}, \orgaddress{\city{Salt Lake City}, \postcode{84112}, \state{UT}, \country{USA}}}
\affil*[3]{\orgdiv{Department of Psychiatry}, \orgname{University of Utah}, \orgaddress{\city{Salt Lake City}, \state{UT}, \country{USA}}}



\abstract{
Forecasting models for health-signal digital twins must preserve the oscillatory, frequency, phase, and state-transition dynamics of physiological signals, yet the pointwise metrics used to benchmark them cannot detect when these fundamental properties are lost. We show that this blind spot misranks models: across 11 architectures, models with comparable pointwise error diverge by up to $53^\circ$ in phase accuracy, equivalent to roughly 123~ms for a 1.2~Hz cardiac rhythm and invisible to standard metrics. To enable development of models that escape such failures, we introduce \textsc{TimeSynth}, a controlled benchmarking framework with two reusable components: a physiologically grounded generator producing signals with analytically known ground-truth dynamics from parametric models fitted to real electroencephalography, electrocardiography and photoplethysmogram signals, along with diagnostics quantifying amplitude, frequency, phase, and state-transition fidelity. Linear and full-sequence attention models systematically lose frequency and phase information despite acceptable amplitude error, whereas architectures with localized temporal structure better preserve dynamical fidelity and adapt to observable state transitions; none, however, reliably preserves stochastic switching. Because the dominant determinant of fidelity is architectural, model choice becomes a principled, use-case-driven decision rather than a search for a single winner. \textsc{TimeSynth} thus supplies the controlled preclinical stress test missing before models are coupled to patient data, with a reusable generator and diagnostics for fidelity-aware development.}

\keywords{Digital Twin , Forecasting Models, Physiological Signals, State Change}



\maketitle

\section{Introduction}

Forecasting physiological time series underpins applications from continuous patient monitoring and early warning systems to emerging health-signal digital twins \cite{li2025advancing,sarani2026technologies,sadee2025medical}: patient-specific computational systems, continuously updated with physiological data, that are expected to predict future biological states \cite{nasem2024}. Across these applications, forecasting models must do more than predict future values accurately. Physiological signals encode clinically relevant information in their temporal dynamics, including oscillatory structure, frequency content, phase relationships, and transitions between physiological states that evolve across multiple timescales \cite{clifford2006advanced,tong1990non,sornmo2005bioelectrical}. A model that reproduces future values while distorting these dynamics may appear numerically accurate yet fail to preserve the physiological behavior it is meant to represent.

Yet modern forecasting models are typically evaluated with pointwise metrics such as mean squared error (MSE) and mean absolute error (MAE) \cite{zeng2023transformers,kim2025comprehensive}, which quantify numerical agreement between predicted and observed values but reveal little about whether a forecast preserves oscillatory timing, frequency evolution, or phase coherence \cite{wang2024comprehensive,kim2025comprehensive}. Models with similar forecasting error may therefore represent the underlying process very differently: one may reproduce amplitude while drifting in phase, another may preserve dominant frequencies while smoothing clinically meaningful structure. This is compounded because the dynamical properties of real recordings are generally unknown; without ground-truth amplitude, frequency, and phase trajectories, errors from amplitude distortion, frequency drift, and loss of temporal coherence cannot be disentangled \cite{zeng2023transformers,wang2024comprehensive}. Together, these limitations create a fundamental gap between forecasting accuracy and physiological fidelity.

This gap is particularly consequential for health-signal digital twins. The National Academies of Sciences, Engineering, and Medicine (NASEM) identify verification, validation, and uncertainty quantification (VVUQ) as essential for trust in digital twins \cite{nasem2024}, and recent reviews note the lack of frameworks for assessing whether predictions preserve physiologically meaningful behavior rather than merely acceptable aggregate accuracy \cite{tudor2025scoping,sel2025vvuq}. Clinical data alone cannot settle this: recordings are noisy, nonstationary, and shaped by many interacting processes \cite{goldberger2000physiobank}, and events such as arrhythmia onset, seizure evolution, and sleep-stage transitions cannot be systematically controlled, repeated, or repositioned \cite{obermeyer2016predicting}. Real recordings therefore offer limited opportunity to isolate individual dynamical properties from the complexity of physiological variability.

A complementary approach is to evaluate architectures in controlled environments where the relevant dynamical properties are known, a strategy with precedent across biomedical machine learning. Physiologically grounded synthetic generators have been adopted as enabling infrastructure when richly annotated real data are scarce, exemplified by a recent myoelectric digital twin that simulates large, perfectly annotated electromyography datasets to train neural decoders \cite{maksymenko2023myoelectric}; there, synthetic ground truth is the contribution, not a limitation. In parallel, rigorous benchmarking has repeatedly shown that headline metrics mislead: state-of-the-art cancer drug-response models barely outperform naive baselines once a reproducible, bias-resistant pipeline is imposed \cite{bernett2026dreval}, and across synthetic electronic-health-record generators no single method is best on all criteria, so generators must be assessed per use case \cite{yan2022multifaceted}. Existing time-series benchmarks, however, primarily target predictive accuracy and rarely provide explicit ground-truth dynamics; the field therefore lacks a controlled framework for measuring amplitude, frequency, phase, and state-transition fidelity directly rather than inferring it from aggregate error.

\begin{figure}
\centering
\includegraphics[clip, trim=0.1cm 0.5cm 0.1cm 2.5cm, width=0.9\textwidth]{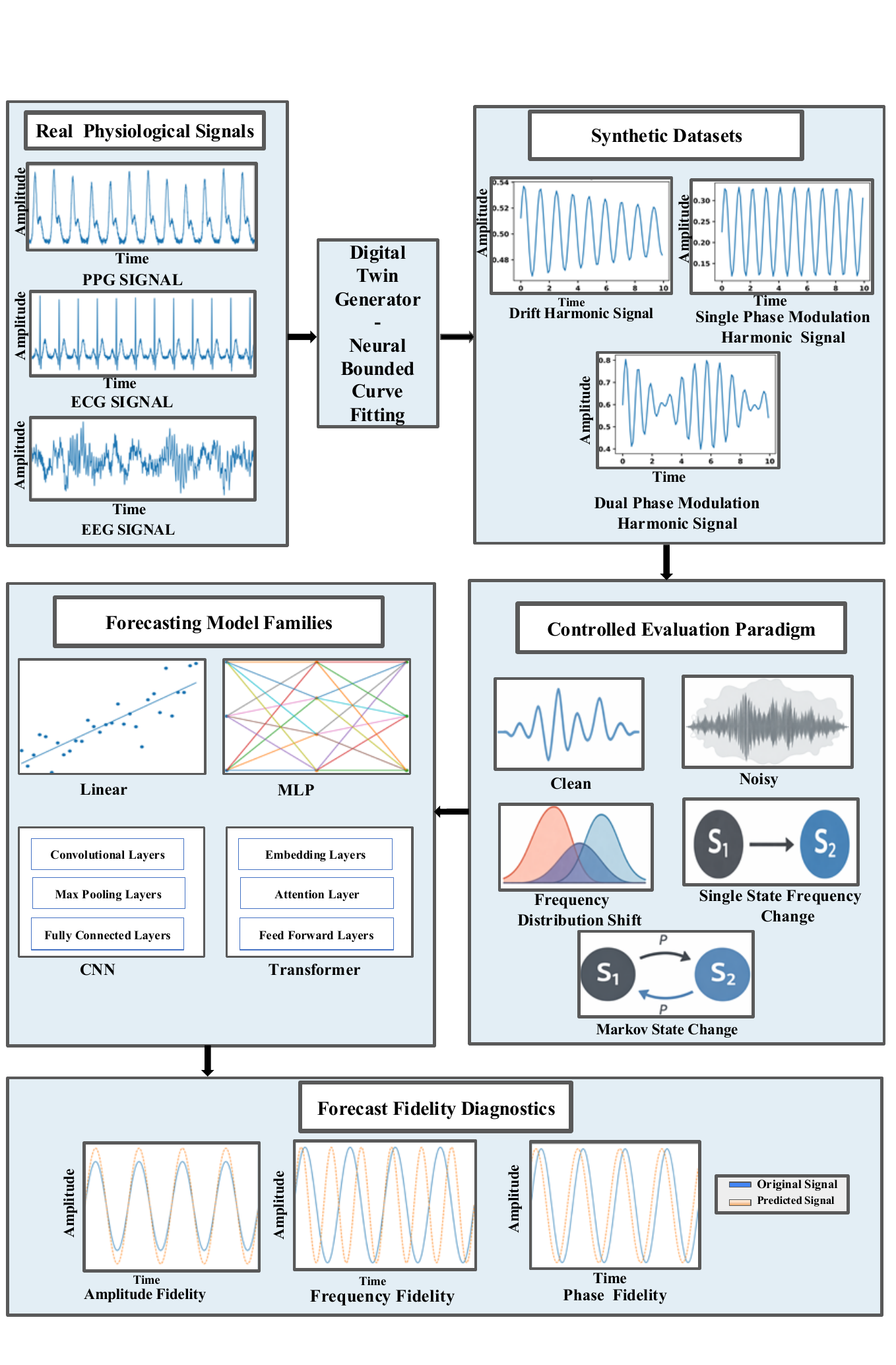}
\caption{\textbf{TimeSynth enables controlled evaluation of dynamical fidelity in physiological forecasting.} Real physiological recordings are used to fit parametric signal models that generate synthetic time series with analytically known dynamical properties. The resulting signal families provide controlled evaluation environments spanning clean forecasting, noise corruption, frequency distribution shifts, deterministic state transitions, and stochastic switching. Forecasting architectures from multiple model families are evaluated using fidelity diagnostics that separately quantify amplitude, frequency, phase, and state-transition preservation. Together, the signal generator and fidelity diagnostics enable systematic assessment of dynamical fidelity beyond conventional forecasting accuracy metrics.}
\label{fig:framework}
\end{figure}

To address this gap, we introduce \textsc{TimeSynth}, a controlled benchmarking framework for evaluating dynamical fidelity in physiological forecasting (Fig.~\ref{fig:framework}). Synthetic signals are generated from closed-form parametric models fitted to real electrocardiogram (ECG), electroencephalogram (EEG), and photoplethysmography (PPG) recordings, providing analytically known amplitude, instantaneous frequency, and instantaneous phase. The framework applies controlled perturbations, including noise corruption, frequency distribution shifts, deterministic state transitions, and stochastic switching, and evaluates architectures with diagnostics that separately quantify amplitude, frequency, phase, and state-transition fidelity.

Using \textsc{TimeSynth}, we evaluate 11 forecasting architectures across four model families. We show that conventional metrics can substantially misrepresent dynamical fidelity: models with similar prediction error differ by up to $53^\circ$ in phase accuracy. Architectures with localized temporal processing preserve dynamical structure most consistently (PatchTST \cite{nie2022time} and MICN \cite{wang2023micn} retain phase and frequency fidelity under noise and distribution shift, and PatchTST and ModernTCN \cite{luo2024moderntcn} adapt fastest to observable state transitions), whereas linear and full-sequence attention models frequently lose phase and frequency information despite competitive pointwise accuracy. No architecture, however, reliably preserves stochastic switching. Together, these findings reveal systematic limitations of current architectures for health-signal digital twins and establish \textsc{TimeSynth} as reusable infrastructure for fidelity-aware evaluation: the controlled, preclinical stress test the VVUQ gap demands before forecasting models are coupled to real patient data.

\section{Results}

We evaluated 11 forecasting architectures spanning four inductive-bias families, comprising linear (Linear, DLinear, FITS), MLP-based (MLinear, NBeats, FreMLP), convolutional and patch-based (MICN, ModernTCN, PatchTST), and full-sequence attention (Transformer, Autoformer) models, as candidate health-signal digital twins. Each was tested under five controlled stress tests that isolate the dynamical demands of physiological forecasting: clean reconstruction, noise corruption, frequency distribution shift, deterministic state transitions, and stochastic Markov switching. The architectural rationale and per-family hypotheses are detailed in Methods. Unless otherwise noted, pairwise comparisons against the linear baseline use paired $t$-tests with Holm correction; the state-transition analysis uses Wilcoxon signed-rank tests (Methods).

\subsection*{Physiologically grounded synthetic signals provide controlled ground truth}

Real biosignals do not expose ground-truth amplitude, frequency, and phase trajectories, so controlled synthetic signals are required to test whether a forecast preserves these dynamics rather than merely matching future values \cite{clifford2006advanced, sornmo2005bioelectrical}. We therefore fit closed-form parametric models to recordings from three public datasets and sampled new signals from the resulting empirical parameter distributions, yielding signals whose amplitude, instantaneous frequency, and instantaneous phase are analytically known at every time point (Methods; Supplementary Tables~\ref{tab:param_bounds}--\ref{tab:fitting_hyperparams_alt}).

\begin{figure}[!htbp]
    \centering
    \includegraphics[width=0.9\linewidth]{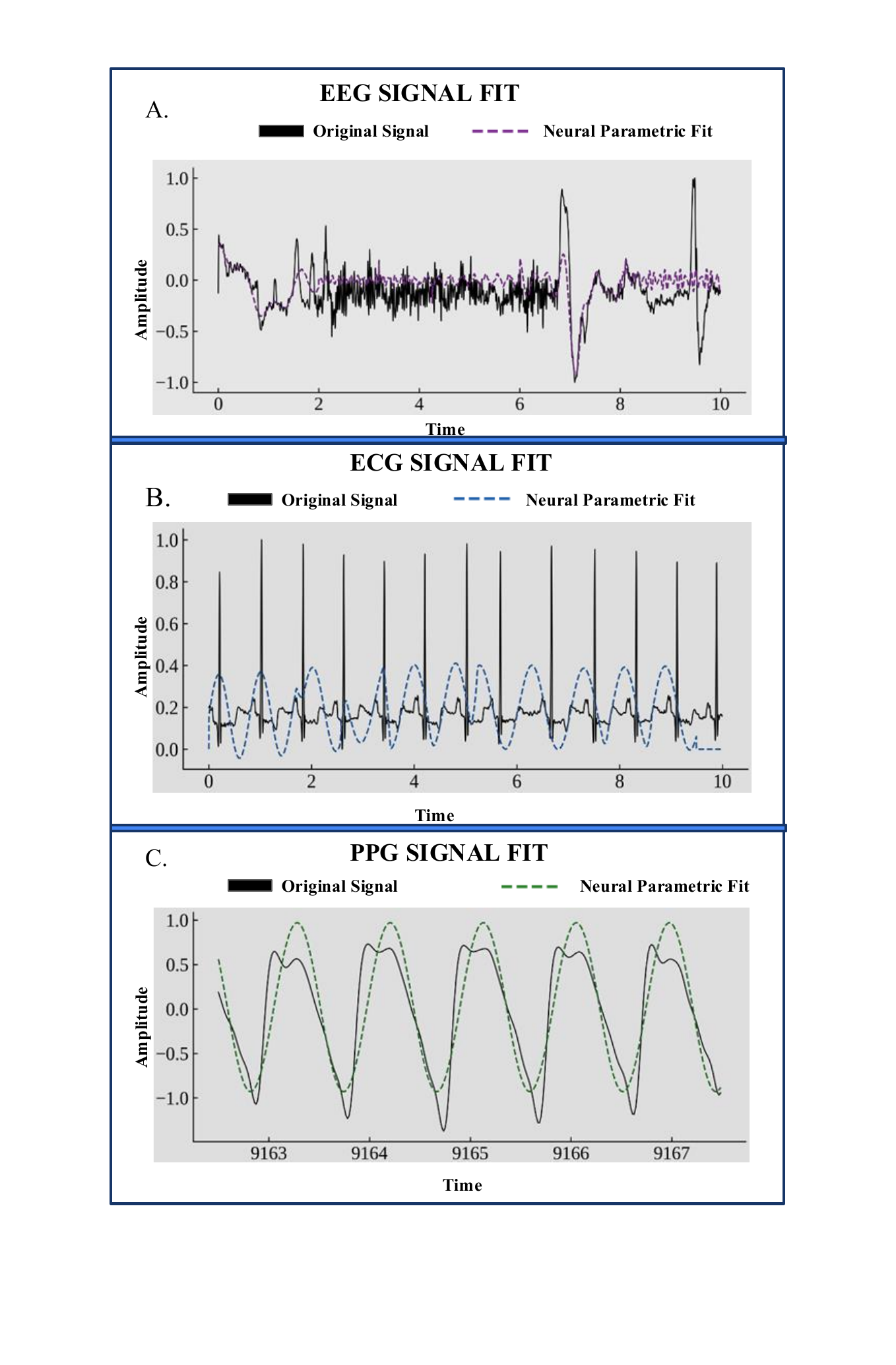}
    \caption{\textbf{Parametric models fitted to real biosignals ground synthetic evaluation in physiological dynamics.} Parametric fits (dashed) overlaid on original recordings (black) for (\textbf{A})~EEG (CHB-MIT, channel FP1-F7, 10\,s segment), (\textbf{B})~ECG (MIT-BIH record 100, 10\,s segment, S-Q interval baseline), and (\textbf{C})~PPG/BVP (PPG-DaLiA subject S1, best-fit 5\,s segment). The fits capture the dominant oscillatory morphology of each signal type while providing closed-form ground truth for amplitude, frequency, and phase.}
    \label{fig:fitting}
\end{figure}

The fitted families reproduced the dominant morphology of each signal type. The drift-harmonic family, derived from blood-volume-pulse segments in PPG-DaLiA \cite{ppg}, captured the quasi-sinusoidal morphology of peripheral cardiovascular signals, with fitted frequencies concentrated near 1.0--1.1~Hz (consistent with resting heart rate) and near-zero drift coefficients (Fig.~\ref{fig:fitting}C). The single and dual phase-modulated families, derived from the MIT-BIH Arrhythmia Database \cite{moody2001impact, goldberger2000physiobank}, captured the smooth oscillatory baseline of the S-Q interval, excluding sharp R-peak transients that violate sinusoidal decomposition, with carrier frequencies clustered near 1.0~Hz across all 48 records and modulation consistent with heart rate variability (Fig.~\ref{fig:fitting}B). The dual phase-modulated family, fit to channel FP1-F7 of the CHB-MIT Scalp EEG Database \cite{shoeb2009application}, captured dominant cortical rhythmic activity while accommodating transient deflections (Fig.~\ref{fig:fitting}A). Table~\ref{tab:table1} maps each synthetic paradigm to the physiological phenomenon it approximates, grounding the preclinical evaluation used throughout.

\begin{table}[!htbp]
\centering
\caption{Physiological events and the synthetic signals used to simulate them in TimeSynth.}
\label{tab:table1}
\begin{tabular}{@{}ll@{}}
\toprule
\textbf{Physiological event} & \textbf{Synthetic signal \& perturbation} \\
\midrule
Baseline wander (ECG, EEG)            & Drift-harmonic (low-frequency drift) \\
Respiration coupling (PPG)            & Single phase modulation \\
Heart rate variability (ECG)          & Dual phase modulation + Markov switching \\
Neural oscillation variability (EEG)  & Multi-frequency modulation + state switching \\
Sensor noise (all signals)            & Additive noise (SNR 0--6\,dB) \\
Exercise-induced rhythm change        & Non-stationary frequency shift \\
Arrhythmia onset (ECG)                & Single deterministic state transition \\
Sleep-stage transition (EEG)          & State transition + Markov switching \\
Electrode artifact (ECG, EEG)         & Drift + additive noise \\
\bottomrule
\end{tabular}
\end{table}

\subsection*{Pointwise accuracy fails to reflect physiological fidelity}

Clinically relevant information in physiological signals is encoded in the timing and rhythm of oscillatory events, not amplitude alone, so a model with low prediction error but mistracked phase or frequency can appear reliable yet fail to preserve the dynamics that motivate a digital twin. Because MSE, MAE, and their normalized variants quantify only pointwise amplitude deviation, we evaluated three complementary fidelity dimensions in parallel: amplitude, phase (timing misalignment), and frequency (dominant-rate mismatch), each defined in Methods.

\begin{figure}[!htbp]
    \centering
    \includegraphics[width=1.0\linewidth]{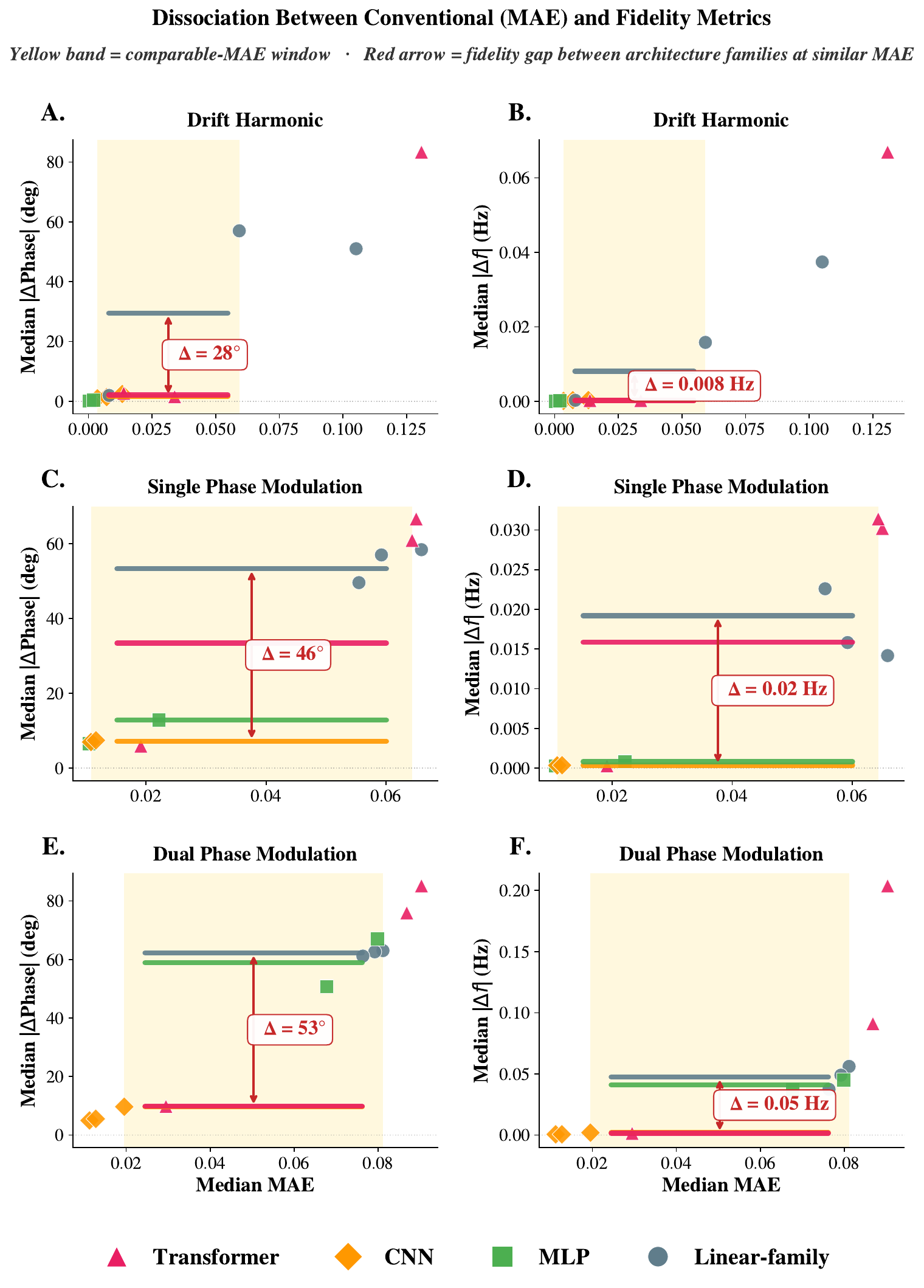}
    \caption{\textbf{Pointwise accuracy ranks models approximately but cannot resolve phase and frequency differences among models with comparable MAE.} Median MAE versus median absolute phase error (left: \textbf{a,c,e}) and median absolute frequency error (right: \textbf{b,d,f}) across drift-harmonic (\textbf{a,b}), single-phase modulation (\textbf{c,d}), and dual-phase modulation (\textbf{e,f}) signals. Points are colored by architecture family. The yellow band marks the comparable-MAE window (central 60\% of MAE values); horizontal colored lines give each family's mean fidelity error within that window and red arrows mark the family-level gap. Error metrics are defined in Methods.}
    \label{fig:dissociation}
\end{figure}

Across all 11 models and three signal families, MAE was strongly rank-correlated with phase error (Spearman $\rho = 0.93$, $0.90$, and $0.97$ on drift-harmonic, single-phase, and dual-phase signals; $p < 0.001$ each), confirming that pointwise error captures coarse ordering, yet models with comparable MAE still diverged sharply in fidelity (Fig.~\ref{fig:dissociation}). Within the comparable-MAE window, linear-family models exhibited phase deviations exceeding $50^{\circ}$, whereas convolutional and patch-based architectures held phase error below $10^{\circ}$ ($p < 0.05$, all pairwise). The dissociation widened with signal complexity, from $\Delta = 28^{\circ}$ on drift-harmonic to $46^{\circ}$ on single-phase and $53^{\circ}$ on dual-phase signals (about 123~ms at a 1.2~Hz cardiac rhythm), with a parallel frequency gap growing from $0.008$ to $0.05$~Hz; outside the comparable-MAE window, full-sequence attention models reached frequency deviations up to $0.20$~Hz. Pointwise metrics therefore rank models only approximately and cannot resolve the dynamical quantities physiological forecasting requires. The remaining analyses evaluate each fidelity dimension separately across the five stress tests.

\subsection*{Localized architectures preserve oscillatory structure on clean signals}

\begin{figure}[!htbp]
    \centering
    \includegraphics[clip, trim=0.3cm 6.5cm 0.3cm 2.5cm, width=1.0\textwidth]{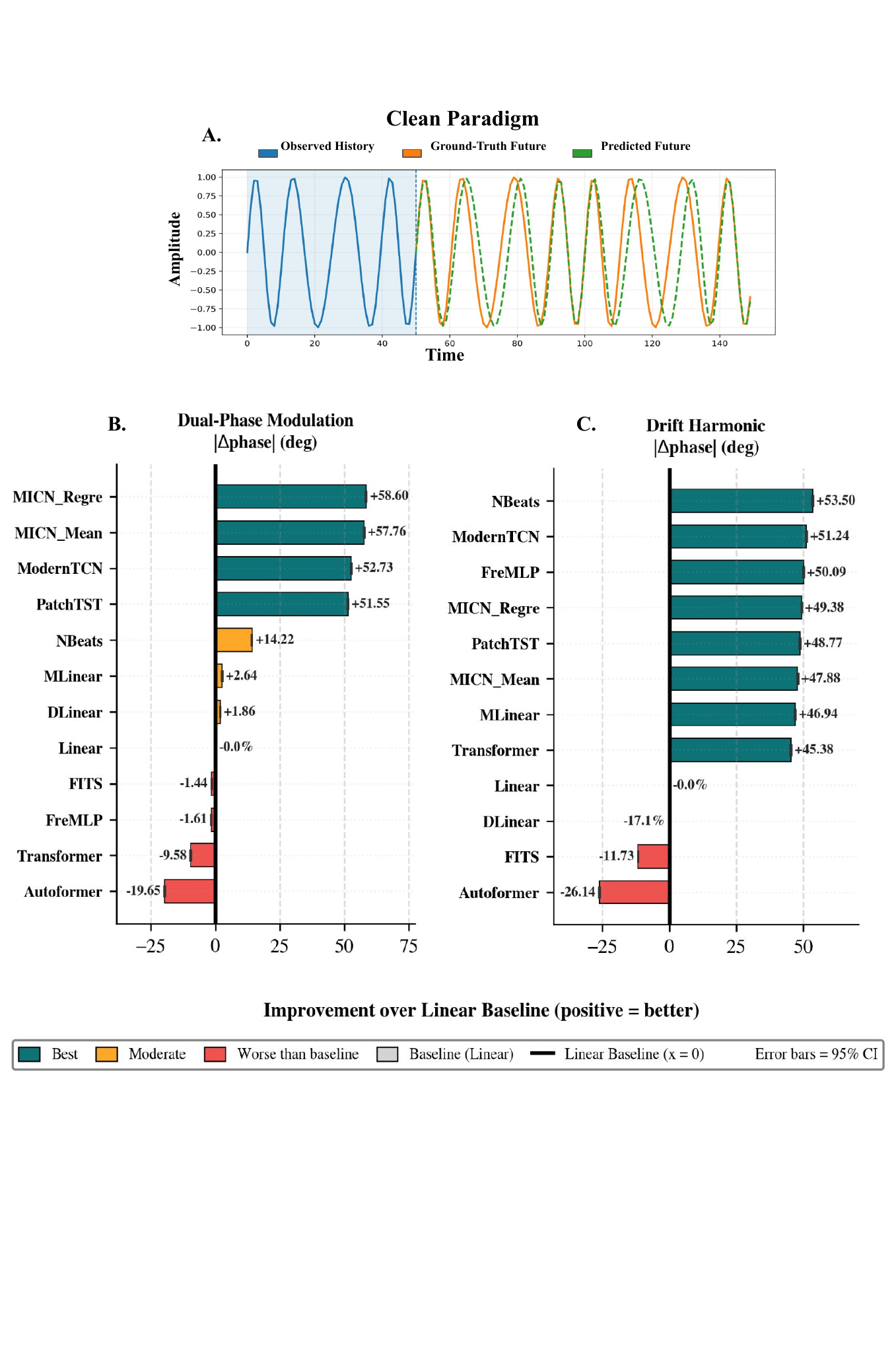}
    \caption{\textbf{PatchTST, MICN, and ModernTCN achieve $>50^{\circ}$ phase-accuracy improvement on dual-phase signals, but rankings reverse on simpler signals.} (\textbf{a})~Clean paradigm: models observe history (blue) and predict the future (green) against ground truth (orange); peak misalignment is invisible to pointwise metrics. (\textbf{b})~Phase-accuracy improvement ($|\Delta\text{phase}|$) over the linear baseline on dual-phase signals. (\textbf{c})~The same on drift-harmonic signals, where the ranking reverses (NBeats and ModernTCN lead and Transformer exceeds baseline), indicating that broad attention preserves phase on simple spectra but fails under multi-frequency modulation. Error bars are 95\% confidence intervals; tests are paired $t$-tests with Holm correction.}
    \label{fig:clean}
\end{figure}

We began by asking which families preserve oscillatory structure with no perturbation present, a baseline that any reliable digital-twin candidate must pass. On the most demanding signals (dual-phase modulation), families separated sharply (Fig.~\ref{fig:clean}a,b). MICN\_Regre ($+58.60^{\circ}$), MICN\_Mean ($+57.76^{\circ}$), ModernTCN \cite{luo2024moderntcn} ($+52.73^{\circ}$), and PatchTST ($+51.55^{\circ}$) each improved phase accuracy by more than $50^{\circ}$ over the linear baseline (all $p < 0.05$), while Transformer ($-9.58^{\circ}$) and Autoformer \cite{wu2021autoformer} ($-19.65^{\circ}$) degraded below it, indicating that broad attention disrupts phase under multi-frequency modulation. FreMLP showed a small deficit ($-1.61^{\circ}$, $p < 0.05$) consistent with spectral leakage.

The ranking shifted systematically with complexity. On drift-harmonic signals, where a single slowly varying frequency dominates, nearly all architectures surpassed baseline, including Transformer ($+45.38^{\circ}$), while FITS ($-11.73^{\circ}$) and Autoformer ($-26.14^{\circ}$) degraded. NBeats showed the strongest complexity dependence, rising from $+14.22^{\circ}$ on dual-phase to $+43.72^{\circ}$ on single-phase and $+53.50^{\circ}$ on drift-harmonic signals; FreMLP followed the same pattern ($-1.61^{\circ}$ to $+50.09^{\circ}$). Autoformer remained the weakest architecture across all families and dimensions. In short, architectures processing the input through localized temporal windows (PatchTST, MICN, ModernTCN) preserved oscillatory structure even on complex signals, whereas global mappings and full-sequence attention held fidelity only when spectral structure was simple. This is the earliest instance of a pattern that recurs under every perturbation below.

\subsection*{Localized architectures degrade gracefully under noise}

\begin{figure}[!htbp]
    \centering
    \includegraphics[clip, trim=1.5cm 9.5cm 0.8cm 0.5cm, width=0.9\textwidth]{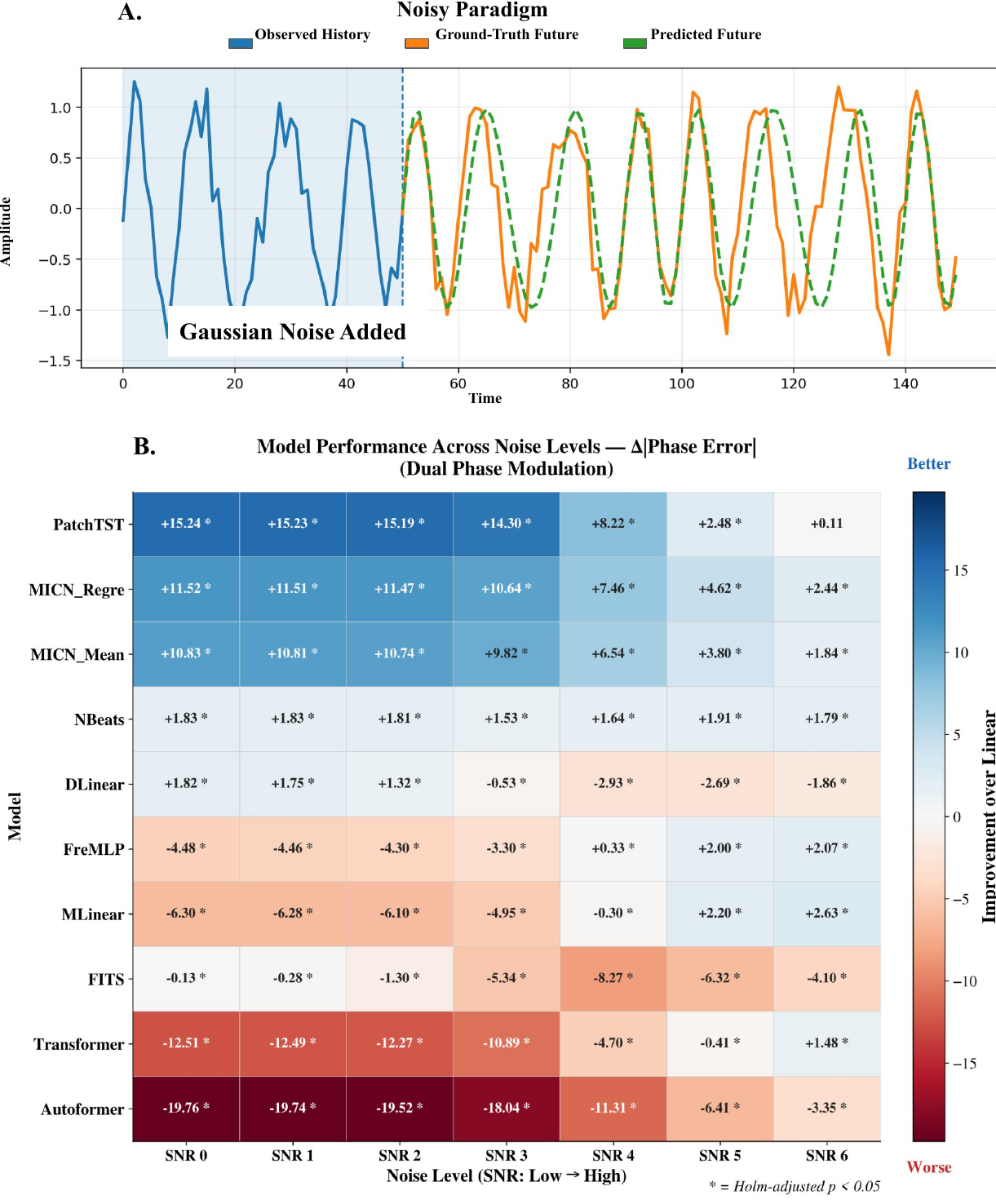}
    \caption{\textbf{PatchTST and MICN variants maintain phase accuracy under noise; linear models degrade systematically.} (\textbf{a})~Noisy paradigm: models trained on clean signals; at test time Gaussian noise is added to observed history (blue) while the future remains clean. (\textbf{b})~Phase-error difference across increasing noise (SNR~0 to SNR~6, higher = more severe) on dual-phase signals. Per-signal-family analyses are in Supplementary Figs.~\ref{fig:supp_noise_single}--\ref{fig:supp_noise_drift}. Differences marked $*$ are Holm-corrected $p < 0.05$.}
    \label{fig:noise}
\end{figure}

Biosignals are invariably corrupted by motion artifact in PPG, impedance fluctuation in ECG, and muscle activity in EEG \cite{clifford2006advanced, sornmo2005bioelectrical}. Training on clean signals and testing on history corrupted with additive Gaussian noise across six SNR levels (clean future targets; Methods), the locality pattern persisted (Fig.~\ref{fig:noise}). On dual-phase signals, PatchTST led at low-to-moderate corruption ($+15.24^{\circ}$ at SNR~0; $+8.22^{\circ}$ at SNR~4), while MICN\_Regre and MICN\_Mean retained significant improvement across the full range, including under severe corruption (SNR~6: $+2.44^{\circ}$ and $+1.84^{\circ}$) where PatchTST's advantage fell to non-significance ($+0.11^{\circ}$). Autoformer showed the largest degradation at every level ($-19.76^{\circ}$ to $-3.35^{\circ}$), and DLinear crossed from positive to negative as noise rose ($+1.82^{\circ}$ at SNR~0 to $-1.86^{\circ}$ at SNR~6).

A consistent division of labor emerged across families: PatchTST led under mild noise, while MICN's multi-scale convolution gave more stable retention under severe corruption. On single-phase signals MICN\_Regre held $+26.37^{\circ}$ at SNR~6, outperforming PatchTST by $+8.91^{\circ}$. Autoformer remained the worst performer across all three families. The complementary robustness profiles of PatchTST and MICN suggest that architecture choice should be matched to the expected noise regime, or combined in an ensemble.

\subsection*{Frequency-shift adaptation is architecture-dependent}

\begin{figure}[!htbp]
    \centering
    \includegraphics[clip, trim=1.5cm 11.0cm 0.8cm 0.5cm, width=0.8\textwidth]{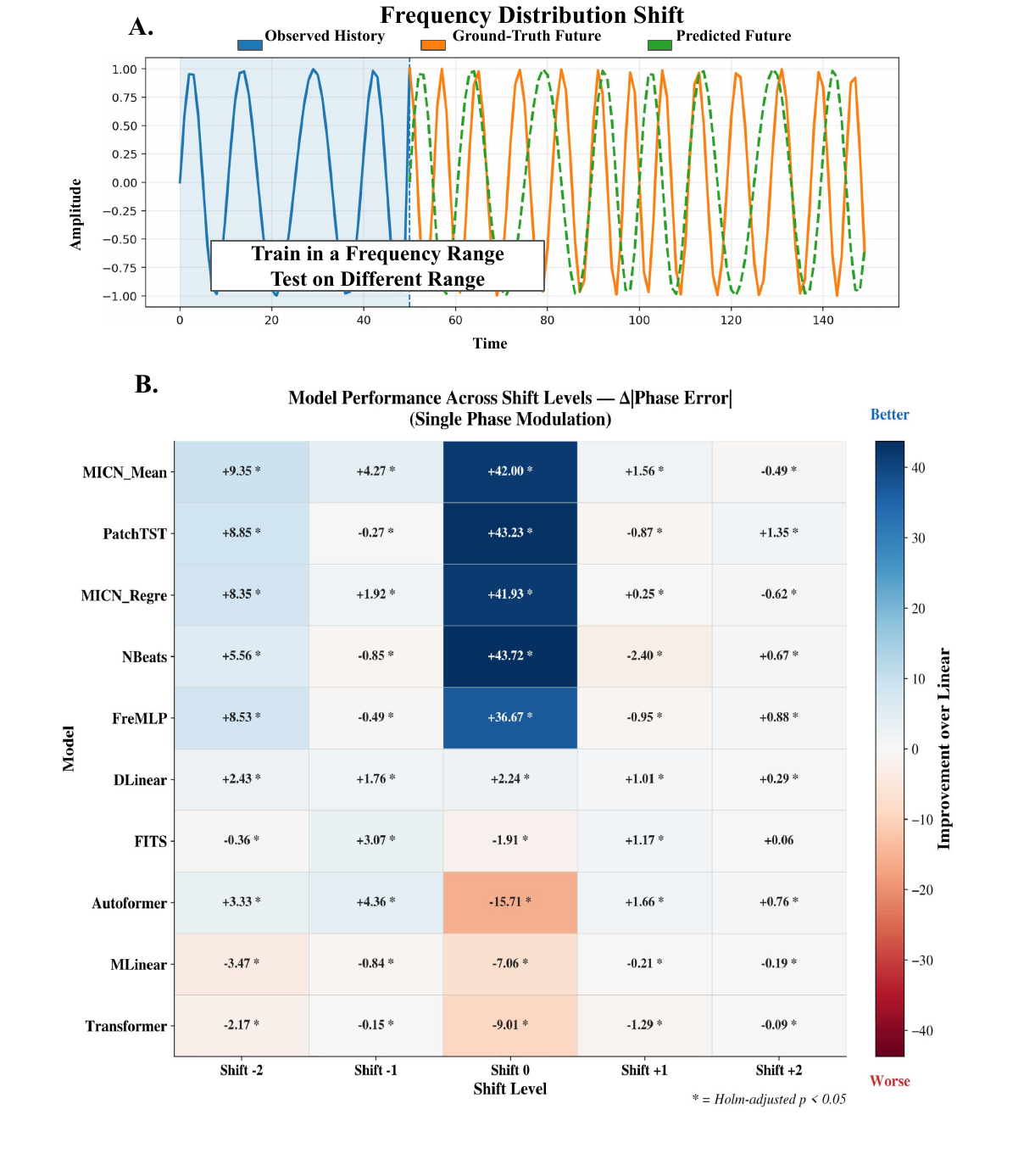}
    \caption{\textbf{PatchTST and MICN variants adapt better to rhythm changes; linear and full-sequence attention models degrade across frequency shifts.} (\textbf{a})~Frequency-shift condition: models trained on one frequency range (history, blue) forecast into a shifted range (future, orange/green). (\textbf{b})~Phase-error difference across shift levels (Shift $-2$ to $+2$) on single-phase signals. Per-signal-family analyses are in Supplementary Figs.~\ref{fig:supp_shift_dual}--\ref{fig:supp_shift_drift}. Differences marked $*$ are Holm-corrected $p < 0.05$.}
    \label{fig:shift}
\end{figure}

Physiological rhythms are nonstationary: heart rate rises with exercise, respiratory rate changes with posture, and EEG bands shift across arousal states \cite{clifford2006advanced, sornmo2005bioelectrical}. Testing models trained on one frequency band on distributions displaced by $-2$ to $+2$ range-widths (Methods), the localized architectures again led at moderate shift (Fig.~\ref{fig:shift}). On single-phase signals, NBeats ($+43.72^{\circ}$), PatchTST ($+43.23^{\circ}$), MICN\_Mean ($+42.00^{\circ}$), and MICN\_Regre ($+41.93^{\circ}$) all exceeded $+40^{\circ}$ at no shift; under shift these declined gracefully rather than collapsing, with MICN\_Mean ($+9.35^{\circ}$) and PatchTST ($+8.85^{\circ}$) retaining the most fidelity at Shift~$-2$. By contrast, Transformer ($-9.01^{\circ}$) and Autoformer ($-15.71^{\circ}$) degraded below baseline even without any shift, indicating that global attention suppresses rather than tracks frequency-specific phase relationships.

Degradation scaled with complexity and showed informative asymmetries. On dual-phase signals the collapse was severe and roughly symmetric (MICN\_Regre: $+58.60^{\circ}$ at Shift~0 to $-4.57^{\circ}$ at Shift~$-2$); on drift-harmonic signals it was directional, with most architectures degrading more under downward than upward shifts (MICN\_Mean: $-10.32^{\circ}$ at Shift~$-2$ versus $-1.94^{\circ}$ at Shift~$+2$), implying downward shifts are harder for models trained on higher-frequency bands. No architecture retained more than approximately $10^{\circ}$ at extreme shifts ($\pm 2$), marking a boundary on out-of-distribution generalization; within that boundary, PatchTST and MICN provided the strongest adaptation where linear and attention models had already failed.

\subsection*{PatchTST and ModernTCN adapt fastest to observable state transitions}

The most clinically consequential events, including atrial-fibrillation onset, the interictal-to-ictal transition, and sleep-stage shifts \cite{sornmo2005bioelectrical}, raise a question no deterministic forecaster can sidestep: not whether it can predict an unseen transition, but how quickly it adapts once evidence of the new state enters the observable history. We introduced deterministic frequency transitions at varying distances from the forecast boundary and tracked phase recovery as post-transition context accrued (in-context tags H2--H40; Methods).

\begin{figure}[!htbp]
    \centering
    \includegraphics[clip, trim=1.5cm 7.5cm 0.8cm 1.0cm, width=1.0\textwidth]{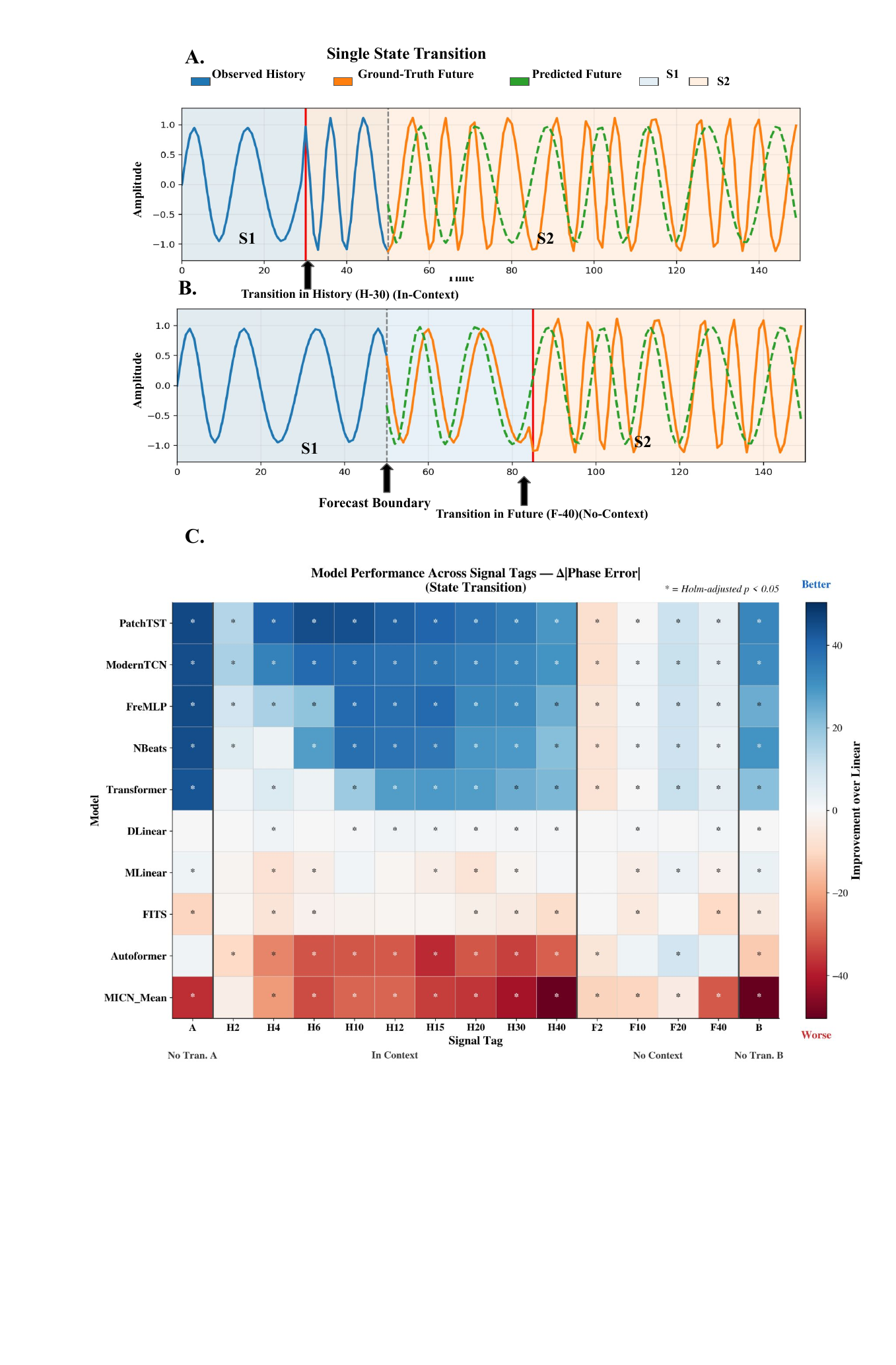}
    \caption{\textbf{PatchTST and ModernTCN adapt fastest to observable state transitions.} (\textbf{a})~Single S1$\rightarrow$S2 transition within the observed history (H-30, in-context), providing post-transition context before the forecast boundary. (\textbf{b})~Transition in the unobserved future (F-40, no-context). (\textbf{c})~Phase improvement over the linear baseline versus post-transition context: in-context (H2--H40) and no-context (F2--F40), where all models forecast the pre-transition state. Per-dimension analyses are in Supplementary Figs.~\ref{fig:supp_transition_mae}--\ref{fig:supp_transition_freq}. Differences marked $*$ are Holm-corrected $p < 0.05$.}
    \label{fig:transition}
\end{figure}

Architectures processing short, overlapping windows adapted fastest (Fig.~\ref{fig:transition}). PatchTST and ModernTCN recovered to within $20^{\circ}$ of their no-transition baselines ($4.7^{\circ}$ and $5.8^{\circ}$) within just 6 timesteps of context (both $p < 0.05$ versus Linear at H6). FreMLP and NBeats needed approximately 15 timesteps; Transformer required approximately 40, reaching $15.4^{\circ}$ only at H40, with global attention diffusing transition evidence across the full context rather than concentrating it at the change point. When transitions fell in the unobserved future (F2--F40), all models forecast the pre-transition state, confirming the paradigm tests adaptation to evidence, not clairvoyance. The hierarchy held beyond phase: PatchTST and ModernTCN recovered fastest across amplitude and frequency too, whereas MICN\_Mean dissociated, with strong frequency adaptation ($p < 0.05$ at H10) but persistent amplitude degradation (H6 through H40), recovering rhythm before magnitude and illustrating why separate diagnostics are necessary. At 10~Hz sampling, PatchTST's 6-timestep recovery corresponds to approximately 0.6~s of context versus approximately 4~s for Transformer, a concrete measure of adaptation lag. Notably, the MICN variants that excelled under noise and shift did not lead here, the only paradigm to separate the localized architectures from one another.

\subsection*{No architecture reliably recovers stochastic switching dynamics}

Finally, cardiac rhythm, sleep staging, and autonomic tone fluctuate stochastically \cite{clifford2006advanced, tong1990non}, and conditions such as paroxysmal atrial fibrillation are defined by switching statistics rather than any single waveform \cite{sornmo2005bioelectrical}. We generated signals from a two-state Markov chain ($p \in \{0.10, 0.30, 0.50, 0.70, 0.90\}$) and asked whether each forecast preserved the switching statistics of the ground-truth future, scored by symmetric KL divergence between switching-probability distributions recovered by a standardized two-state Gaussian HMM probe applied identically to true and predicted signals (recovery summarized at symmetric KL $< 0.05$; Methods).

This paradigm exposed a limitation no architecture overcame. PatchTST was the only model to recover switching at more than one transition probability ($p = 0.70$, KL $= 0.008$; $p = 0.90$, KL $= 0.046$), both at high probabilities where frequent alternation supplies more within-window evidence. Six architectures recovered at a single probability each (DLinear, Linear, FreMLP, ModernTCN, MICN\_Mean, MICN\_Regre; KL $= 0.014$--$0.034$), and five failed everywhere (Autoformer, FITS, MLinear, NBeats, Transformer; KL up to $2.020$). NBeats' complete failure is striking given its fast deterministic-transition adaptation, showing that adapting to an observable transition and preserving probabilistic switching are distinct capabilities. Even PatchTST succeeded in fewer than half the conditions. The continuous KL values and a threshold-sensitivity analysis preserve this ranking, and the uniform difficulty across all four families indicates the limitation is architectural-general: deterministic input-output forecasters are suited to waveform extrapolation, not to reproducing transition structure.

\begin{figure}[!htbp]
    \centering
    \includegraphics[clip, trim=2.6cm 11.5cm 0.8cm 1.0cm, width=0.9\textwidth]{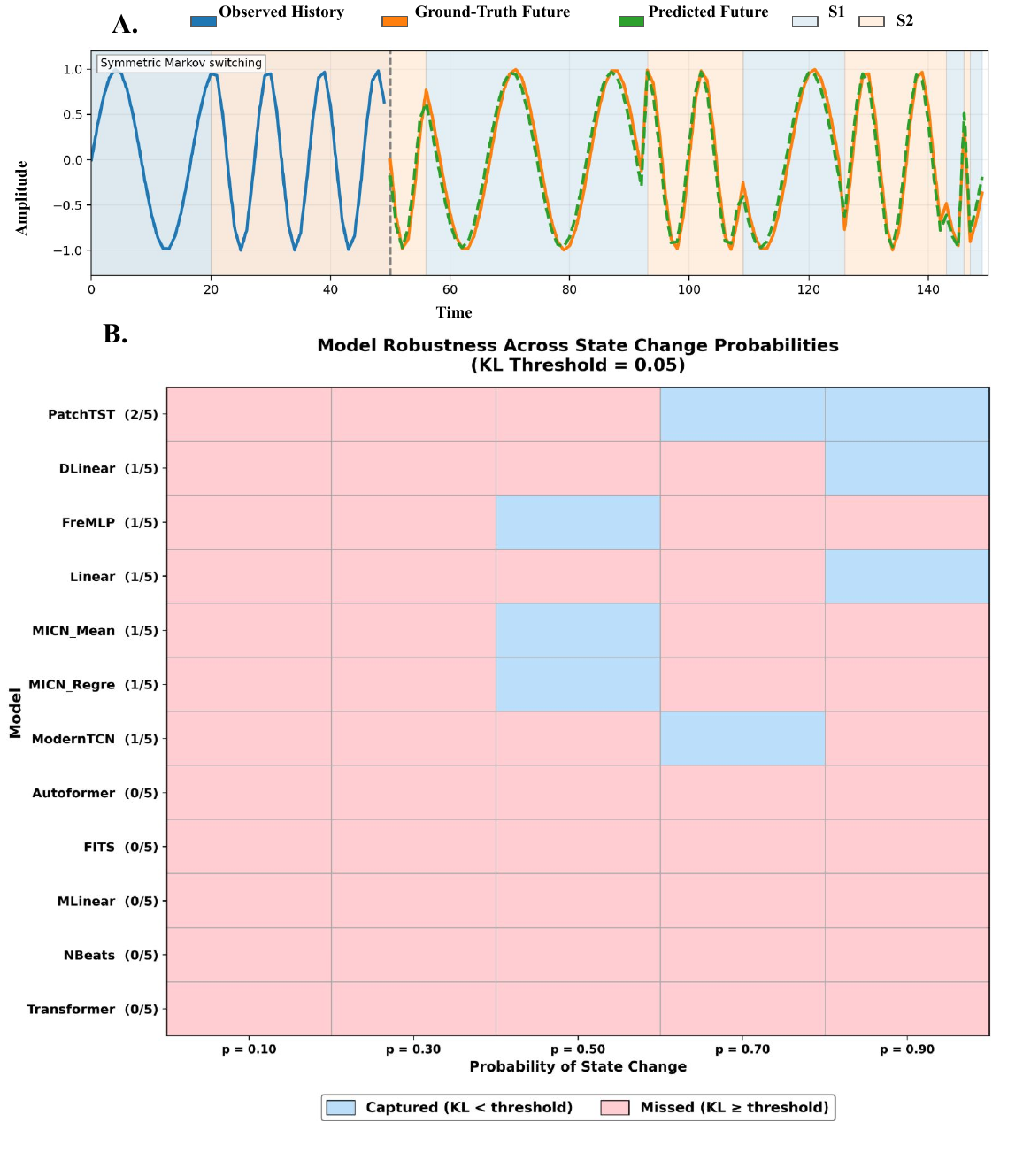}
    \caption{\textbf{PatchTST partially recovers switching dynamics; most architectures show limited recovery.} (\textbf{a})~Markov-switching signal alternating between states S1 and S2 with transition probability $p$. (\textbf{b})~Pass/fail summary across $p = 0.10$ to $0.90$ using the HMM probe; blue denotes recovery (symmetric KL $< 0.05$), red limited recovery. The HMM is a standardized downstream probe of switching statistics, not the true latent simulator. Full continuous KL values are in Supplementary Table~\ref{tab:kl_full}; threshold-sensitivity analysis in Supplementary Fig.~\ref{fig:kl_threshold_sensitivity}.}
    \label{fig:markov}
\end{figure}

\subsection*{Localized temporal processing achieves the most balanced fidelity}

Each preceding test isolates one perturbation, but a deployed digital twin faces them jointly. To identify architectures that hold fidelity across all five paradigms simultaneously, we performed a Pareto analysis, training each model separately per paradigm to isolate architectural effects (Methods). Four models reached the frontier, but for different reasons: MICN\_Regre and MICN\_Mean through clean, noise, and shift performance but weak state transitions (normalized $0.54$ and $0.00$); ModernTCN through state transitions alone ($0.97$) at the cost of noise ($0.31$) and shift ($0.23$). Only PatchTST exceeded $0.8$ on every dimension (clean $0.97$, noise $0.98$, shift $1.00$, transition $1.00$, Markov $1.00$); NBeats, though dominated, placed third overall ($0.85$ mean, nothing below $0.77$). No other architecture exceeded $0.8$ on more than three dimensions.

The same answer thus emerged under every perturbation: localized temporal processing preserved physiological fidelity, global mappings and full-sequence attention did not, and no deterministic architecture captured stochastic switching. Among all models, PatchTST held the most balanced fidelity across noise, shift, deterministic transitions, and switching. For health-signal digital twins, architectures with localized temporal processing are therefore the strongest candidates for further development, though final model choice should be matched to the perturbation dimensions most relevant to the target application and validated on real patient recordings.

\begin{figure}[!htbp]
    \centering
    \includegraphics[width=0.8\linewidth]{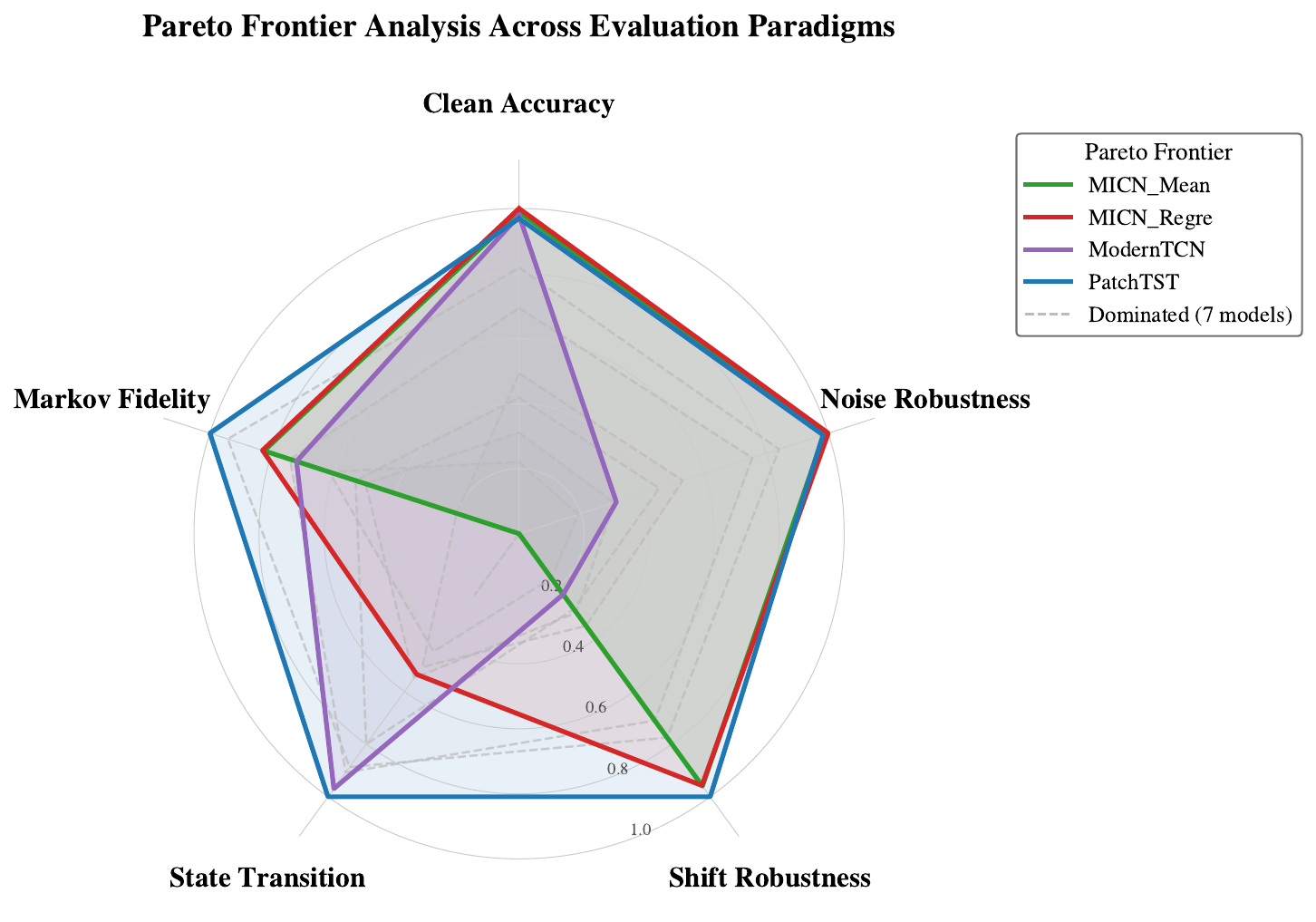}
    \caption{\textbf{PatchTST maintains the most balanced fidelity across all evaluation paradigms.} Radar plot of 11 models across five paradigms (clean accuracy, noise robustness, shift robustness, state transition, Markov fidelity), each normalized to 0--1. Solid polygons denote Pareto-optimal models; dashed gray polygons denote dominated models. Axes represent aggregate improvement over the linear baseline, averaged across signals and conditions. Full normalized scores are in Supplementary Table~\ref{tab:pareto_scores}.}
    \label{fig:pareto}
\end{figure}

\section{Discussion}

Using TimeSynth, we found that conventional accuracy metrics obscure how forecasting architectures preserve the oscillatory, nonstationary, and stochastic dynamics of physiological signals, and that architectural choice strongly determines what a digital twin forecast can faithfully represent. This speaks directly to a recognized gap: the NASEM report identifies verification, validation, and uncertainty quantification as prerequisites for trustworthy digital twins \cite{nasem2024}, yet such frameworks for biosignal forecasting remain underdeveloped \cite{tudor2025scoping,sel2025vvuq}. Synthetic signals with analytically known dynamics are what make this verification possible: ground-truth amplitude, frequency, and phase trajectories cannot be recovered from real recordings, so controlled generation is a prerequisite for the stress test rather than a weakness of it, the same enabling role that synthetic data plays in other physiological deep-learning systems \cite{maksymenko2023myoelectric}. TimeSynth accordingly provides a controlled preclinical environment for exposing dynamical failure modes before forecasting models are coupled to real patient time series.

The central finding is a dissociation between pointwise accuracy and dynamical fidelity (Fig.~\ref{fig:dissociation}): models with comparable MAE differed by up to $53^{\circ}$ in phase error, so a forecast can appear accurate by standard benchmarks while mistracking oscillatory timing. The clinical scale of such an error depends on signal frequency. A $53^{\circ}$ phase error corresponds to roughly 123~ms for a 1.2~Hz cardiac rhythm, a magnitude relevant to R-R interval variability and rhythm monitoring \cite{clifford2006advanced}; to 12--18~ms in the EEG alpha band (8--12~Hz), where it may disrupt phase-sensitive neural analyses \cite{sornmo2005bioelectrical}; and to nearly 590~ms for respiratory-modulated PPG near 0.25~Hz, potentially obscuring respiratory-cardiovascular coupling \cite{sornmo2005bioelectrical}. These translations do not demonstrate clinical failure, but they identify timing errors invisible to aggregate accuracy that warrant evaluation before deployment, consistent with concerns that many digital twin studies validate only on aggregate measures \cite{tudor2025scoping}.

Across paradigms, the dominant determinant of fidelity was the locality of temporal processing. Patch-based and convolutional architectures preserved phase and frequency structure most consistently under clean, noisy, and shifted conditions (Figs.~\ref{fig:clean}--\ref{fig:shift}), and PatchTST achieved the most balanced fidelity across all five paradigms (Fig.~\ref{fig:pareto}), consistent with patch tokenization capturing local periodicity more effectively than pointwise mappings or broad-context attention \cite{nie2022time} and with prior reports that vanilla Transformers underperform on structured time series \cite{zeng2023transformers}. The same mechanism accounts for adaptation speed after observable state changes: PatchTST and ModernTCN recovered to within $20^{\circ}$ of baseline within six timesteps (about 0.6~s at 10~Hz), whereas full-sequence attention required roughly 40, diluting transition evidence across the input (Fig.~\ref{fig:transition}). Locality was not sufficient on its own, however. MICN variants were strong under noise and shift but weak under deterministic transitions, and NBeats adapted to deterministic transitions yet failed to preserve stochastic switching, so no single inductive bias guaranteed fidelity across all regimes. The frequency-shift results marked a shared ceiling: no architecture retained more than about $10^{\circ}$ of improvement at extreme shifts ($\pm 2$ range-widths), with a directional asymmetry suggesting that rhythms moving away from the training band are intrinsically harder to track.

The sharpest limitation appeared under stochastic switching (Fig.~\ref{fig:markov}), where most architectures failed and even PatchTST recovered the switching statistics in only two of five regimes. We assessed recovery with a two-state Gaussian HMM applied identically to true and predicted signals as a standardized downstream probe, not as an oracle or as the true latent simulator, asking whether a forecast preserves enough temporal-spectral structure for comparable switching statistics to be inferred; the conclusion held across continuous KL values and a threshold-sensitivity analysis. Interpreted this way, the failure is informative: deterministic point forecasters can reproduce plausible waveforms while discarding distributional state-transition structure. Forecasting models for digital twins may therefore require explicit latent-state, probabilistic, or distributional mechanisms rather than point-estimate prediction alone.

These results reframe how forecasting models for digital twins should be selected and validated. Because no single architecture dominates every fidelity dimension, the practical conclusion is not a universal winner but a principled selection rule, mirroring how benchmarking of synthetic-data generators in other biomedical domains resolves into use-case-specific guidance rather than a single ranking \cite{yan2022multifaceted}. Selection should therefore weigh balanced fidelity across perturbations rather than clean-condition accuracy alone, with the weighting matched to the use case: ambulatory cardiac monitoring may prioritize noise robustness, intensive care may prioritize rapid adaptation to state changes, and sleep or seizure applications may prioritize preservation of switching statistics. Beyond benchmarking, TimeSynth offers two further roles for model development. Its signals with known ground-truth dynamics can support physiologically structured pretraining before fine-tuning on limited patient data, consistent with evidence that well-designed synthetic data can rival domain-specific training \cite{taga2025timepfn}, and its fidelity diagnostics provide interpretable feedback on which dynamical properties an architecture preserves or destroys. A natural extension is to move from analytic generators to constrained generative models trained on real biosignals, with latent variables aligned to amplitude, frequency, phase, and state transitions, increasing realism while retaining the interpretability that makes the framework useful for verification.

Several limitations qualify these conclusions. The benchmark uses univariate signals, fixed input-output horizons, and deliberately simplified perturbations, whereas clinical digital twins typically require multivariate inputs, irregular sampling, patient-specific adaptation, and uncertainty quantification. Physiological dynamics also vary with age, sex, comorbidity, sensor type, and care setting \cite{goldberger2000physiobank}, so architectures intended for broad use should be evaluated across patient subgroups to ensure that fidelity failures are not unevenly distributed. Finally, the switching analysis depends on a downstream HMM probe and should not be read as definitive latent-state validation; comparing alternative probes and distributional distances on real state-transition datasets is an important next step.

In summary, pointwise accuracy is insufficient for evaluating forecasting models intended for health-signal digital twins. Architectures with localized temporal processing, particularly PatchTST, MICN, and ModernTCN, better preserved amplitude, frequency, and phase fidelity under controlled perturbations, while deterministic architectures generally struggled with stochastic switching. TimeSynth provides a controlled preclinical framework for exposing these failure modes, guiding architecture development, and clarifying which models warrant further evaluation in patient-specific digital twin workflows.

\section{Methods}
 
\subsection*{Synthetic signal generation}
 
We generated synthetic signals from closed-form parametric models whose terms map to interpretable physiological quantities, in three families each grounded in a real biosignal modality. Single phase-modulated (SPM) signals model quasi-periodic dynamics with one source of phase variation, using a carrier whose phase is sinusoidally modulated at a slower rate:
\begin{equation}
x(t) = A \sin\!\bigl(2\pi f\, t + \beta \sin(2\pi f_{\mathrm{mod}}\, t)\bigr) + c,
\label{eq:spm}
\end{equation}
where $A$ is the carrier amplitude, $f$ the carrier frequency (Hz), $\beta$ the phase-modulation index, $f_{\mathrm{mod}}$ the modulation frequency (Hz), and $c$ a DC offset. Dual phase-modulated (DPM) signals sum two independently modulated components with a shared offset, capturing multi-component harmonic structure:
\begin{equation}
x(t) = \sum_{i=0}^{1} A_i \sin\!\bigl(2\pi f_i\, t + \beta_i \sin(2\pi f_{\mathrm{mod},i}\, t)\bigr) + c,
\label{eq:dpm}
\end{equation}
with each component's parameters ($A_i$, $f_i$, $\beta_i$, $f_{\mathrm{mod},i}$) sampled independently from the SPM bounds. Drift-harmonic (DH) signals combine a carrier with a time-varying amplitude envelope and linear trend, reproducing slow nonstationarity in peripheral cardiovascular signals:
\begin{equation}
x(t) = (1 + \varepsilon\, t)\sin(2\pi f\, t + \varphi) + a\, t,
\label{eq:dh}
\end{equation}
where $\varepsilon$ is the envelope drift rate (fixed at $-0.05$ across realizations), $\varphi$ the initial phase, and $a$ the linear trend coefficient; DH signals were min-max normalized to $[0, 1]$.
 
To ground parameter ranges in physiology, we fit each model to real segments from three public datasets: the MIT-BIH Arrhythmia Database \cite{moody2001impact} (ECG, 48 records, 360~Hz), PPG-DaLiA \cite{ppg} (PPG/BVP, 15 subjects, 64~Hz), and the CHB-MIT Scalp EEG Database \cite{shoeb2009application} (EEG, channel FP1-F7). SPM and DPM were fit to the smooth oscillatory S-Q-interval baseline of MIT-BIH ECG, DH to PPG-DaLiA blood-volume-pulse segments, and the extended DPM model to CHB-MIT EEG. Each model was implemented as a differentiable parametric module in PyTorch and optimized with Adam over overlapping sliding windows, with physiological plausibility enforced by per-step box-constraint clamping on learnable parameters. The fitted ranges for carrier frequency, modulation depth, amplitude, phase, offset, and drift rate defined the generation space for each family (Fig.~\ref{fig:fitting}; Supplementary Tables~\ref{tab:param_bounds},~\ref{tab:fitting_hyperparams_alt}).
 
Signals were sampled at $f_s = 10$~Hz for 300~s (3,000 points per instance). The 10~Hz rate captures modulation-envelope dynamics relevant to digital twin forecasting rather than high-frequency waveform morphology, since cardiac, respiratory, and neural monitoring often operate on derived quantities such as heart rate, respiratory rate, and spectral power that evolve over seconds to minutes. Each family contained 100 unique parametric realizations split into 70 training, 10 validation, and 20 test instances, with uniqueness enforced by deterministic parameter hashing (Supplementary Appendix~\ref{app:generation}).
 
\subsection*{Controlled evaluation paradigms}
 
We evaluated forecasters under five conditions that stress-test dynamical fidelity in settings relevant to health-signal digital twins. In the \textbf{clean} paradigm, models train and test on noise-free signals, establishing a per-family ceiling for each fidelity dimension. For \textbf{noise robustness}, models trained on clean signals are tested across seven SNR levels, a clean baseline (SNR~0) plus six additive-Gaussian-noise levels from 40 to 1~dB (SNR~1 to SNR~6, higher index more severe), with noise power calibrated relative to zero-mean signal power so that the DC offset does not inflate the estimate (Supplementary Appendix~\ref{app:noise}). For \textbf{frequency distribution shift}, models trained on one frequency range are tested on bands displaced by $-2$ to $+2$ range-widths; within each shifted band, 20 test signals are generated with carrier frequency sampled uniformly from the shifted range and all non-frequency parameters held at training bounds (Supplementary Appendix~\ref{app:shift}; Supplementary Table~\ref{tab:freq_shift}). For the \textbf{single state transition} paradigm, a deterministic frequency change-point is placed either within the observed history (in-context, tags H2--H40) or in the unobserved future (no-context, tags F2--F40), with phase continuity maintained by recursive accumulation to avoid artificial discontinuities at the change-point; each split contains 600 training, 100 validation, and 200 test instances (Supplementary Appendix~\ref{app:state-transition}). For \textbf{Markov switching}, signals alternate stochastically between two frequency states under a symmetric two-state Markov chain with transition probabilities $p \in \{0.10, 0.30, 0.50, 0.70, 0.90\}$, with 70 training, 10 validation, and 20 test instances per value of $p$ (Supplementary Appendix~\ref{app:markov}). Together these paradigms approximate sensor noise, evolving rhythms, abrupt state changes, and stochastic variability encountered in patient monitoring; full specifications are provided in Supplementary Appendix~\ref{app:paradigms}.
 
\subsection*{Forecasting models and inductive-bias hypotheses}
 
We benchmarked 11 models spanning four architectural families selected to represent dominant approaches in time series forecasting and to isolate how temporal processing affects dynamical fidelity. The \textbf{linear} family (Linear, DLinear \cite{zeng2023transformers}, FITS \cite{xu2023fits}) maps input to output through learned linear projections. The \textbf{MLP} family (MLinear, NBeats \cite{oreshkin2019n}, FreMLP \cite{yi2023frequency}) adds nonlinear transformations, with FreMLP operating in the frequency domain. The \textbf{convolutional} family (ModernTCN \cite{luo2024moderntcn}, MICN \cite{wang2023micn}) processes inputs through convolutional kernels over local windows, with MICN evaluated in two trend-prediction variants (MICN\_Mean, MICN\_Regre). The \textbf{transformer} family (Transformer, Autoformer \cite{wu2021autoformer}, PatchTST \cite{nie2022time}) uses attention ranging from full-sequence to patch-based.
 
Our analysis groups these models by the inductive bias most relevant to physiological dynamics, namely the locality of temporal processing. Linear and full-sequence attention models (Transformer, Autoformer) integrate information globally, whereas convolutional (ModernTCN, MICN) and patch-based (PatchTST) models process short local windows; although PatchTST is attention-based, its patchwise tokenization first constrains the model to local temporal segments, giving it an effectively localized bias. We therefore hypothesized that localized processing would better preserve phase, short-range frequency structure, and recent state evidence, so that convolutional and patch-based models would retain fidelity under clean conditions, degrade more gracefully under noise, adapt faster once post-transition context becomes observable, and remain more robust under moderate frequency shifts, whereas linear and full-sequence attention models would preserve fidelity only when spectral structure is simple. We further expected stochastic Markov switching to be the hardest paradigm for all deterministic forecasters, since it requires preserving transition probabilities and dwell-time structure rather than extrapolating a single most-likely waveform.
 
All models were configured for univariate forecasting with a 50-step input and a 100-step prediction horizon. The 50-step input (5~s at 10~Hz) spans roughly five oscillatory cycles at a 1~Hz cardiac frequency, sufficient to observe the carrier and its modulation envelope; the 100-step horizon (10~s) tests whether fidelity is sustained over an extended window. Models were trained under a unified protocol (AdamW optimizer, OneCycleLR scheduling, MSE loss, batch size 128, early stopping on validation loss), independently on each signal type and paradigm to isolate architectural effects from task-specific optimization. Architectural descriptions, model-specific hyperparameters, and family-level deviations are reported in Supplementary Appendix~\ref{app:models}.
 
\subsection*{Fidelity metrics}
 
We evaluate forecasts along three complementary dimensions that isolate distinct failure modes: whether the prediction preserves oscillation magnitude, rate, and timing. Standard pointwise metrics conflate these and can mask a failure in one dimension behind acceptable performance in another. Amplitude error is the mean absolute error (MAE) over the $H$-step horizon,
\begin{equation}
\text{MAE} = \frac{1}{H} \sum_{t=1}^{H} \left| \hat{y}(t) - y(t) \right|,
\label{eq:mae}
\end{equation}
capturing oscillation magnitude, the property most directly measured by conventional benchmarks. Frequency error captures mismatch in dominant oscillation rate, computed from the power spectrum $P(k) = |\mathcal{F}(x)|^2$ after DC removal. The peak bin $k^*$ is refined by three-point parabolic interpolation,
\begin{equation}
\hat{f} = \left(k^* + \frac{P(k^*\!-\!1) - P(k^*\!+\!1)}{2\bigl(P(k^*\!-\!1) - 2P(k^*) + P(k^*\!+\!1)\bigr)}\right) \cdot \frac{f_s}{N},
\label{eq:freq}
\end{equation}
where $N$ is the FFT length; estimates with peak power below 10\% of total spectral power are discarded, and frequency error is $|\hat{f}_{\text{pred}} - \hat{f}_{\text{true}}|$. Phase error captures temporal misalignment via the analytic signal $z(t)$, obtained by the frequency-domain Hilbert transform (zeroing negative-frequency bins, doubling positive-frequency bins, inverse-transforming with zero-padding at pad factor 2 to limit edge effects). Instantaneous phase is $\varphi(t) = \text{unwrap}\bigl(\arg(z(t))\bigr)$, and to avoid spurious estimates in low-amplitude regions the difference is evaluated only where the true instantaneous amplitude exceeds 20\% of its median,
\begin{equation}
\Delta\varphi = \frac{1}{|\mathcal{M}|} \sum_{t \in \mathcal{M}} \left| \text{wrap}_\pi\!\bigl(\varphi_{\text{pred}}(t) - \varphi_{\text{true}}(t)\bigr) \right|, \quad \mathcal{M} = \{t : |z_{\text{true}}(t)| > 0.2 \cdot \text{median}(|z_{\text{true}}|)\},
\label{eq:phase}
\end{equation}
where $\text{wrap}_\pi$ maps the difference to $(-\pi, \pi]$ and phase error is reported in degrees. A model may achieve low MAE while mistracking oscillation rate or peak timing, so only these separate diagnostics distinguish such failure modes. Full implementation details, including padding factors, smoothing windows, and edge-case handling, are provided in Supplementary Appendix~\ref{app:metrics}.
 
\subsection*{Statistical analysis}
 
Models were compared using paired tests on per-sequence metric distributions, ensuring each model was evaluated on the same sequences. For frequency and phase error, where reliability filtering can produce missing values, intersection-valid masking included a sequence only if all compared models had finite values for that metric, preventing differences in sample composition from confounding the comparison. For each metric we computed paired differences against the Linear baseline per sequence. For the clean, noise, and frequency-shift paradigms, significance was assessed with paired $t$-tests and 95\% confidence intervals ($\bar{d} \pm 1.96\, s_d / \sqrt{n}$); for state-transition analyses, where mixed in-context and no-context conditions yield non-Gaussian distributions, we used the Wilcoxon signed-rank test with tie correction. Within each metric and paradigm, $p$-values were adjusted for multiple comparisons using the Holm step-down procedure, controlling the family-wise error rate at $\alpha = 0.05$ with greater power than Bonferroni correction.
 
To assess whether forecasts preserve stochastic switching statistics, we used a two-state Gaussian hidden Markov model (HMM) as a standardized downstream probe rather than as the true latent model. The HMM was fit to windowed spectral features (Welch periodogram, window 16, hop 8) of the true history, selecting the best of eight random seeds by log-likelihood, then applied unchanged to true and predicted futures. Decoded switching-probability distributions were compared using symmetric KL divergence, which penalizes both under- and over-switching. The $\mathrm{KL_{sym}} < 0.05$ cutoff was used only for heatmap summarization; continuous values and threshold sensitivity at 0.05, 0.10, 0.15, and 0.20 are reported in Supplementary Fig.~\ref{fig:kl_threshold_sensitivity} and Supplementary Appendix~\ref{app:hmm}.
 
To identify architectures that maintain fidelity across all conditions, we performed a Pareto analysis over the five paradigms (clean accuracy, noise robustness, shift robustness, state-transition adaptation, and Markov fidelity). Within each paradigm, scores were computed as aggregate improvement over the Linear baseline and min-max normalized to $[0, 1]$ across models. A model was classified as Pareto-optimal if no other model matched or exceeded it on all five dimensions while strictly exceeding it on at least one.

\subsection*{Ethics}

All datasets used in this study (MIT-BIH Arrhythmia Database, PPG-DaLiA, CHB-MIT Scalp EEG Database) are publicly available and contain de-identified recordings. No ethics approval was required because no new human data were collected.

\subsection*{Data availability}

The three source datasets are publicly available: the MIT-BIH Arrhythmia Database and CHB-MIT Scalp EEG Database via PhysioNet, and PPG-DaLiA via the UCI Machine Learning Repository. The synthetic signal families generated for this study, together with the fitted parameter distributions used to produce them, can be regenerated exactly using the released code (\url{https://github.com/RakibulHaqueSajal/TimeSynth}), so that all evaluation paradigms are fully reproducible.

\subsection*{Code availability}

\textsc{TimeSynth} is released as reusable infrastructure at \url{https://github.com/RakibulHaqueSajal/TimeSynth}: the physiologically grounded signal generator, the controlled perturbation paradigms, the amplitude, frequency, phase, and state-transition fidelity diagnostics, and the scripts reproducing every figure and statistical test. The release is intended to let others benchmark new architectures, generate controlled ground-truth signals for pretraining, and audit fidelity in their own digital-twin workflows.


\bibliography{sn-bibliography}

\section*{Author Contributions}
\textbf{M.R.H. }conceived the TimeSynth framework, designed the synthetic signal generation pipeline, implemented the parametric models fitted to real biosignals (ECG, PPG, EEG), developed the fidelity diagnostic metrics (amplitude, frequency, and phase error), implemented and trained all 11 forecasting architectures, designed and executed all five controlled evaluation paradigms (clean, noise, frequency shift, single state transition, and Markov switching), performed the statistical analyses including Pareto frontier analysis, generated all figures, and wrote the manuscript.
\textbf{W.W.P.} supervised the project, conceived the framing of dynamical fidelity for health-signal digital twins, guided the physiological grounding of the synthetic signal families and their mapping to clinical phenomena, advised on experimental design and the controlled evaluation paradigms, contributed to interpretation of results in the clinical context, and revised the manuscript.
\textbf{S.E.} supervised the project, advised on the computational methodology including parametric model fitting and the design of the fidelity diagnostic framework, contributed to the experimental design and statistical analysis approach, provided guidance on the architectural inductive-bias hypotheses, and revised the manuscript.





\section*{Acknowledgements}

This work was supported in part by the Huntsman Mental Health Foundation, the University of Utah Digital Health Initiative, and the National Institutes of Health under grant numbers L70AG096751, 2R01MH122412-06, and 2R01MH123489-06.

\newpage


\appendix

\renewcommand{\thesection}{A\arabic{section}}
\renewcommand{\thesubsection}{A\arabic{section}.\arabic{subsection}}
\renewcommand{\thefigure}{A\arabic{figure}}
\renewcommand{\thetable}{A\arabic{table}}
\renewcommand{\theequation}{A\arabic{equation}}
\setcounter{figure}{0}
\setcounter{table}{0}
\setcounter{equation}{0}

\section*{Appendix A: Supplementary Information}
\addcontentsline{toc}{section}{Appendix A: Supplementary Information}

\noindent
This appendix provides the full technical specifications, supporting analyses, and extended results for the TimeSynth framework. It is organized as follows: \textbf{A1} synthetic signal generation, \textbf{A2} controlled evaluation paradigms, \textbf{A3} forecasting model architectures and hyperparameters, \textbf{A4} fidelity metric computation, \textbf{A5} statistical analysis, and \textbf{A6} extended
results across signal families and evaluation paradigms.

\section{Synthetic Signal Generation}
\label{app:signal-generation}

\subsection{Extended signal model for EEG fitting}
\label{app:eeg-model}

The SPM, DPM, and DH signal family equations are defined in the main text (Eqs.~1--3). For fitting to real EEG recordings, the DH model was extended with additional frequency bands and transient Gaussian spike components to accommodate the multi-band structure and transient deflections characteristic
of scalp EEG:
\begin{equation}
x(t) = \sum_{i=1}^{n_{\mathrm{bands}}} A_i\bigl(1 + e_i \sin(0.5\pi t)\bigr)\sin(2\pi f_i\, t + \varphi_i)
       + \sum_{j=1}^{n_{\mathrm{spikes}}} A_j^{(\mathrm{sp})} \exp\!\Bigl(-\frac{(t - t_{c,j})^2}{2\sigma_j^2}\Bigr)
\label{eq:eeg-extended}
\end{equation}
where $n_{\mathrm{bands}} = 2$ frequency bands and $n_{\mathrm{spikes}} = 2$ transient Gaussian spike components, $e_i$ denotes the envelope modulation depth, $A_j^{(\mathrm{sp})}$ and $t_{c,j}$ the spike amplitude and center, and $\sigma_j$ the spike width. This richer model was used exclusively for parametric fitting to CHB-MIT recordings to derive physiologically grounded parameter bounds; the synthetic DH signals used for evaluation were generated from the simpler closed-form equation defined in the main text (Eq.~3).

\subsection{Parametric fitting to real biosignals}
\label{app:fitting}

To derive physiologically grounded parameter bounds, each parametric model was fit to real signal segments from three clinical-grade datasets. Fitting was performed using differentiable parametric modules implemented in PyTorch and optimized with Adam. After each gradient step, all learnable parameters were clamped to predefined physiological bounds (box constraints). Fitting hyperparameters for each dataset are summarized in Table~\ref{tab:param_bounds}.

\paragraph{PPG fitting.}
Blood volume pulse (BVP) recordings from PPG-DaLiA (15 subjects, $f_s = 64$~Hz) were fit using the drift-harmonic model (main text Eq.~3) with learnable parameters $\varepsilon$, $f$, $\varphi$, and $a$. Segments of 5.0~s with 50\% overlap were optimized for 500 epochs using MSE loss (lr $= 0.01$). Parameter bounds: $\varepsilon \in [-0.05, 0.05]$, $f \in [0.85, 1.1]$~Hz,
$\varphi \in [-\pi/4, \pi/4]$~rad, $a \in [-0.1, 0.1]$.

\paragraph{ECG fitting.}
S-Q interval segments from the MIT-BIH Arrhythmia Database (48 records, $f_s = 360$~Hz, 10~s per record) were fit using the phase-modulated multisine model (main text Eq.~2) with $N = 2$ components. Segments of
1.0~s with 15\% overlap and 10\% blend ratio were optimized for 500 epochs using L1 loss (lr $= 0.001$). Parameter bounds: $A_i \in [0.1, 0.4]$, $f_i \in [0.5, 3.0]$~Hz, $\beta_i \in [0.01, 0.3]$, $f_{\mathrm{mod},i} \in [0.01, 0.1]$~Hz, $c \in [0.0, 1.0]$.

\paragraph{EEG fitting.}
Scalp EEG recordings from the CHB-MIT database (channel FP1-F7, 100~s extracted segments) were fit using the extended multi-band model (Eq.~\ref{eq:eeg-extended}) with $n_{\mathrm{bands}} = 2$ and
$n_{\mathrm{spikes}} = 2$. Signals were normalized to $[-1, 1]$ prior to fitting. Segments of 1.0~s were optimized for 100 epochs using L1 loss (lr $= 0.01$).

\paragraph{Cosine blending.}
Overlapping fitted segments were combined using cosine ramp weights to prevent discontinuities at segment boundaries:
\begin{equation}
w(n) = \tfrac{1}{2}\Bigl(1 - \cos\!\bigl(\tfrac{\pi n}{N-1}\bigr)\Bigr),
\quad n = 0, 1, \ldots, N-1
\end{equation}
where $N$ is the segment length. The final fitted signal at each time point is the weighted average $\hat{x}(t) = \sum_k w_k(t)\hat{x}_k(t) \big/ \sum_k w_k(t)$, summing over all segments $k$ covering time $t$.

\begin{table}[ht]
\centering
\caption{Parameter bounds for synthetic signal generation. All parameters are sampled from uniform distributions over the stated ranges unless marked as fixed.}
\label{tab:param_bounds}
\begin{tabular}{llccc}
\toprule
\textbf{Signal family} & \textbf{Parameter} & \textbf{Symbol} & \textbf{Range} & \textbf{Unit} \\
\midrule
SPM & Carrier amplitude       & $A$              & [0.1, 0.1227]            & --   \\
SPM & Carrier frequency       & $f$              & [0.6782, 1.4112]         & Hz   \\
SPM & Modulation index        & $\beta$          & [0.01, 0.3]              & rad  \\
SPM & Modulation frequency    & $f_{\mathrm{mod}}$ & [0.01, 0.1]            & Hz   \\
SPM & DC offset               & $c$              & [0.1937, 0.7418]         & --   \\
\midrule
DPM & Per-component parameters & $A_i, f_i, \beta_i, f_{\mathrm{mod},i}$ & Same as SPM & -- \\
DPM & Shared offset           & $c$              & [0.1937, 0.7418]         & --   \\
\midrule
DH  & Envelope drift rate     & $\varepsilon$    & $-0.05$ (fixed)          & s$^{-1}$ \\
DH  & Carrier frequency       & $f$              & [0.85, 1.10]             & Hz   \\
DH  & Initial phase           & $\varphi$        & [$-0.65$, 0.75]          & rad  \\
DH  & Linear trend coefficient & $a$             & [$-6\times10^{-5}$, $8\times10^{-5}$] & s$^{-1}$ \\
\bottomrule
\end{tabular}
\end{table}

\begin{table}[ht]
\centering
\caption{Fitting hyperparameters for real biosignal datasets.
Seg.\ = segment length; Ovl.\ = overlap.}
\label{tab:fitting_hyperparams_alt}
\small
\setlength{\tabcolsep}{4pt}
\begin{tabular}{@{}llcccccll@{}}
\toprule
\textbf{Dataset} & \textbf{Signal} & $f_s$\,\textbf{(Hz)} & \textbf{Seg.\,(s)} & \textbf{Ovl.} & \textbf{Epochs} & \textbf{lr} & \textbf{Loss} & \textbf{Model} \\
\midrule
PPG-DaLiA & PPG/BVP      & 64  & 5.0 & 50\% & 500 & 0.01  & MSE & DH      \\
MIT-BIH   & ECG (S-Q)    & 360 & 1.0 & 15\% & 500 & 0.001 & L1  & SPM  and DPM \\
CHB-MIT   & EEG (FP1-F7) & --  & 1.0 & --   & 100 & 0.01  & L1  &  DH and  spike   \\
\bottomrule
\end{tabular}
\end{table}

\subsection{Generation procedure and parameter uniqueness}
\label{app:generation}

For each signal family, 100 unique parameter realizations were generated (70 training, 10 validation, 20 test). All parameters were sampled independently from uniform distributions over the bounds in
Table~\ref{tab:param_bounds} using a fixed random seed (seed $= 42$) for reproducibility.

To guarantee that no two signals share the same parameter configuration, including across splits, each candidate parameter tuple was converted to a canonical string at six-decimal precision and hashed via MD5:
\begin{equation}
h = \mathrm{MD5}\bigl(\texttt{"}p_1\texttt{"}\_p_2\texttt{"}\_\cdots\texttt{"}\_p_k\texttt{"}\bigr)
\end{equation}
If the resulting hash matched any previously accepted configuration, the candidate was rejected and a new tuple was sampled. For SPM signals, the hash key comprised five values ($A$, $f$, $\beta$, $f_{\mathrm{mod}}$, $c$). For DPM signals, it was extended to nine values ($A_0$, $A_1$, $f_0$, $f_1$, $\beta_0$, $\beta_1$, $f_{\mathrm{mod},0}$, $f_{\mathrm{mod},1}$, $c$). For DH signals, it covered $f$, $\varphi$, and $a$ (with $\varepsilon$ fixed).

All generation parameters were embedded directly in each output filename for full traceability. Signals were sampled at $f_s = 10$~Hz for 300~s, yielding $T = 3{,}000$ samples per instance over an evenly spaced time vector $t \in [0, 300)$.

The 10~Hz sampling rate was chosen to capture modulation-envelope dynamics relevant to digital twin operation rather than high-frequency waveform morphology. Clinical digital twins for cardiac, respiratory, and neural monitoring typically operate on derived features (heart rate, respiratory rate, spectral power) that evolve on timescales of seconds to minutes, and the modulation-envelope structure captured at 10~Hz represents the dynamical layer at which forecasting fidelity most directly impacts clinical decision-making.

\section{Controlled Evaluation Paradigms}
\label{app:paradigms}

The main text describes five evaluation paradigms in terms of their clinical motivation and what each tests for digital twin deployment. This section provides the full technical specifications, including signal generation equations, parameter separation protocols, phase-continuity mechanisms, and dataset sizes needed for reproduction.

\subsection{Noise robustness}
\label{app:noise}

Models trained exclusively on clean signals are evaluated at seven SNR levels (Table~\ref{tab:snr}): a clean baseline (SNR~0) and six additive white Gaussian noise (AWGN) levels (SNR~1--6). Noise power is calibrated relative to the zero-mean signal power to prevent the DC offset from inflating the power estimate:
\begin{align}
P_{\mathrm{sig}} &= \frac{1}{T}\sum_{t=1}^{T}\bigl(x(t) - \bar{x}\bigr)^2 + \epsilon \\
\sigma_{\mathrm{noise}} &= \sqrt{\frac{P_{\mathrm{sig}}}{10^{\,\mathrm{SNR}_{\mathrm{dB}}/10}}} \\
\tilde{x}(t) &= x(t) + \mathcal{N}(0,\, \sigma_{\mathrm{noise}}^2)
\end{align}
where $\epsilon = 10^{-12}$ prevents division by zero. For DH signals, noise is added after min-max normalization (main text Eq.~3). Each SNR level uses
an independent random number generator seeded as $\mathrm{seed}_l = \mathrm{seed}_{\mathrm{base}} + l \times 1000$ (where $l$ is the SNR level index), ensuring independent noise realizations across
levels while maintaining within-level reproducibility. The same parameter configurations are used across all SNR levels and the clean condition, so that performance differences are attributable solely to noise. Each noise condition contains 70 training, 10 validation, and 20 test instances, mirroring the clean splits with identical parameter configurations.

\begin{table}[ht]
\centering
\caption{SNR levels for noise robustness evaluation. SNR~0 is the clean
baseline with no noise added. SNR~1 through SNR~6 represent progressively
increasing corruption.}
\label{tab:snr}
\begin{tabular}{@{}cccl@{}}
\toprule
\textbf{SNR level} & \textbf{SNR (dB)} & \textbf{Description} & \textbf{Clinical analogue} \\
\midrule
0 & $\infty$ (clean) & No noise added   & Ideal / ground truth \\
1 & 40              & Near-clean       & Controlled laboratory recording \\
2 & 30              & Very low noise   & Stationary bedside monitoring \\
3 & 20              & Low noise        & Ambulatory recording, minimal movement \\
4 & 10              & Moderate noise   & Ambulatory recording, routine activity \\
5 & 5               & High noise       & Wearable during exercise \\
6 & 1               & Severe noise     & Electrode displacement, heavy artifact \\
\bottomrule
\end{tabular}
\end{table}

\subsection{Frequency distribution shift}
\label{app:shift}

To quantify out-of-distribution generalization, the carrier frequency range used during training is treated as shift-0. Additional test sets are generated from frequency bands systematically displaced from the training range. Given a training frequency interval $(f_{\mathrm{low}},\, f_{\mathrm{high}})$ with width $w = f_{\mathrm{high}} - f_{\mathrm{low}}$:
\begin{itemize}
\item \textbf{Below training range}: the interval $[0,\, f_{\mathrm{low}})$ is divided into $n_{\mathrm{below}} = 2$ equally spaced sub-intervals, producing two buckets progressively further below the training distribution.
\item \textbf{Above training range}: $n_{\mathrm{above}} = 2$ additional intervals of width $w$ are placed by stepping upward from $f_{\mathrm{high}}$, i.e.\ bucket $k$ spans $[f_{\mathrm{low}} + kw,\; f_{\mathrm{high}} + kw]$ for $k = 1, 2$.
\end{itemize}
This yields five frequency buckets in total. The frequency ranges for all three signal families are reported in Table~\ref{tab:freq_shift}. Within each out-of-distribution bucket, 20 test signals are generated with the carrier frequency sampled uniformly from the bucket range and all non-frequency parameters ($A$, $\beta$, $f_{\mathrm{mod}}$, $c$) sampled from the original training bounds. This design isolates the effect of frequency shift from other parameter variations.

\begin{table}[ht]
\centering
\caption{Frequency distribution shift setup for the three signal families.
Shift $-2$ and $-1$ denote lower-frequency ranges, Shift~0 is the training
distribution, and Shift $+1$ and $+2$ denote higher-frequency ranges. All
values in Hz.}
\label{tab:freq_shift}
\begin{tabular}{@{}lccccc@{}}
\toprule
\textbf{Family} & \textbf{Shift $-2$} & \textbf{Shift $-1$} & \textbf{Shift 0 (Train)} & \textbf{Shift $+1$} & \textbf{Shift $+2$} \\
\midrule
Drift-Harmonic & [0.35, 0.60] & [0.60, 0.85] & [0.85, 1.10] & [1.10, 1.35] & [1.35, 1.60] \\
SPM-Harmonic   & [0.00, 0.34] & [0.34, 0.68] & [0.68, 1.41] & [1.41, 2.14] & [2.14, 2.88] \\
DPM-Harmonic   & [0.00, 0.34] & [0.34, 0.68] & [0.68, 1.41] & [1.41, 2.14] & [2.14, 2.88] \\
\bottomrule
\end{tabular}
\end{table}

\subsection{Single state transition}
\label{app:state-transition}

A deterministic frequency change-point is placed at a variable position $t^*$ within the signal, testing how quickly models adapt to abrupt state changes once they become observable. The signal alternates between two frequency states whose ranges are deliberately separated to ensure identifiability:
\begin{equation}
f_0 \sim \mathcal{U}\!\bigl(f_{\mathrm{low}} + 0.05\,\Delta f,\;\; f_{\mathrm{low}} + 0.25\,\Delta f\bigr),
\quad
f_1 \sim \mathcal{U}\!\bigl(f_{\mathrm{low}} + 0.55\,\Delta f,\;\; f_{\mathrm{low}} + 0.75\,\Delta f\bigr)
\label{eq:state-freq}
\end{equation}
where $\Delta f = f_{\mathrm{high}} - f_{\mathrm{low}}$ is the span of the
training frequency range. For SPM with $(f_{\mathrm{low}}, f_{\mathrm{high}}) = (0.6782, 1.4112)$, this yields  $f_0 \in [0.71, 0.94]$ Hz and $f_1 \in [1.07, 1.21]$ Hz. The same
separation logic is applied to modulation frequencies:
\begin{equation}
f_{\mathrm{mod},0} \sim \mathcal{U}\!\bigl(f_{\mathrm{mod}}^{\mathrm{low}} + 0.05\,\Delta f_{\mathrm{mod}},\;\; f_{\mathrm{mod}}^{\mathrm{low}} + 0.30\,\Delta f_{\mathrm{mod}}\bigr),
\;
f_{\mathrm{mod},1} \sim \mathcal{U}\!\bigl(f_{\mathrm{mod}}^{\mathrm{low}} + 0.55\,\Delta f_{\mathrm{mod}},\;\; f_{\mathrm{mod}}^{\mathrm{low}} + 0.90\,\Delta f_{\mathrm{mod}}\bigr)
\end{equation}
Small per-realization perturbations are added to amplitude and modulation
depth to further differentiate states: $\Delta A \sim \mathcal{U}(0.01, 0.03)$
and $\Delta\beta \sim \mathcal{U}(0.02, 0.04)$.

\paragraph{Phase continuity.}
To prevent artificial discontinuities at the change-point, which would introduce a nonphysiological transient that models could exploit rather than genuinely adapting to the new frequency regime, the instantaneous phase is computed recursively:
\begin{equation}
\theta(k) = \theta(k-1) + 2\pi\, f_{S(k-1)}\, \Delta t, \quad k = 1, \ldots, T-1
\label{eq:phase-recursion}
\end{equation}
where $S(k) \in \{0, 1\}$ is the state at sample $k$, $f_{S(k)}$ is the corresponding carrier frequency, and $\Delta t = 1/f_s$. This ensures that the phase accumulates smoothly across state boundaries, mimicking the continuous phase evolution of real physiological oscillators during state transitions.

\paragraph{Change-point placement.}
The change-point $t^*$ is sampled uniformly over $[0.25\,T,\; 0.75\,T]$, where $T$ is the total signal length. Given a history window of $H = 50$ samples and a prediction window of $P = 100$ samples, this placement allows the transition to fall either within the observed history (in-context: positions H-2 through H-40 samples before the forecast boundary) or within the unobserved future (no-context: positions F-2 through F-40 samples after the forecast boundary). Each split contains 600 training, 100 validation, and 200 test instances.

\subsection{Markov switching}
\label{app:markov}

Signals alternate stochastically between two frequency states governed by a symmetric two-state Markov chain, simulating the probabilistic state dynamics that characterize cardiac rhythm variability, sleep-stage cycling, and autonomic fluctuations. At each time step, the state transitions with probability $p$ or persists with probability $1 - p$:
\begin{equation}
S(k) =
\begin{cases}
1 - S(k-1) & \text{with probability } p \\
S(k-1)     & \text{with probability } 1 - p
\end{cases}
\end{equation}
with $S(0) = 0$. This is evaluated at five transition probabilities: $p \in \{0.10,\, 0.30,\, 0.50,\, 0.70,\, 0.90\}$, spanning infrequent switching ($p = 0.10$) to near-continuous alternation ($p = 0.90$). Per-state frequency and modulation parameter ranges follow the well-separated protocol (Eq.~\ref{eq:state-freq}). Modulation depth is additionally perturbed between states: $\beta_1 = \beta_0 + \Delta\beta$ with $\Delta\beta \sim \mathcal{U}(0.02, 0.04)$. Phase continuity is maintained via recursive accumulation (Eq.~\ref{eq:phase-recursion}). For each value of $p$, the dataset contains 70 training, 10 validation, and 20 test instances (100 per $p$; 500 total across all five conditions). Signals are sampled at $f_s = 10$~Hz for 300~s ($T = 3{,}000$ samples).

\section{Forecasting Model Architectures and Hyperparameters}
\label{app:models}

The main text benchmarks 11 forecasting models across four architectural families. This section provides architectural descriptions and full hyperparameter specifications for each family. All models share a unified
training protocol (AdamW optimizer, OneCycleLR scheduling, MSE loss, batch size 128, early stopping on validation loss) unless noted otherwise in the per-family tables. Hyperparameters across different signal families were largely consistent, with only minor adjustments to learning rate and weight decay. Each model was trained independently on each signal type and evaluation paradigm.

\subsection{Linear models}
\label{app:linear}

Three linear models serve as computational efficiency baselines. \textbf{Linear} applies a single linear layer mapping from the input sequence to the prediction horizon.\textbf{DLinear}~\cite{zeng2023transformers} decomposes the input into trend and seasonal components using a moving average kernel before applying separate linear transformations to each component. \textbf{FITS}~\cite{xu2023fits} operates in the complex frequency domain, interpolating low-frequency components to generate forecasts. All three models use Reversible Instance Normalization (RevIN) to handle distribution shift.

\begin{table}[ht]
\centering
\caption{Training hyperparameters for linear-family models.}
\label{tab:hp_linear}
\begin{tabular}{lccc}
\toprule
\textbf{Hyperparameter} & \textbf{Linear} & \textbf{DLinear} & \textbf{FITS} \\
\midrule
Training epochs       & 300        & 300        & 300        \\
Learning rate         & 0.0001     & 0.0001     & 0.0001     \\
Weight decay          & 0.001      & 0.001      & 0.001      \\
Batch size            & 128        & 128        & 128        \\
Patience              & 70         & 70         & 70         \\
LR schedule           & OneCycleLR & OneCycleLR & OneCycleLR \\
RevIN                 & Yes        & Yes        & Yes        \\
Decomposition kernel  & --         & 25         & --         \\
Cutoff frequency      & --         & --         & 15\,Hz     \\
\bottomrule
\end{tabular}
\end{table}

\subsection{MLP-based models}
\label{app:mlp}

Three MLP-based models introduce nonlinear transformations. \textbf{MLinear} is a two-layer multilayer perceptron with hidden dimensions [256, 512] and dropout regularization, serving as a nonlinear baseline. \textbf{N-BEATS}~\cite{oreshkin2019n} (Neural Basis Expansion Analysis for Time Series) introduces a deep architecture with backward and forward residual links organized into stacks of fully connected blocks. Each block produces both a backcast (reconstruction of the input) and a forecast, enabling interpretable decomposition. N-BEATS is the only model in our benchmark that produces an explicit backcast output. \textbf{FreMLP} (FreTS)~\cite{yi2023frequency} is a frequency-domain MLP that operates in two stages: domain conversion, which maps time-domain signals into complex-valued frequency components via FFT, and frequency learning, where redesigned MLPs jointly learn the real and imaginary parts of these components.

\begin{table}[ht]
\centering
\caption{Architectural and training hyperparameters for MLP-based models.}
\label{tab:hp_mlp}
\begin{tabular}{lccc}
\toprule
\textbf{Hyperparameter} & \textbf{MLinear} & \textbf{N-BEATS} & \textbf{FreMLP} \\
\midrule
Number of layers      & 2          & 5 per block & 2          \\
Number of blocks      & --         & 6           & --         \\
Hidden dimensions     & 256, 512   & 256, 512    & 256        \\
Embed size            & --         & --          & 128        \\
Activation            & GELU       & ReLU        & ReLU       \\
Block type            & --         & Generic     & --         \\
Backcast              & No         & Yes         & No         \\
MLP dropout           & 0.3        & 0.3         & 0.3        \\
Weight decay          & 0.0001     & 0.0001      & 0.0001     \\
Learning rate         & 0.0001     & 0.0001      & 0.0001     \\
Training epochs       & 300        & 300         & 300        \\
Patience              & 30         & 30          & 30         \\
Batch size            & 128        & 128         & 128        \\
LR schedule           & OneCycleLR & OneCycleLR  & OneCycleLR \\
\bottomrule
\end{tabular}
\end{table}

\subsection{CNN-based models}
\label{app:cnn}

Two CNN-based architectures process temporal structure through convolutional kernels with localized temporal processing windows. \textbf{ModernTCN}~\cite{luo2024moderntcn} is a temporal convolutional architecture featuring depthwise separable convolutions, residual connections, and structural reparameterization that fuses large and small kernels during inference to capture both short- and long-range temporal dependencies. \textbf{MICN}~\cite{wang2023micn} (Multi-scale Isometric Convolution Network) employs a multi-branch structure: local features are extracted through downsampling convolutions, while global dependencies are modeled using isometric convolutions with linear complexity in sequence length. We evaluate both \textbf{MICN\_Mean} and \textbf{MICN\_Regre}, which implement different strategies for handling trend-cyclical components: MICN\_Mean uses the mean of the decomposed trend for prediction, while MICN\_Regre applies a regression-based approach.

\begin{table}[ht]
\centering
\caption{Architectural and training hyperparameters for CNN-based models.}
\label{tab:hp_cnn}
\begin{tabular}{lcc}
\toprule
\textbf{Hyperparameter} & \textbf{ModernTCN} & \textbf{MICN} \\
\midrule
Number of blocks       & [2, 2, 2, 2]        & --              \\
Large kernel sizes     & [21, 19, 17, 13]    & --              \\
Small kernel sizes     & [3, 3, 3, 3]        & --              \\
Embedding dims         & [64, 128, 256, 512] & --              \\
FFN ratio              & 4                   & --              \\
Patch size / stride    & 20 / 10             & --              \\
Conv kernels           & --                  & [7, 17]         \\
Decomposition kernels  & --                  & [25, 49]        \\
Isometric kernels      & --                  & [17, 49]        \\
Hidden dimensions      & --                  & 256, 512        \\
Label length           & --                  & 50              \\
Trend prediction mode  & --                  & Regre / Mean    \\
Dropout                & 0.2                 & --              \\
Head dropout           & 0.1                 & --              \\
MLP dropout            & --                  & 0.3             \\
Learning rate          & 0.001               & 0.0001          \\
Weight decay           & 0.001               & 0.0001          \\
Training epochs        & 300                 & 300             \\
Patience               & 30                  & 30              \\
Batch size             & 128                 & 128             \\
LR schedule            & OneCycleLR          & OneCycleLR      \\
\bottomrule
\end{tabular}
\end{table}

\subsection{Transformer-based models}
\label{app:transformer}

Three transformer variants represent the range from global to local attention. A standard \textbf{Transformer} adapted for time series forecasting with an encoder-decoder architecture serves as the baseline. \textbf{Autoformer}~\cite{wu2021autoformer} replaces the self-attention mechanism with an auto-correlation module to capture long-range periodic dependencies and incorporates series decomposition within the architecture. \textbf{PatchTST}~\cite{nie2022time} divides the input time series into patches and applies transformer encoders over these patch-level representations, enabling localized attention patterns that preserve within-patch periodicity. For Transformer and Autoformer, half of the history window (25 time steps) was used as the label length to warm up the decoder. PatchTST uses 3 encoder layers (compared to 2 for the other transformer variants) and does not require a decoder.

\begin{table}[ht]
\centering
\caption{Architectural and training hyperparameters for transformer-based models.}
\label{tab:hp_transformer}
\begin{tabular}{lccc}
\toprule
\textbf{Hyperparameter} & \textbf{PatchTST} & \textbf{Autoformer} & \textbf{Transformer} \\
\midrule
Encoder layers    & 3          & 2          & 2          \\
Attention heads   & 8          & 8          & 8          \\
Embed dimension   & 256        & 256        & 256        \\
Feed-forward dim  & 256        & 256        & 256        \\
Dropout           & 0.2        & 0.2        & 0.2        \\
FC dropout        & 0.2        & 0.2        & 0.2        \\
Head dropout      & 0.2        & 0.2        & 0.2        \\
Patch length      & 15         & --         & --         \\
Stride            & 10         & --         & --         \\
Label length      & --         & 25         & 25         \\
Factor            & --         & --         & 3          \\
RevIN             & Yes        & Yes        & Yes        \\
Decomposition     & No         & No         & No         \\
Training epochs   & 300        & 300        & 300        \\
Patience          & 30         & 30         & 30         \\
Learning rate     & 0.0001     & 0.0001     & 0.0001     \\
Weight decay      & 0.0001     & 0.0001     & 0.0001     \\
Batch size        & 128        & 128        & 128        \\
LR schedule       & OneCycleLR & OneCycleLR & OneCycleLR \\
\bottomrule
\end{tabular}
\end{table}

\section{Fidelity Metric Computation}
\label{app:metrics}

The main text defines three fidelity metrics (amplitude error, frequency error, and phase error) and their clinical rationale. This section provides full algorithmic details, including preprocessing steps, edge case handling, reliability filtering, and masking procedures needed for reproduction. All metrics are computed per-sequence over the $H$-step forecast horizon (excluding the history window), and per-sequence values are aggregated via the median when comparing models.

\subsection{Amplitude error (MAE)}
\label{app:mae}

For each forecast sequence, amplitude error is computed as the mean absolute error between the predicted and true values over the prediction horizon:
\begin{equation}
\mathrm{MAE}_i = \frac{1}{H} \sum_{t=1}^{H} \bigl|\,\hat{y}_i(t) - y_i(t)\,\bigr|
\end{equation}
where $i$ indexes the sequence and $H = 100$ is the prediction length. No normalization or scaling is applied; all signals share the same amplitude range by construction.

\subsection{Frequency error}
\label{app:freq-error}

Frequency error quantifies the mismatch in dominant oscillation rate between the predicted and true signals. The estimation proceeds in four steps.

\paragraph{Step 1: DC removal.}
The signal mean is subtracted to eliminate the zero-frequency component:
$x(t) \leftarrow x(t) - \bar{x}$.

\paragraph{Step 2: Power spectrum.}
The one-sided power spectrum is computed via the real-valued FFT:
\begin{equation}
X(k) = \mathrm{FFT}_{\mathrm{real}}(x), \qquad
P(k) = |X(k)|^2, \qquad
k = 0, 1, \ldots, \lfloor N/2 \rfloor
\end{equation}
where $N$ is the FFT length (equal to the signal length; no zero-padding is applied for frequency estimation).

\paragraph{Step 3: Peak detection with parabolic refinement.}
The bin $k^*$ with maximum power (excluding the DC bin $k = 0$) is identified. For non-edge bins ($1 < k^* < \lfloor N/2 \rfloor$), the frequency estimate is refined using three-point parabolic interpolation:
\begin{equation}
\delta = \frac{P(k^*\!-\!1) - P(k^*\!+\!1)}{2\bigl(P(k^*\!-\!1) - 2\,P(k^*) + P(k^*\!+\!1)\bigr)}, \qquad
\hat{f} = (k^* + \delta)\,\frac{f_s}{N}
\end{equation}
where $f_s$ is the sampling rate. This provides sub-bin accuracy without increasing the FFT length. For edge bins ($k^* = 1$ or $k^* = \lfloor N/2 \rfloor$), the unrefined bin-center estimate $\hat{f} = k^* \cdot f_s / N$ is used.

\paragraph{Step 4: Reliability filtering.}
An estimate is marked as unreliable (NaN) and excluded from downstream analysis if either of the following conditions holds:
\begin{itemize}
\item \textbf{Low total power:} $\sum_k P(k) < 10^{-8}$ (effectively a flat
or constant signal).
\item \textbf{Diffuse spectrum:} $P(k^*) < 0.10 \cdot \sum_k P(k)$ (no
single frequency dominates; the signal lacks a clear periodicity).
\end{itemize}
Both thresholds are applied identically to the true and predicted signals. If both estimates are reliable, the per-sequence frequency error is $\Delta f_i = |\hat{f}_{\mathrm{pred},i} - \hat{f}_{\mathrm{true},i}|$. If either estimate is unreliable, $\Delta f_i$ is set to NaN and excluded. For model-level comparisons, only sequences where all models under comparison
have finite frequency error are retained (intersection-valid masking), ensuring that pairwise tests are conducted on identical sample sets.

\subsection{Phase error}
\label{app:phase-error}

Phase error quantifies temporal misalignment between the predicted and true signals. The computation involves constructing the analytic signal, extracting instantaneous phase, and averaging the phase difference over reliable regions.

\paragraph{Step 1: Preprocessing.}
The signal mean is subtracted: $x(t) \leftarrow x(t) - \bar{x}$.

\paragraph{Step 2: Analytic signal via frequency-domain Hilbert transform.}
The analytic signal $z(t)$ is constructed by:
\begin{enumerate}
\item Zero-pad the signal to length $N_{\mathrm{fft}} = 2N$ (pad factor $= 2$)
to reduce circular convolution edge effects.
\item Compute the FFT: $X(k) = \mathrm{FFT}(x_{\mathrm{padded}})$.
\item Construct the one-sided spectral mask:
\begin{equation}
H(k) = \begin{cases}
1 & k = 0 \\
2 & 1 \leq k < N_{\mathrm{fft}}/2 \\
1 & k = N_{\mathrm{fft}}/2 \\
0 & k > N_{\mathrm{fft}}/2
\end{cases}
\end{equation}
\item Inverse transform and crop to the original length:
$z(t) = \mathrm{IFFT}(X \cdot H)\big|_{t=0}^{N-1}$.
\end{enumerate}
The real part of $z(t)$ approximates the original signal, and the imaginary part is its Hilbert transform. This implementation is equivalent to \texttt{scipy.signal.hilbert} but provides explicit control over the padding factor. The pad factor of 2 was chosen to minimize edge artifacts; increasing it further did not meaningfully change the phase estimates on our signal families.

\paragraph{Step 3: Instantaneous phase extraction.}
The instantaneous phase is extracted and unwrapped for temporal continuity:
\begin{equation}
\varphi(t) = \mathrm{unwrap}\!\bigl(\arg(z(t))\bigr)
\end{equation}
Unwrapping removes $2\pi$ discontinuities, producing a monotonically evolving phase suitable for computing differences.

\paragraph{Step 4: Amplitude-based masking.}
Phase estimates are unreliable where the signal amplitude is low (e.g., near zero crossings of a modulated signal). A binary mask selects only time points where the true signal has sufficient amplitude:
\begin{equation}
\mathcal{M} = \bigl\{\,t : |z_{\mathrm{true}}(t)| > \alpha \cdot \mathrm{median}\!\bigl(|z_{\mathrm{true}}|\bigr) \,\bigr\},
\qquad \alpha = 0.2
\end{equation}
The threshold $\alpha = 0.2$ (20\% of the median instantaneous amplitude) was chosen to exclude low-amplitude regions while retaining the majority of the signal. If the median amplitude is zero or non-finite, or if no time points pass the mask, the sequence is marked NaN and excluded.

\paragraph{Step 5: Phase difference and wrapping.}
The phase difference is computed at each masked time point and wrapped to
$(-\pi, \pi]$:
\begin{equation}
\Delta\varphi(t) = \mathrm{wrap}_\pi\!\bigl(\varphi_{\mathrm{pred}}(t) - \varphi_{\mathrm{true}}(t)\bigr),
\qquad
\mathrm{wrap}_\pi(\theta) = \bigl((\theta + \pi) \bmod 2\pi\bigr) - \pi
\end{equation}

\paragraph{Step 6: Per-sequence phase error.}
The per-sequence metric is the mean absolute wrapped phase difference over the mask, converted to degrees:
\begin{equation}
\Delta\varphi_i = \frac{180}{\pi} \cdot \frac{1}{|\mathcal{M}|} \sum_{t \in \mathcal{M}} |\Delta\varphi(t)|
\end{equation}
As with frequency error, intersection-valid masking is applied for model-level comparisons.

\subsection{Residual variance analysis at comparable MAE}
\label{app:residual}
 
To quantify the extent to which MAE fails to resolve fidelity differences across architectures, we defined a \emph{comparable-MAE window} for each signal family as the central 60\% of median-MAE values across the 11 models (20th--80th percentile of MAE). Within this window, models are, by construction, effectively indistinguishable by conventional pointwise accuracy. For each architecture family (CNN, MLP, Linear-family,
Transformer) we then computed the mean phase error and the mean frequency error over the family's models whose MAE fell inside the window. The dissociation $\Delta$ reported in Table~\ref{tab:comparable_mae_dissociation} and marked by the red arrow in main-text Fig.~3 is the gap between the highest and lowest family mean in a given panel, quantifying the fidelity spread that MAE alone cannot detect.
 
\begin{table}[ht]
\centering
\caption{Family-level fidelity at comparable MAE. The comparable-MAE window is the central 60\% (20th--80th percentile) of MAE values across the 11 models. Entries are mean fidelity error per architecture family over models inside the window; per-family model counts in parentheses. $\Delta$ is the gap between the highest and lowest family mean in each row and corresponds to the red arrow in main-text Fig.~3. DH = Drift Harmonic; SPM = Single-Phase Modulation; DPM = Dual-Phase Modulation. Linear-family models account for the upper extreme of both metrics in all three signal families.}
\label{tab:comparable_mae_dissociation}
\footnotesize
\setlength{\tabcolsep}{5pt}
\renewcommand{\arraystretch}{1.1}
\begin{tabular}{@{}llcccccc@{}}
\toprule
\textbf{Metric} & \textbf{Family} & \textbf{MAE Window} &
\textbf{CNN} & \textbf{MLP} & \textbf{Linear} & \textbf{Transf.} &
$\boldsymbol{\Delta}$ \\
\midrule
\multirow{3}{*}{Phase error ($^\circ$)}
  & DH  & [0.003, 0.059] & 1.62 (3)   & --         & 29.48 (2)  & 2.14 (2)   & 27.87 \\
  & SPM & [0.011, 0.064] & 7.20 (2)   & 12.83 (1)  & 53.30 (2)  & 33.42 (2)  & 46.10 \\
  & DPM & [0.020, 0.081] & 9.68 (1)   & 58.87 (2)  & 62.28 (3)  & 9.80 (1)   & 52.59 \\
\midrule
\multirow{3}{*}{Frequency error (Hz)}
  & DH  & [0.003, 0.059] & 0.0002 (3) & --         & 0.0080 (2) & 0.0002 (2) & 0.0078 \\
  & SPM & [0.011, 0.064] & 0.0004 (2) & 0.0008 (1) & 0.0192 (2) & 0.0158 (2) & 0.0189 \\
  & DPM & [0.020, 0.081] & 0.0020 (1) & 0.0411 (2) & 0.0475 (3) & 0.0015 (1) & 0.0460 \\
\bottomrule
\end{tabular}
\end{table}

\section{Statistical Analysis}
\label{app:stats}

\subsection{Paired testing}
\label{app:paired}

For each fidelity metric, we computed paired differences between each model and the Linear baseline on a per-sequence basis. For clean, noise, and frequency shift paradigms, significance was assessed using the paired $t$-test:
\begin{equation}
t = \frac{\bar{d}}{s_d / \sqrt{n}}, \qquad p = 2\bigl(1 - \Phi(|t|)\bigr)
\end{equation}
where $\bar{d}$ is the mean paired difference, $s_d$ the standard deviation, $n$ the number of paired sequences, and $\Phi$ the standard normal CDF.
95\% confidence intervals were computed as $\bar{d} \pm 1.96 \cdot s_d / \sqrt{n}$.

For state-transition analyses, where error distributions are non-Gaussian due to the mixture of in-context and no-context conditions, we used the Wilcoxon signed-rank test with tie correction. Zero differences are removed, the absolute values of remaining differences are ranked (with average ranks assigned to ties), and the sum of ranks for positive differences $W^+$ is computed. The $z$-statistic with tie-corrected variance is:
\begin{equation}
z = \frac{W^+ - \mu}{\sqrt{\sigma^2}}, \qquad
\mu = \frac{n(n+1)}{4}, \qquad
\sigma^2 = \frac{n(n+1)(2n+1)}{24} - \sum_g \frac{t_g^3 - t_g}{48}
\end{equation}
where $t_g$ is the number of ties in group $g$. Two-sided $p$-value:
$p = 2(1 - \Phi(|z|))$.

\subsection{Multiple comparison correction}
\label{app:holm}

All $p$-values within each metric and paradigm were adjusted using the Holm
step-down procedure:
\begin{enumerate}
\item Sort the $m$ raw $p$-values in ascending order:
$p_{(1)} \leq p_{(2)} \leq \cdots \leq p_{(m)}$.
\item Multiply each by its rank-dependent factor:
$\tilde{p}_{(k)} = (m - k + 1) \cdot p_{(k)}$.
\item Enforce monotonicity:
$p^{\mathrm{Holm}}_{(k)} = \max\bigl(\tilde{p}_{(k)},\, p^{\mathrm{Holm}}_{(k-1)}\bigr)$,
capped at 1.0.
\end{enumerate}
The Holm procedure controls the family-wise error rate at $\alpha = 0.05$ while providing uniformly greater power than the classical Bonferroni
correction.

\subsection{Intersection-valid masking}
\label{app:masking}

For frequency and phase error, spectral reliability filtering (\S\ref{app:freq-error}, \S\ref{app:phase-error}) can produce NaN values for individual sequences. To ensure that all models are compared on exactly the same set of sequences, we apply intersection-valid masking: a sequence is included in the comparison only if all models under evaluation have a finite (non-NaN) value for that metric on that sequence. This prevents differences in sample composition from confounding pairwise comparisons.

\subsection{State-transition adaptation analysis}
\label{app:adaptation}

For the state-transition paradigm, sequences are grouped by distance tags indicating how far the transition lies from the forecast boundary. Tags take the form H-$XX$ (transition is $XX$ timesteps before the boundary, within the observable history) and F-$XX$ (transition is $XX$ timesteps after the boundary, in the unobserved future). The distance bins are $XX \in \{2, 4, 6, 10, 12, 15, 20, 30, 40\}$. Two additional tags, A and B, denote sequences with no transition in the evaluation window, serving as steady-state baselines for each frequency state.

Within each tag, a Wilcoxon signed-rank test is performed comparing each model to the Linear baseline, with Holm correction applied separately per tag. Adaptation speed is characterized by the first history tag at which a model's median phase error drops below a clinically meaningful threshold of $20^{\circ}$. For a digital twin receiving data at $10$~Hz, this provides a direct translation from timesteps to seconds of delayed state detection. For example, a model reaching $20^{\circ}$ at tag $H-6$ requires approximately 0.6 seconds of post-transition context, while one reaching $20^{\circ}$ at tag $H-40$ requires $4.0$ seconds.

\subsection{Markov fidelity assessment (HMM proxy)}
\label{app:hmm}

The HMM-based evaluation assesses whether forecasting models preserve the temporal structure of stochastic state switching. The procedure is as
follows:
\begin{enumerate}
        \item \textbf{Feature extraction.} Extract the dominant frequency from each
        window using a Welch periodogram (window $= 16$ samples, hop $= 8$ samples,
        $f_s = 10$~Hz).
        \item \textbf{Normalization.} Z-score normalize features across all sequences.
        \item \textbf{HMM fitting.} Fit a two-state Gaussian HMM on the true-history
        features, selecting the best model across eight random seeds
        (0, 1, 2, 3, 4, 5, 10, 20) by log-likelihood.
        \item \textbf{State canonicalization.} Relabel states so that state~0
        always has the lower emission mean.
        \item \textbf{Decoding.} Decode state sequences for both true-future and
        predicted-future using the fitted HMM.
        \item \textbf{Switching probability.} Compute the windowed switching
        probability (flip rate) for each decoded sequence.
        \item \textbf{Distributional fit.} Fit Gaussian distributions to the
        switching-probability distributions of true-history. 
            \item \textbf{Comparison.} Compare distributions via symmetric KL divergence:
    \begin{equation}
    \mathrm{KL_{sym}} = \mathrm{KL}(P\|Q) + \mathrm{KL}(Q\|P)
    \end{equation}
    where, for univariate Gaussians:
    \begin{equation}
    \mathrm{KL}(P\|Q) = \ln\!\frac{\sigma_Q}{\sigma_P}
    + \frac{\sigma_P^{2} + (\mu_P - \mu_Q)^{2}}{2\,\sigma_Q^{2}} - \frac{1}{2}
    \end{equation}
    \end{enumerate}
A model is classified as capturing the switching dynamics at a given transition probability if $\mathrm{KL_{sym}} < 0.05$. Sensitivity to this threshold is analyzed in \S\ref{app:markov-results}.

\subsection{Pareto frontier construction}
\label{app:pareto}

Models were scored across five evaluation paradigms: clean accuracy, noise robustness, shift robustness, state-transition adaptation, and Markov fidelity. Within each paradigm, scores were computed as the aggregate
improvement over the Linear baseline (averaged across signal families and fidelity metrics), then min-max normalized to $[0, 1]$ across models so that a score of 1.0 corresponds to the best-performing model on that paradigm and 0.0 to the worst.

A model $A$ is said to dominate model $B$ if $A$ scores greater than or equal to $B$ on all five paradigms and strictly greater on at least one. The Pareto frontier consists of all non-dominated models. Models not on the frontier are classified as dominated. The full normalized scores and Pareto classifications are reported in Table~\ref{tab:pareto_scores}.

\section{Extended Results}
\label{app:results}

This section reports extended results across signal families and evaluation paradigms. The structure mirrors the order in which the analyses are referenced from the main text: clean-condition fidelity profiles
(\S\ref{app:clean-results}), noise robustness across signal families (\S\ref{app:noise-results}), frequency-shift robustness across signal families (\S\ref{app:shift-results}), state-transition adaptation across fidelity dimensions (\S\ref{app:transition-results}), Markov switching with threshold sensitivity (\S\ref{app:markov-results}), and multi-paradigm performance profiles (\S\ref{app:pareto-results}).

\subsection{Three-dimensional fidelity profiles under clean conditions}
\label{app:clean-results}

\begin{figure}[ht]
    \centering
    \includegraphics[width=1.0\linewidth]{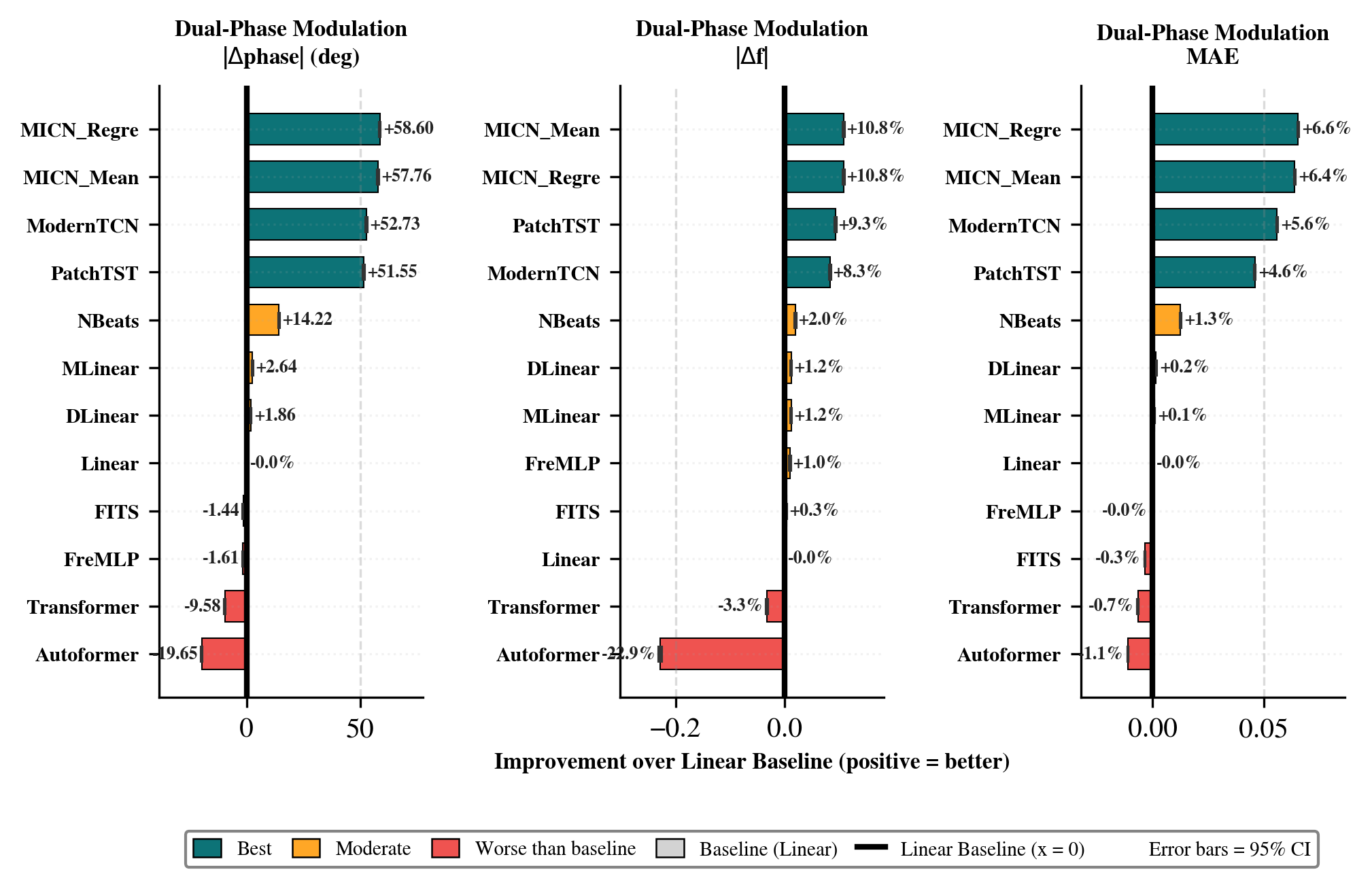}
    \caption{\textbf{Full fidelity profile for dual-phase modulation signals under clean conditions.} Phase improvement ($|\Delta\text{phase}|$,
    degrees; left), frequency improvement ($|\Delta f|$; center), and amplitude improvement (MAE; right) over the linear baseline for all 11
    architectures. MICN variants lead all three dimensions: MICN\_Mean and MICN\_Regre achieve $+10.8\%$ frequency improvement and $+6.4\%$ to
    $+6.6\%$ amplitude improvement alongside $>57^{\circ}$ phase gains.  PatchTST and ModernTCN follow closely in frequency ($+9.3\%$ and $+8.3\%$) and amplitude ($+4.6\%$ and $+5.6\%$). Transformer and
    Autoformer degrade across all three dimensions (Transformer:  $-9.58^{\circ}$ phase, $-3.3\%$ frequency, $-0.7\%$ amplitude;  Autoformer: $-19.65^{\circ}$ phase, $-22.9\%$ frequency, $-1.1\%$
    amplitude), confirming that their phase failure reflects a broader inability to preserve dynamical structure rather than an isolated timing deficit. Error bars represent 95\% confidence intervals. Color
    denotes performance tier: best (teal), moderate (orange), baseline (gray), worse than baseline (red).}
    \label{fig:supp_dual}
\end{figure}

\begin{figure}[ht]
    \centering
    \includegraphics[width=1.0\linewidth]{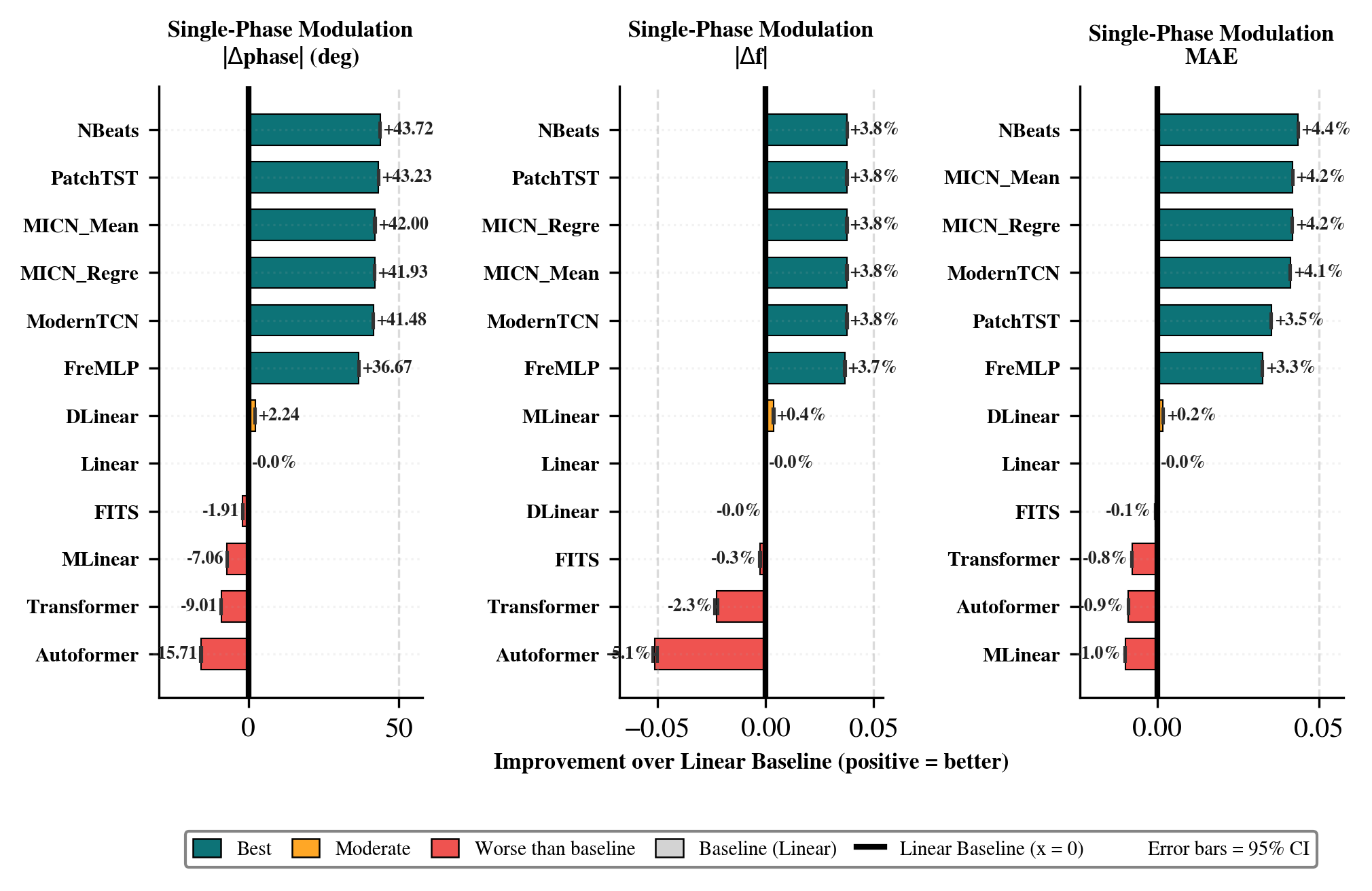}
    \caption{\textbf{Full fidelity profile for single-phase modulation  signals under clean conditions.} Phase improvement (left), frequency   improvement (center), and amplitude improvement (right) over the   linear baseline. The top tier compresses relative to dual-phase signals: NBeats ($+43.72^{\circ}$), PatchTST ($+43.23^{\circ}$),  MICN\_Mean ($+42.00^{\circ}$), MICN\_Regre ($+41.93^{\circ}$),
    ModernTCN ($+41.48^{\circ}$), and FreMLP ($+36.67^{\circ}$) all achieve substantial phase improvement, with frequency improvements tightly clustered between $+3.7\%$ and $+3.8\%$ and amplitude improvements  between $+3.3\%$ and $+4.4\%$. MLinear shows a dimension-specific  dissociation: moderate frequency improvement ($+0.4\%$) alongside phase degradation ($-7.06^{\circ}$), illustrating a failure mode that
    only separate fidelity diagnostics can detect. Transformer ($-9.01^{\circ}$ phase, $-2.3\%$ frequency, $-0.8\%$ amplitude) and Autoformer ($-15.71^{\circ}$ phase, $-9.1\%$ frequency, $-0.9\%$
    amplitude) again degrade across all dimensions. Error bars represent  95\% confidence intervals.}
    \label{fig:supp_single}
\end{figure}

\begin{figure}[ht]
    \centering
    \includegraphics[width=1.0\linewidth]{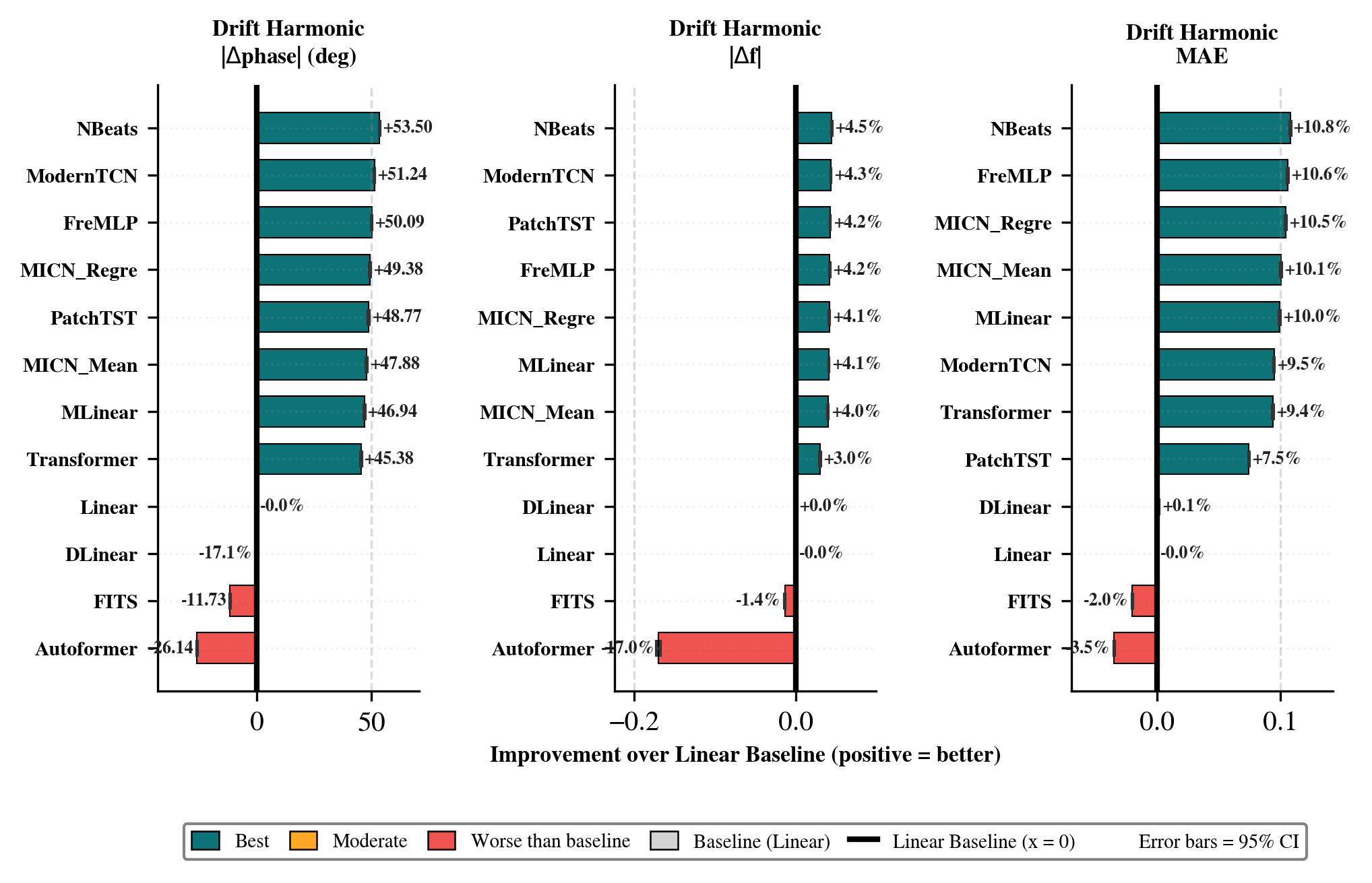}
    \caption{\textbf{Full fidelity profile for drift-harmonic signals  under clean conditions.} Phase improvement (left), frequency  improvement (center), and amplitude improvement (right) over the
    linear baseline. Nearly all architectures improve over baseline  across all three dimensions, with NBeats leading in phase  ($+53.50^{\circ}$), frequency ($+4.5\%$), and amplitude ($+10.8\%$).  Transformer achieves strong performance across all dimensions ($+45.38^{\circ}$ phase, $+3.0\%$ frequency, $+9.4\%$ amplitude), confirming that global attention preserves dynamical structure when spectral complexity is low. Even on these simplest signals, Autoformer remains the weakest architecture ($-26.14^{\circ}$ phase, $-17.0\%$  frequency, $-3.5\%$ amplitude) and FITS degrades across all dimensions  ($-11.73^{\circ}$ phase, $-1.4\%$ frequency, $-2.0\%$ amplitude), indicating fundamental mismatches to oscillatory signal preservation regardless of complexity. Error bars represent 95\% confidence intervals.}
    \label{fig:supp_drift}
\end{figure}

The main text reports phase fidelity as the primary diagnostic for evaluating architectural suitability, as phase preservation showed the largest dissociation from conventional MAE and the clearest architectural
separation. Here we present the full three-dimensional fidelity profiles, covering phase, frequency, and amplitude improvement over the linear baseline, for all 11 architectures across all three signal families
(Figs.~\ref{fig:supp_dual}--\ref{fig:supp_drift}). These results demonstrate that the architectural hierarchy observed for phase fidelity extends consistently to frequency and amplitude dimensions, with
informative differences in magnitude and ranking that validate the need for separate diagnostics.

On dual-phase modulation signals (Fig.~\ref{fig:supp_dual}), the most spectrally complex family, the top-performing architectures maintained their advantage across all three fidelity dimensions. MICN variants
achieved the highest frequency improvement ($+10.8\%$ for both MICN\_Mean and MICN\_Regre) and amplitude improvement ($+6.4\%$ and $+6.6\%$), consistent with their leading phase performance ($+57.76^{\circ}$ and
$+58.60^{\circ}$). PatchTST and ModernTCN followed in all three dimensions. Critically, Transformer and Autoformer degraded not only in phase but also in frequency ($-3.3\%$ and $-22.9\%$) and amplitude ($-0.7\%$ and $-1.1\%$), confirming that their failure is not limited to timing but extends to the full dynamical profile. Had evaluation relied on MAE alone, Autoformer's $-22.9\%$ frequency degradation would have been invisible.

On single-phase modulation signals (Fig.~\ref{fig:supp_single}), the architectural spread narrowed as the reduced spectral complexity placed lower demands on temporal processing window structure. Six architectures
achieved phase improvements between $+36^{\circ}$ and $+44^{\circ}$, with frequency improvements tightly clustered near $+3.8\%$ and amplitude improvements between $+3.3\%$ and $+4.4\%$. The narrower separation is
consistent with the signal-complexity interaction reported in the main text. A notable finding was the dimension-specific dissociation exhibited by MLinear: it achieved $+0.4\%$ frequency improvement and $+1.2\%$ phase improvement on this signal family, but degraded to $-7.06^{\circ}$ in phase on the same signals, demonstrating that a model can preserve one fidelity dimension while failing on another. This is precisely the failure mode that separate diagnostics are designed to detect and that conventional MAE evaluation would miss entirely.

On drift-harmonic signals (Fig.~\ref{fig:supp_drift}), the simplest family, nearly all architectures improved over baseline across all three dimensions. NBeats led in phase ($+53.50^{\circ}$), frequency ($+4.5\%$),
and amplitude ($+10.8\%$). The most informative finding was Transformer's strong three-dimensional performance ($+45.38^{\circ}$ phase, $+3.0\%$ frequency, $+9.4\%$ amplitude), which stands in sharp contrast to its degradation on dual-phase signals. This confirms that the Transformer's failure on complex signals is not a general architectural deficiency but a specific inability to handle multi-frequency modulation, a distinction that would be obscured by reporting performance on any single signal family alone.

\subsection{Noise robustness across signal families}
\label{app:noise-results}

\begin{figure}[ht]
    \centering
    \includegraphics[width=1.0\linewidth]{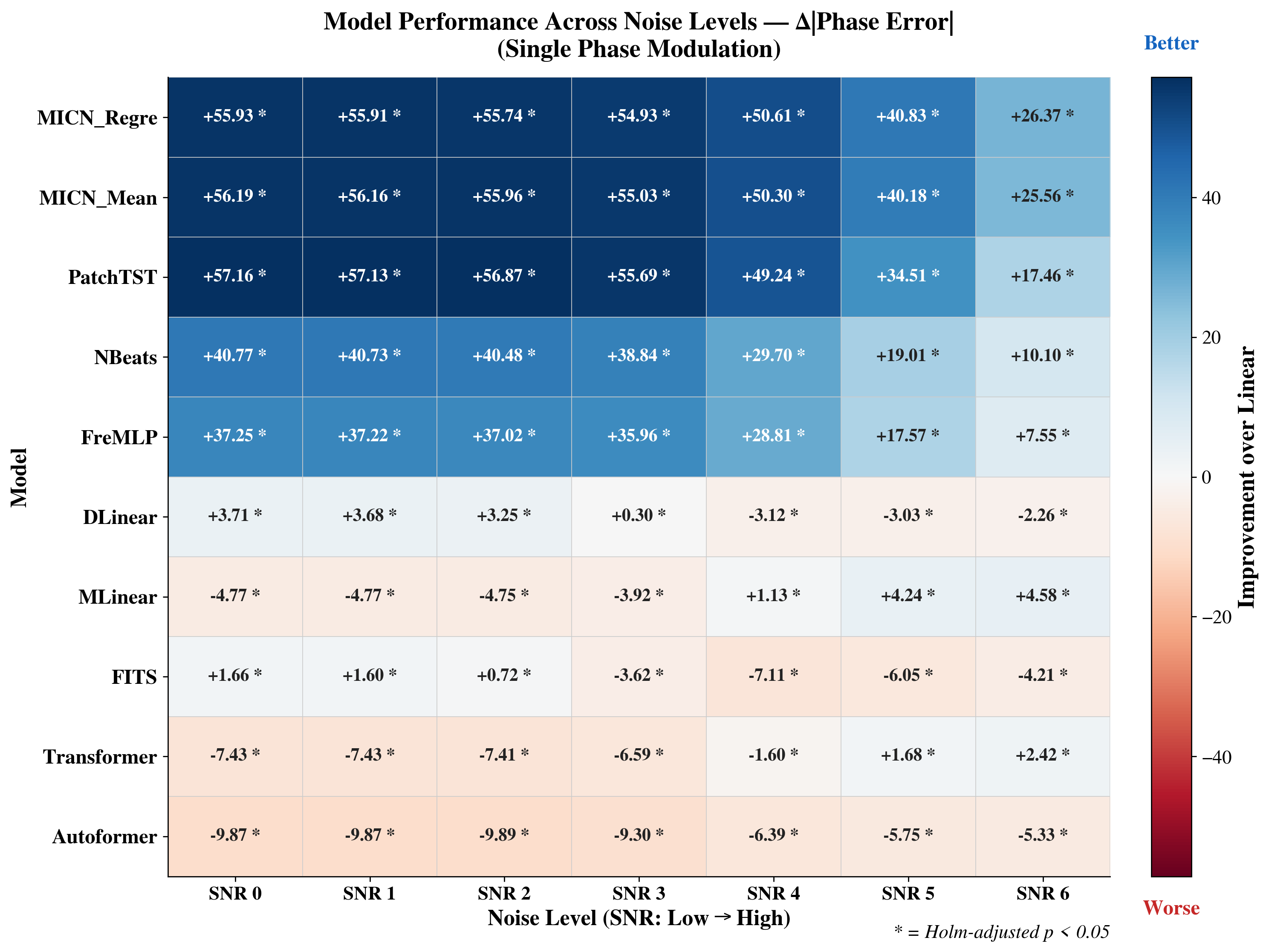}
    \caption{\textbf{Noise robustness for single-phase modulation signals.}
    Phase improvement ($\Delta|\text{phase}|$, degrees) over the linear baseline across seven noise levels (SNR~0 to SNR~6, where higher SNR number corresponds to more severe corruption). MICN\_Regre and
    MICN\_Mean lead across all noise levels, retaining $+26.37^{\circ}$ and $+25.56^{\circ}$ at SNR~6. PatchTST retains $+17.46^{\circ}$ at SNR~6, remaining in the top tier despite a steeper decline from
    $+57.16^{\circ}$ at SNR~0. Their relative ordering reverses compared with dual-phase signals (main-text Fig.~5), with MICN variants outperforming PatchTST under severe corruption, but both remain
    substantially ahead of all other architectures. MLinear exhibits a  crossover from negative improvement at low noise ($-4.77^{\circ}$ at SNR~0) to positive at high noise ($+4.58^{\circ}$ at SNR~6), and
    Transformer shows a similar reversal ($-7.43^{\circ}$ to $+2.42^{\circ}$), indicating that global mappings gain a relative advantage once noise obscures localized temporal structure, though  their absolute improvement remains far below PatchTST and MICN. Autoformer remains the worst performer across all noise levels ($-9.87^{\circ}$ to $-5.33^{\circ}$). All differences marked with $*$ are Holm-corrected $p < 0.05$.}
    \label{fig:supp_noise_single}
\end{figure}

\begin{figure}[ht]
    \centering
    \includegraphics[width=1.0\linewidth]{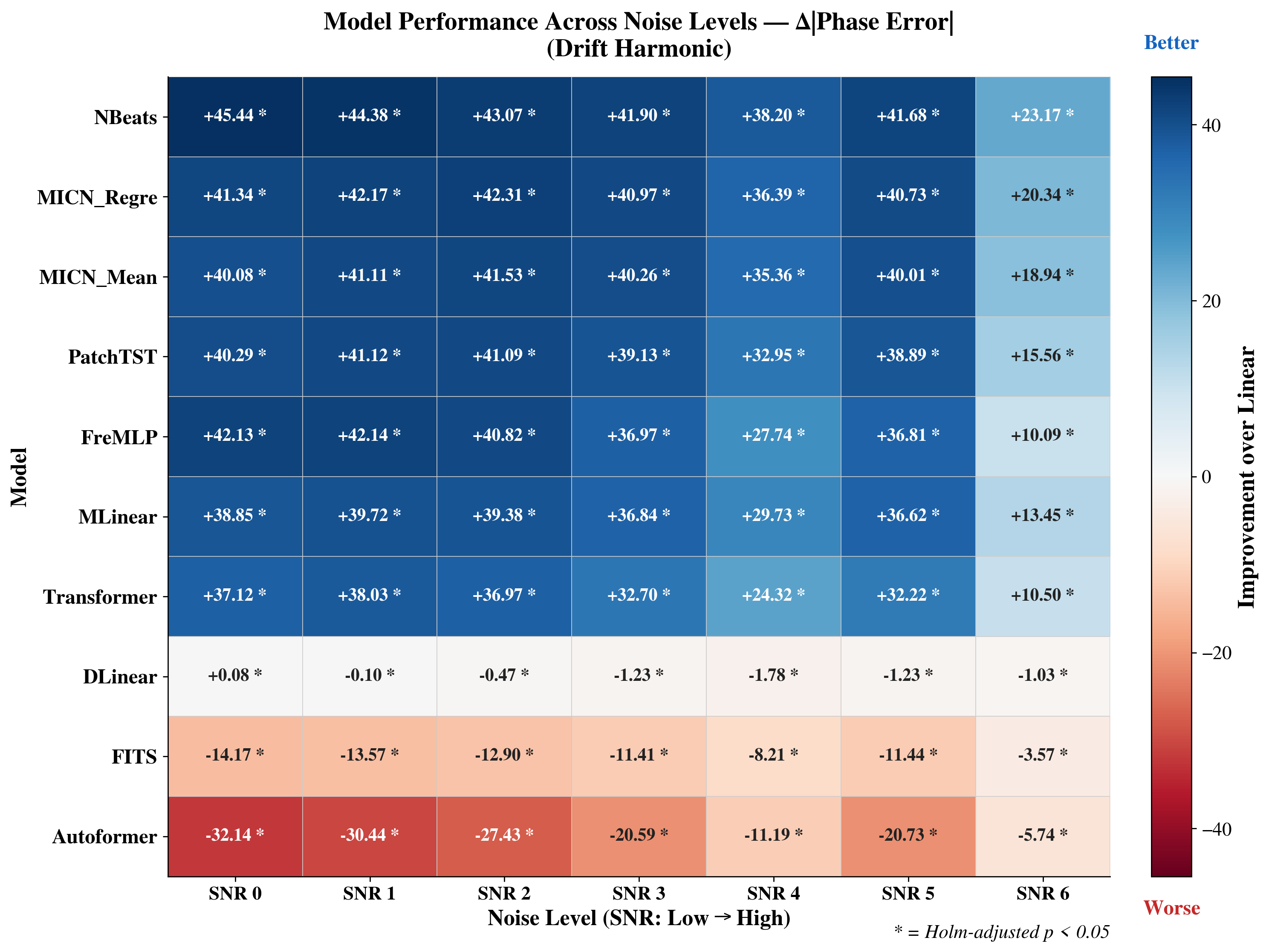}
    \caption{\textbf{Noise robustness for drift-harmonic signals.} Phase improvement over the linear baseline across seven noise levels.
    NBeats leads at SNR~6 ($+23.17^{\circ}$), with MICN\_Regre ($+20.34^{\circ}$), MICN\_Mean ($+18.94^{\circ}$), and PatchTST ($+15.56^{\circ}$) close behind. The architectural spread compresses
    relative to more complex signal families, consistent with reduced demands on temporal processing window structure when only a single slowly varying frequency is present. Nearly all nonlinear
    architectures retain positive improvement across the full noise range, including Transformer ($+37.12^{\circ}$ to $+10.50^{\circ}$) and  MLinear ($+38.85^{\circ}$ to $+13.45^{\circ}$). DLinear hovers near
    zero throughout. FITS ($-14.17^{\circ}$ to $-3.57^{\circ}$) and  Autoformer ($-32.14^{\circ}$ to $-5.74^{\circ}$) remain the only architectures with consistently negative improvement. All differences marked with $*$ are Holm-corrected $p < 0.05$.}
    \label{fig:supp_noise_drift}
\end{figure}

The main text reports noise robustness on dual-phase modulation signals,
the most spectrally complex family. Here we present the full noise
robustness profiles for single-phase modulation
(Fig.~\ref{fig:supp_noise_single}) and drift-harmonic
(Fig.~\ref{fig:supp_noise_drift}) signals, confirming that PatchTST and
MICN variants consistently occupy the top tier across signal families
despite shifts in their relative ordering.

On single-phase modulation signals (Fig.~\ref{fig:supp_noise_single}), MICN\_Regre and MICN\_Mean retained the strongest phase improvement across all noise levels, with MICN\_Regre holding $+55.93^{\circ}$ at SNR~0 and $+26.37^{\circ}$ at SNR~6. PatchTST followed closely ($+57.16^{\circ}$ at SNR~0, $+17.46^{\circ}$ at SNR~6), with both architectures substantially ahead of NBeats ($+40.77^{\circ}$ to $+10.10^{\circ}$) and FreMLP ($+37.25^{\circ}$ to $+7.55^{\circ}$). The relative ordering between PatchTST and MICN reversed compared with dual-phase signals: MICN\_Regre outperformed PatchTST at SNR~6 by $+8.91^{\circ}$ despite comparable clean-condition baselines, indicating that MICN's multi-scale convolution provides more stable retention under severe corruption on single-frequency signals. Two architectures exhibited informative crossovers: MLinear shifted from $-4.77^{\circ}$ at SNR~0 to $+4.58^{\circ}$ at SNR~6, and Transformer shifted from $-7.43^{\circ}$ to $+2.42^{\circ}$, suggesting that global mappings gain a relative advantage once noise obscures the localized temporal structure that patch-based and convolutional models exploit. However, even at their best noise level these architectures remained far below the PatchTST and MICN tier. Autoformer remained the worst performer ($-9.87^{\circ}$ to $-5.33^{\circ}$), and FITS degraded from marginally positive ($+1.66^{\circ}$) to consistently negative ($-4.21^{\circ}$).

On drift-harmonic signals (Fig.~\ref{fig:supp_noise_drift}), the gap between architectures compressed substantially as the simpler spectral structure reduced demands on temporal processing windows. NBeats showed
the best retention at SNR~6 ($+23.17^{\circ}$), followed by MICN\_Regre ($+20.34^{\circ}$), MICN\_Mean ($+18.94^{\circ}$), and PatchTST ($+15.56^{\circ}$). Notably, nearly all nonlinear architectures retained
positive improvement across the full noise range, including Transformer($+37.12^{\circ}$ to $+10.50^{\circ}$) and MLinear ($+38.85^{\circ}$ to $+13.45^{\circ}$), both of which had degraded on more complex signal
families. FreMLP exhibited the steepest decline among top models($+42.13^{\circ}$ to $+10.09^{\circ}$), consistent with noise amplifying spectral leakage in frequency-domain representations. Even on these
simplest signals, Autoformer remained the worst performer ($-32.14^{\circ}$ to $-5.74^{\circ}$) and FITS showed consistent degradation ($-14.17^{\circ}$ to $-3.57^{\circ}$).

Taken together, the cross-signal noise results reinforce two findings from the main text. First, PatchTST and MICN variants consistently occupied the top tier across all three signal families, with their relative
ordering shifting (PatchTST leading on dual-phase, MICN leading on single-phase, NBeats leading on drift-harmonic at the highest noise) but both remaining substantially ahead of linear-family and global attention
architectures. Second, the noise robustness hierarchy interacted with signal complexity: on spectrally complex signals, only PatchTST and MICN maintained strong improvement under noise, whereas on simpler signals a
broader range of architectures retained fidelity.

\subsection{Frequency shift robustness across signal families}
\label{app:shift-results}

\begin{figure}[ht]
    \centering
    \includegraphics[width=1.0\linewidth]{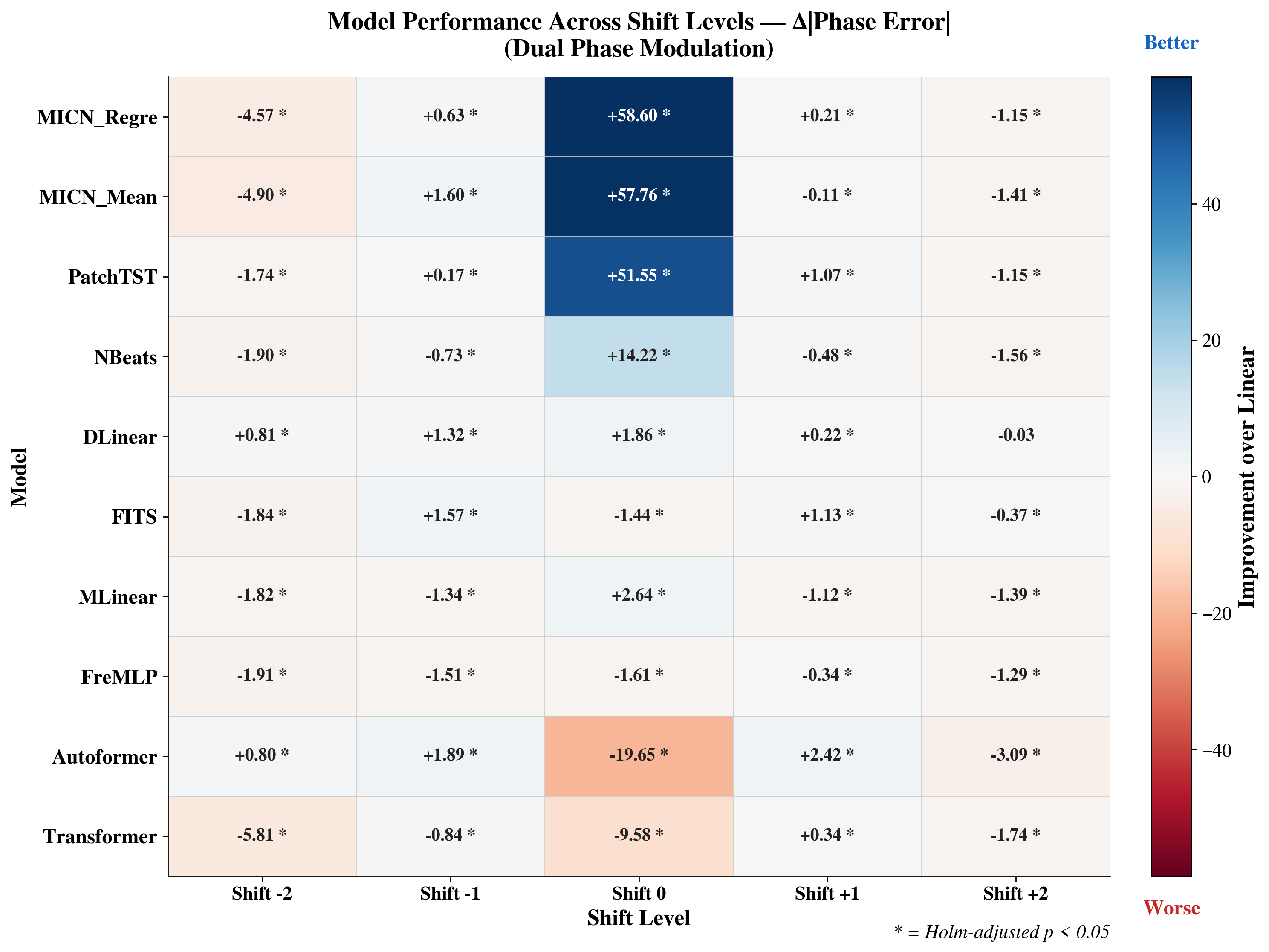}
    \caption{\textbf{Frequency shift robustness for dual-phase modulation
    signals.} Phase improvement ($\Delta|\text{phase}|$, degrees) over the linear baseline across five shift levels (Shift~$-2$ to Shift~$+2$). At Shift~0, MICN\_Regre ($+58.60^{\circ}$), MICN\_Mean
    ($+57.76^{\circ}$), and PatchTST ($+51.55^{\circ}$) achieve strong improvement, consistent with clean-condition results. Under any frequency shift, these advantages collapse: MICN\_Regre falls to
    $-4.57^{\circ}$ at Shift~$-2$ and $-1.15^{\circ}$ at Shift~$+2$, and PatchTST falls to $-1.74^{\circ}$ and $-1.15^{\circ}$. The collapse is roughly symmetric and more severe than on single-phase signals
    (main-text Fig.~6b), indicating that multi-frequency modulation amplifies sensitivity to distributional mismatch. NBeats ($+14.22^{\circ}$ at Shift~0) also collapses under shift ($-1.90^{\circ}$ at Shift~$-2$, $-1.56^{\circ}$ at Shift~$+2$). DLinear shows modest but stable improvement across shifts  ($+0.81^{\circ}$ to $+1.86^{\circ}$). Transformer ($-9.58^{\circ}$) and Autoformer ($-19.65^{\circ}$) degrade even at Shift~0. All differences marked with $*$ are Holm-corrected $p < 0.05$.}
    \label{fig:supp_shift_dual}
\end{figure}

\begin{figure}[ht]
    \centering
    \includegraphics[width=1.0\linewidth]{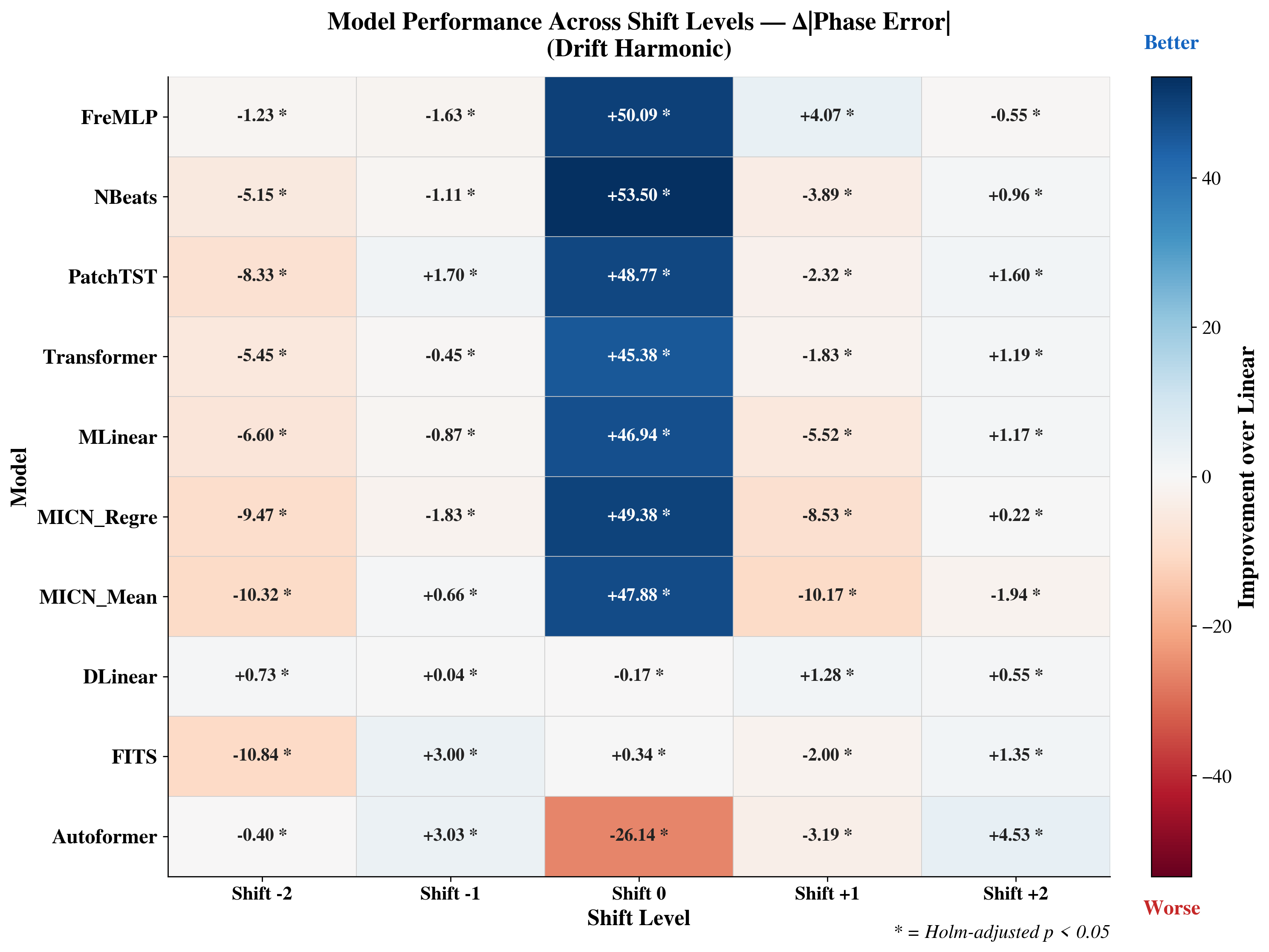}
    \caption{\textbf{Frequency shift robustness for drift-harmonic
    signals.} Phase improvement over the linear baseline across five shift levels. At Shift~0, NBeats ($+53.50^{\circ}$), FreMLP($+50.09^{\circ}$), MICN\_Regre ($+49.38^{\circ}$), PatchTST($+48.77^{\circ}$), and MICN\_Mean ($+47.88^{\circ}$) all achieve strong improvement. A directional asymmetry emerges under shift:  negative shifts produce larger degradation than positive shifts  across most architectures. MICN\_Mean falls to $-10.32^{\circ}$ at Shift~$-2$ but only $-1.94^{\circ}$ at Shift~$+2$. MICN\_Regre falls to $-9.47^{\circ}$ at Shift~$-2$ but retains $+0.22^{\circ}$ at Shift~$+2$. PatchTST falls to $-8.33^{\circ}$ at Shift~$-2$ but retains $+1.60^{\circ}$ at Shift~$+2$. This asymmetry suggests that downward frequency shifts, such as transitions from active to resting states, pose a greater challenge than upward shifts for models trained on higher-frequency bands. Autoformer ($-26.14^{\circ}$ atShift~0) remains the worst performer. All differences marked with $*$ are Holm-corrected $p < 0.05$.}
    \label{fig:supp_shift_drift}
\end{figure}

The main text reports frequency shift robustness on single-phase modulation
signals. Here we present the full shift robustness profiles for dual-phase
modulation (Fig.~\ref{fig:supp_shift_dual}) and drift-harmonic
(Fig.~\ref{fig:supp_shift_drift}) signals, revealing that the severity of
frequency shift degradation scales with signal complexity and that a
directional asymmetry emerges on simpler signals.

On dual-phase modulation signals (Fig.~\ref{fig:supp_shift_dual}), the collapse under frequency shift was substantially more severe than on single-phase signals. At Shift~0, the architectural hierarchy matched the
clean-condition results: MICN\_Regre ($+58.60^{\circ}$), MICN\_Mean ($+57.76^{\circ}$), and PatchTST ($+51.55^{\circ}$) led, with NBeats at $+14.22^{\circ}$. Under any degree of shift, however, these advantages effectively disappeared. MICN\_Regre fell to $-4.57^{\circ}$ at Shift~$-2$ and $-1.15^{\circ}$ at Shift~$+2$, PatchTST fell to $-1.74^{\circ}$ and $-1.15^{\circ}$, and NBeats fell to $-1.90^{\circ}$ and $-1.56^{\circ}$. The collapse was roughly symmetric across positive and negative shifts, and no architecture maintained more than $+2^{\circ}$ improvement at anyshifted condition. This contrasts with single-phase signals where PatchTST retained $+8.85^{\circ}$ at Shift~$-2$, indicating that multi-frequency modulation amplifies sensitivity to distributional mismatch. DLinear showed modest but stable improvement across all shifts ($+0.81^{\circ}$ to $+1.86^{\circ}$), consistent with its shift-invariant linear mapping. Transformer ($-9.58^{\circ}$ at Shift~0, $-5.81^{\circ}$ at Shift~$-2$) and Autoformer ($-19.65^{\circ}$ at Shift~0) degraded even without shift, confirming their unsuitability for nonstationary monitoring regardless of shift severity.

On drift-harmonic signals (Fig.~\ref{fig:supp_shift_drift}), a directional asymmetry emerged that was not apparent on the other signal families. At Shift~0, nearly all architectures achieved strong improvement, with NBeats ($+53.50^{\circ}$), FreMLP ($+50.09^{\circ}$), MICN\_Regre ($+49.38^{\circ}$), PatchTST ($+48.77^{\circ}$), and MICN\_Mean ($+47.88^{\circ}$) all above $+47^{\circ}$. Under negative shifts (lower
frequencies at test time), degradation was substantially greater than under positive shifts. MICN\_Mean fell to $-10.32^{\circ}$ at Shift~$-2$ but only $-1.94^{\circ}$ at Shift~$+2$. MICN\_Regre showed a comparable pattern ($-9.47^{\circ}$ vs.\ $+0.22^{\circ}$), and PatchTST followed the same trend ($-8.33^{\circ}$ vs.\ $+1.60^{\circ}$). This directional asymmetry suggests that downward frequency shifts, such as transitions from active to resting physiological states, pose a greater challenge than upward shifts for models trained on higher-frequency bands. Notably, Transformer ($+45.38^{\circ}$ at Shift~0) also showed this asymmetry ($-5.45^{\circ}$ at Shift~$-2$ vs.\ $+1.19^{\circ}$ at Shift~$+2$), indicating that the effect is not architecture-specific but reflects a general property of how models extrapolate beyond their training frequency range. Autoformer remained the worst performer at Shift~0 ($-26.14^{\circ}$) but paradoxically showed less degradation under shift ($-0.40^{\circ}$ at Shift~$-2$, $+4.53^{\circ}$ at Shift~$+2$), likely because its already-poor baseline left less room to degrade.

Taken together, the cross-signal shift results reinforce two findings from the main text. First, PatchTST and MICN variants maintained the strongest improvement at Shift~0 across all signal families, confirming their suitability as the default architectural choice for digital twins operating within their training distribution. Second, no architecture maintained substantial improvement beyond moderate shifts ($\pm 1$) on any signal family, and the severity of collapse increased with signal complexity. The directional asymmetry on drift-harmonic signals adds a new consideration: digital twins monitoring signals where downward frequency shifts are clinically relevant, such as transitions from active wakefulness to drowsiness in EEG, may require more frequent model updating than those tracking upward shifts.

\subsection{State transition adaptation across fidelity dimensions}
\label{app:transition-results}

\begin{figure}[ht]
    \centering
    \includegraphics[width=1.0\linewidth]{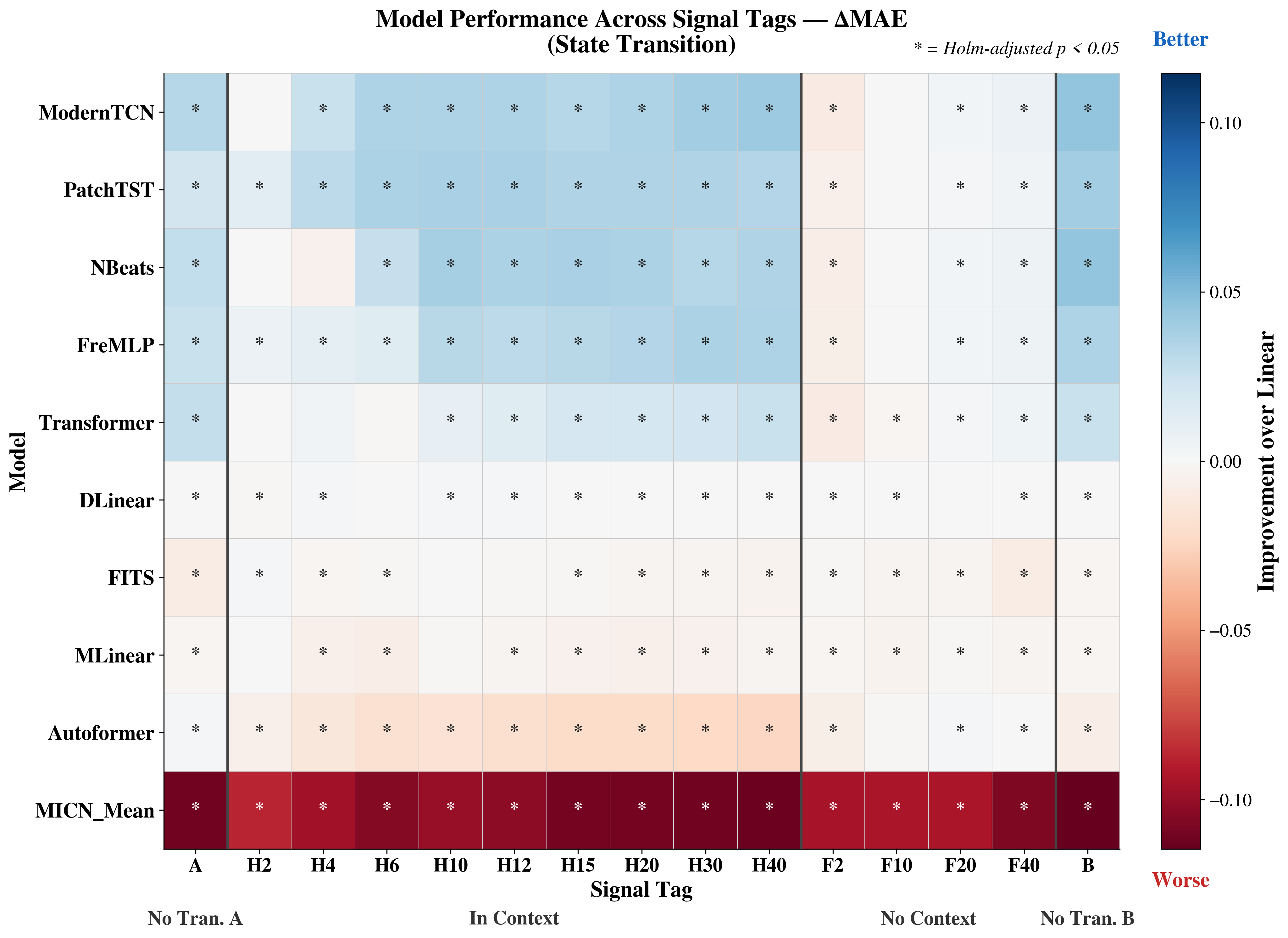}
    \caption{\textbf{Amplitude fidelity (MAE) across state transition tags.} Improvement in MAE over the linear baseline for all 11 architectures across signal tags: no-transition baselines (A, B),
    in-context transitions (H2 to H40), and no-context transitions (F2 to F40). ModernTCN and PatchTST show the strongest and most consistent amplitude improvement across in-context tags, with improvement
    intensifying as more post-transition context becomes available. FreMLP shows consistent moderate improvement across all in-context tags. NBeats shows weaker amplitude adaptation than its phase
    adaptation, with modest improvement emerging only after H6. MICN\_Mean shows consistent amplitude degradation across all tags (dark red), contrasting with its strong clean-condition amplitude performance and indicating that state transitions specifically  disrupt its amplitude preservation. Autoformer shows degradation concentrated at in-context tags with moderate context (H6 to H40). Linear-family models (DLinear, MLinear, FITS) show mixed patterns with generally weak improvement. All differences marked with $*$ are  Holm-corrected $p < 0.05$.}
    \label{fig:supp_transition_mae}
\end{figure}

\begin{figure}[ht]
    \centering
    \includegraphics[width=1.0\linewidth]{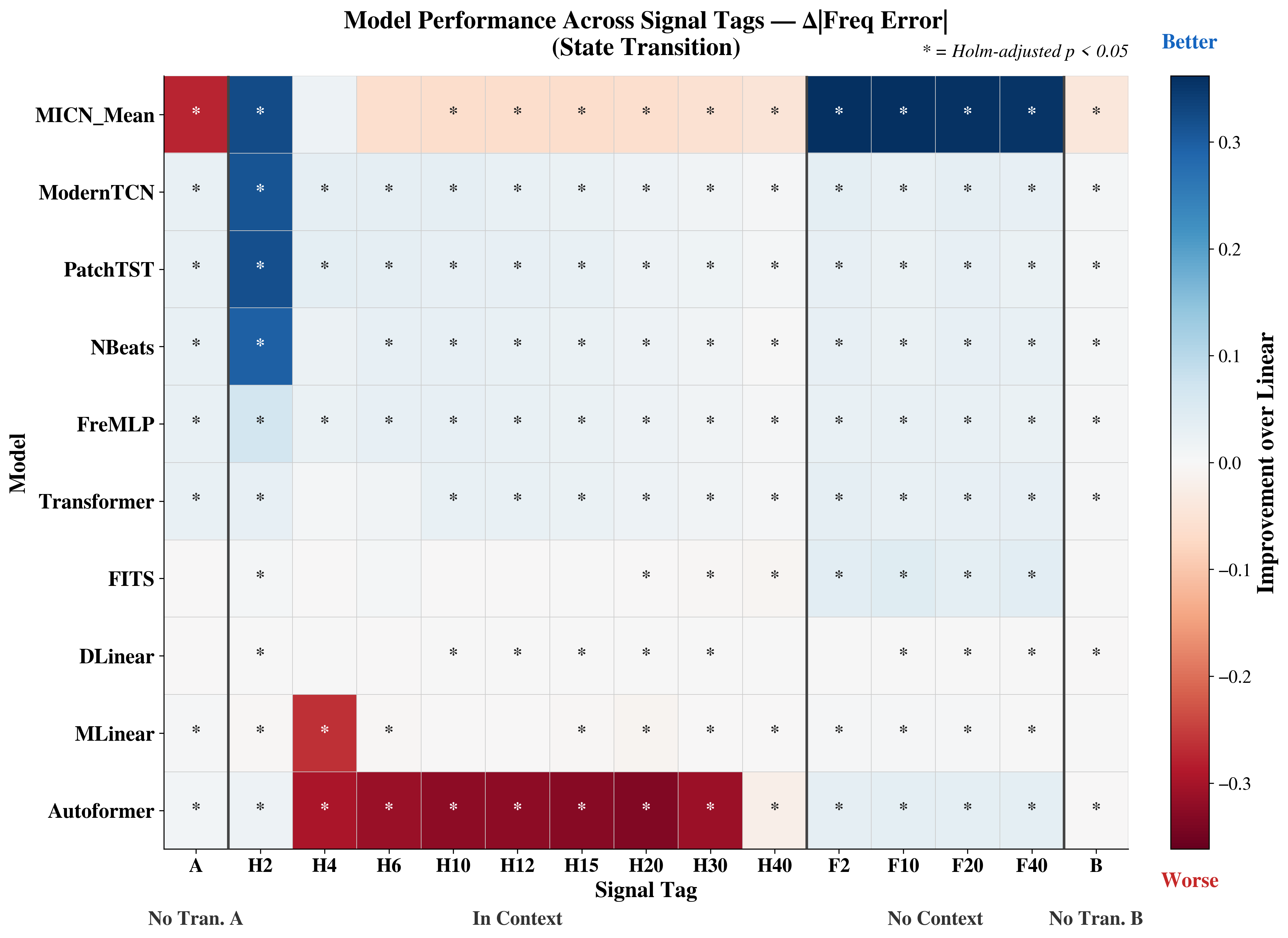}
    \caption{\textbf{Frequency fidelity across state transition tags.} Improvement in frequency error ($\Delta|f|$) over the linear baseline across signal tags. MICN\_Mean achieves the strongest frequency
    improvement across in-context tags (H10 to H40), with deep blue  indicating substantial improvement over Linear. ModernTCN and  PatchTST show strong frequency improvement at early tags (H2), consistent with their rapid phase adaptation. NBeats shows strong  frequency improvement at H2 that diminishes at intermediate tags before recovering, suggesting non-monotonic frequency adaptation. Autoformer shows the largest frequency degradation across in-context tags (dark red, H4 to H40), and MLinear shows a sharp frequency degradation at H4. Transformer shows weak frequency adaptation throughout. In the no-context condition (F2 to F40), MICN\_Mean  retains substantial frequency improvement while most other  architectures show moderate improvement, indicating that MICN's  frequency preservation is more robust to unseen transitions than its amplitude preservation. All differences marked with $*$ are  Holm-corrected $p < 0.05$.}
    \label{fig:supp_transition_freq}
\end{figure}

The main text reports state transition adaptation speed using phase fidelity as the primary diagnostic, as phase preservation showed the clearest differentiation between architectural adaptation rates. Here we present the full tag-wise profiles for amplitude fidelity (MAE; Fig.~\ref{fig:supp_transition_mae}) and frequency fidelity (Fig.~\ref{fig:supp_transition_freq}) across all signal tags, confirming that the adaptation speed hierarchy identified through phase error extends consistently to the other two fidelity dimensions, with additional dimension-specific patterns.

For amplitude fidelity (Fig.~\ref{fig:supp_transition_mae}), the adaptation speed hierarchy broadly matched the phase results but with notable differences. ModernTCN and PatchTST showed the strongest and most
consistent amplitude improvement across in-context tags, with improvement intensifying as more post-transition context became available, consistent with their rapid phase adaptation reported in the main text. FreMLP maintained consistent moderate improvement across all in-context tags. NBeats, which showed strong phase adaptation, exhibited weaker amplitude adaptation, with modest improvement emerging only after H6, suggesting that its basis expansion architecture recovers oscillatory timing before it recovers oscillation magnitude. The most striking finding was MICN\_Mean's consistent amplitude degradation across all tags including no-transition baselines, shown as uniformly dark red in the heatmap. This contrasts with MICN\_Mean's strong clean-condition amplitude performance (Fig.~\ref{fig:supp_dual}, $+6.4\%$), indicating that the introduction of state transitions specifically disrupts MICN's amplitude preservation even when the transition itself has not yet occurred, possibly through sensitivity of its multi-scale decomposition to nonstationarity in the training distribution. When transitions occurred in the unobserved future (F2 to F40), amplitude improvement generally weakened across all architectures, paralleling the phase results.

For frequency fidelity (Fig.~\ref{fig:supp_transition_freq}), a different pattern emerged that reveals dimension-specific adaptation dynamics. MICN\_Mean achieved the strongest frequency improvement across in-context tags (H10 to H40), contrasting sharply with its amplitude degradation on the same tags. This dimension-specific dissociation, strong frequency preservation alongside amplitude degradation, is precisely the kind of failure mode that separate fidelity diagnostics are designed to detect: had evaluation relied on either metric alone, MICN\_Mean would have appeared either excellent (frequency) or poor (amplitude) for state transition adaptation, when in reality it exhibits a complex, dimension-dependent profile. ModernTCN and PatchTST showed strong frequency improvement at early tags (H2), consistent with their rapid
phase adaptation and confirming that localized temporal processing windows enable fast recovery across multiple fidelity dimensions simultaneously. NBeats showed an unexpected non-monotonic pattern: strong
frequency improvement at H2 that diminished at intermediate tags before recovering at later tags, suggesting that its basis expansion architecture undergoes a transient frequency adjustment period during adaptation. Autoformer showed the largest frequency degradation across in-context tags, with dark red cells from H4 to H40, confirming its consistent unsuitability across all fidelity dimensions. MLinear exhibited a sharp frequency degradation specifically at H4, indicating a narrow window of instability during early adaptation.

These cross-dimensional profiles reinforce the main text finding that localized temporal processing architectures (PatchTST, ModernTCN) adapt fastest, while adding the nuance that adaptation speed can vary across
fidelity dimensions within the same architecture. MICN\_Mean's contrasting performance across frequency (strong) and amplitude (weak) during state transitions illustrates that architecture selection for
digital twins monitoring state-change-prone signals should consider which fidelity dimension is most clinically relevant: if frequency tracking is paramount (as in seizure detection where frequency shifts precede generalization), MICN may be suitable despite its amplitude vulnerability, whereas if amplitude preservation is critical (as in hemodynamic monitoring), PatchTST or ModernTCN offer more balanced adaptation.

\subsection{Stochastic switching: full results and threshold sensitivity}
\label{app:markov-results}

\begin{figure}[ht]
\centering
\includegraphics[width=0.48\textwidth]{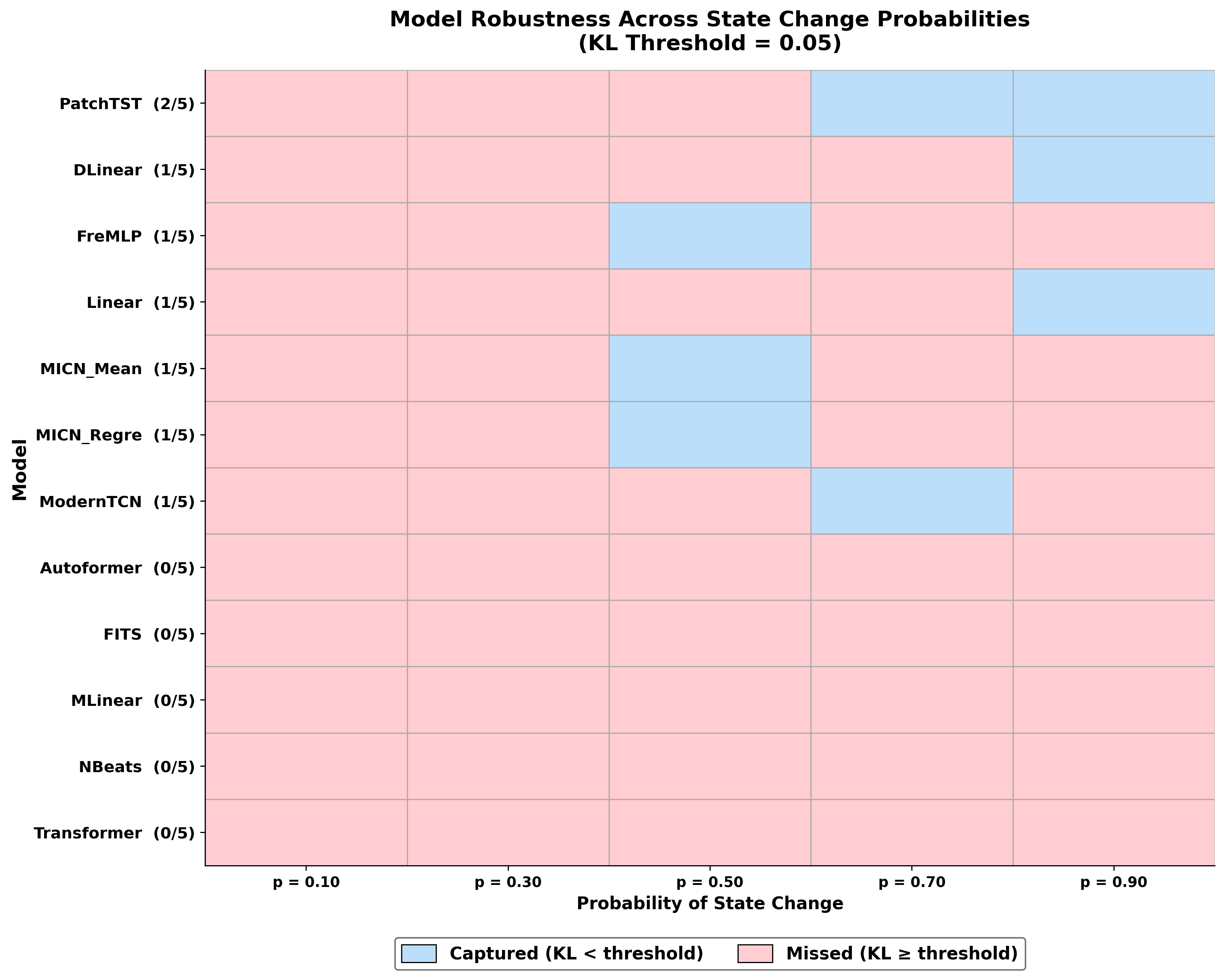}%
\hfill
\includegraphics[width=0.48\textwidth]{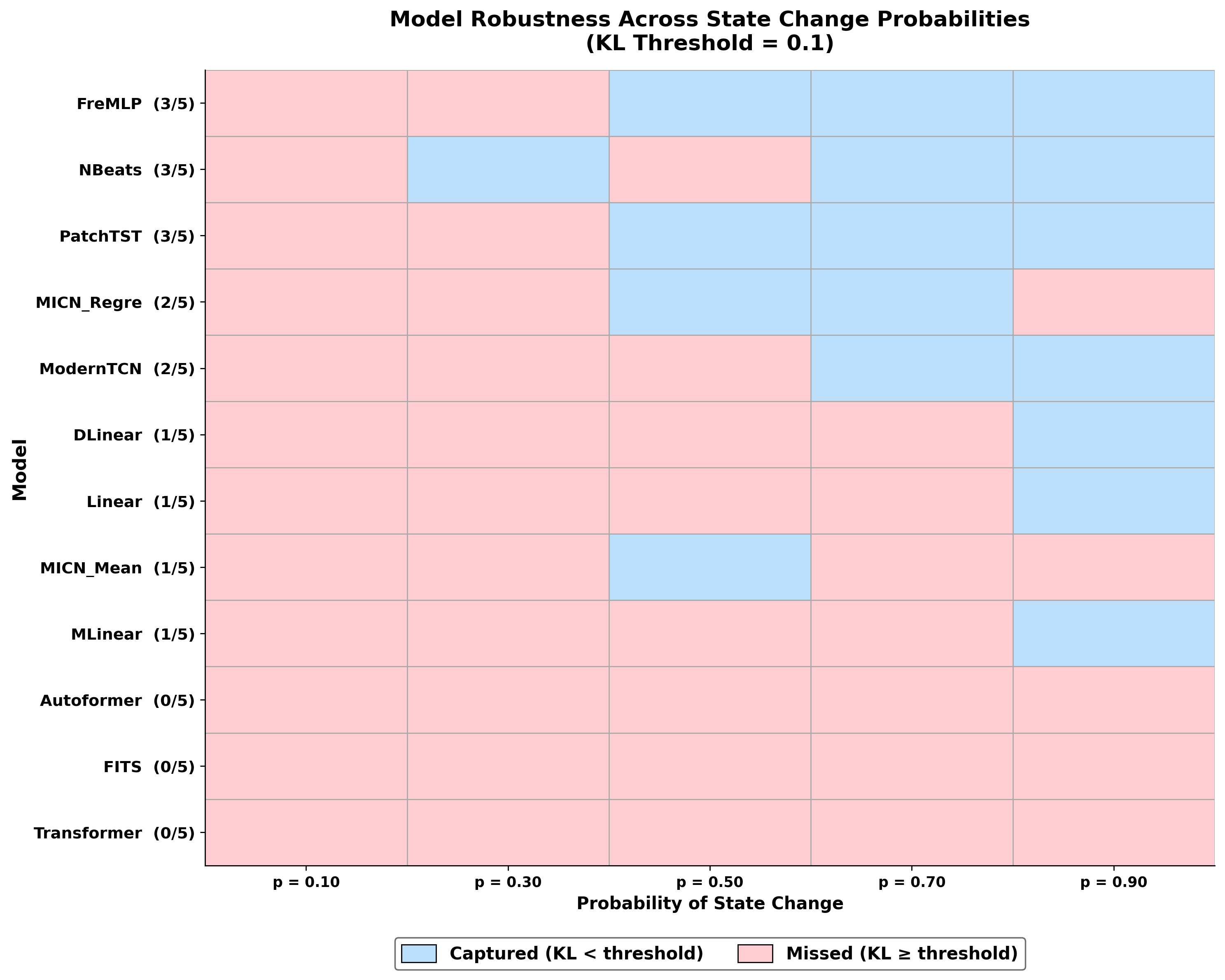}\\[4pt]
\includegraphics[width=0.48\textwidth]{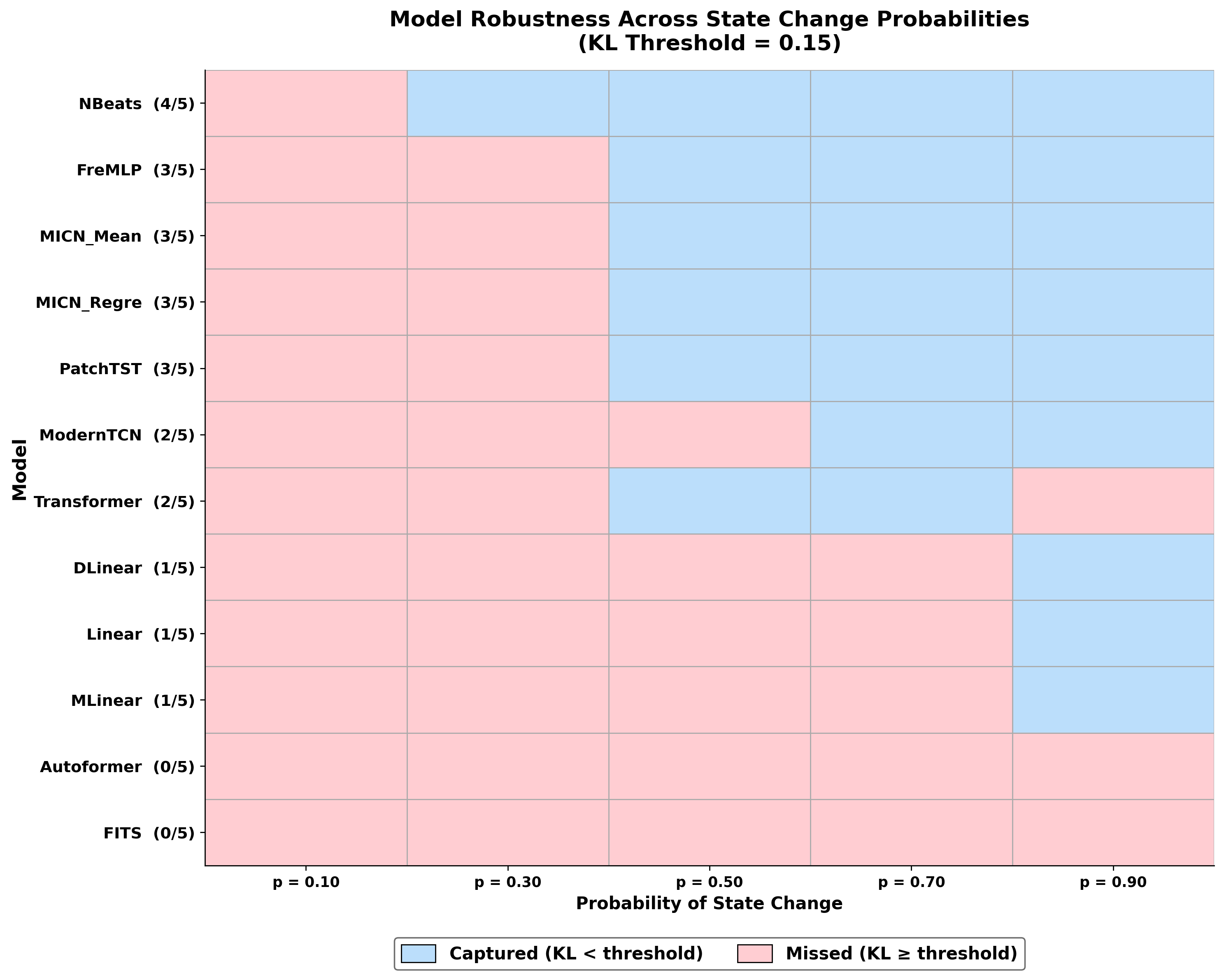}%
\hfill
\includegraphics[width=0.48\textwidth]{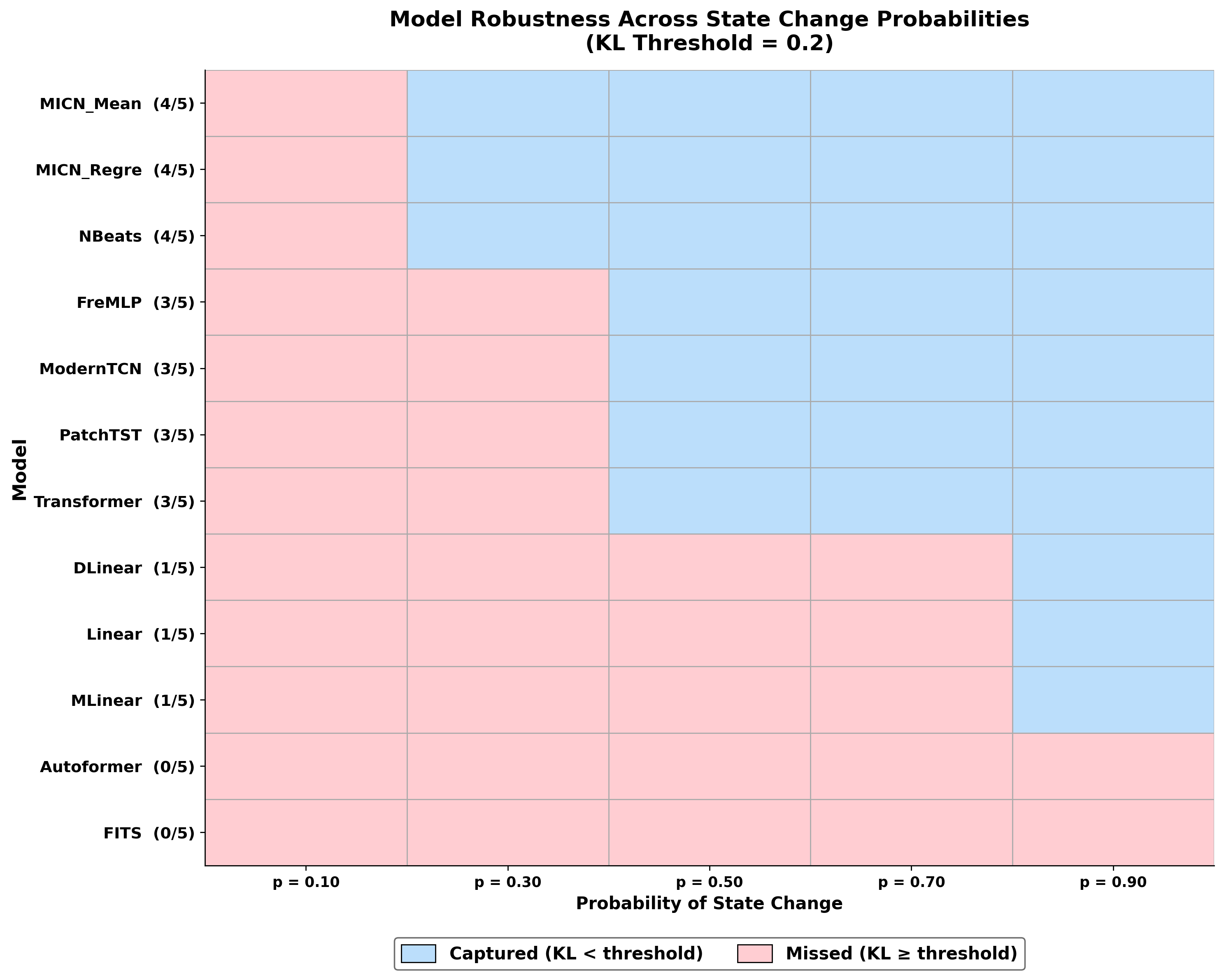}
\caption{\textbf{Pass/fail classification is robust to KL threshold
choice.} Each panel shows the model $\times$ transition-probability pass/fail matrix at a different symmetric KL threshold. Blue: KL below threshold (captured); red: KL at or above threshold (missed). Models are
sorted by pass rate (descending). At $\mathrm{KL} < 0.05$ (main-text threshold), PatchTST leads with 2/5. At $\mathrm{KL} < 0.10$, FreMLP, NBeats, and PatchTST each reach 3/5. At $\mathrm{KL} < 0.20$, PatchTST and NBeats reach 4/5. Autoformer remains at 0/5 across all thresholds (KL range: 0.521 to 2.020). The rank ordering is stable across all fourthresholds, confirming that the main-text conclusions are not an artifact
of a single threshold choice.}
\label{fig:kl_threshold_sensitivity}
\end{figure}

The main text reports that stochastic switching was the most challenging paradigm across all 11 architectures, with PatchTST succeeding at 2 of 5 transition probabilities and five architectures failing entirely. Here we present the complete KL divergence table across all models and transition probabilities (Table~\ref{tab:kl_full}), and a sensitivity analysis confirming that the architectural ranking is robust to threshold choice (Fig.~\ref{fig:kl_threshold_sensitivity}).

\paragraph{Full KL divergence results.}
Table~\ref{tab:kl_full} reports the symmetric KL divergence between true-history and predicted-future state-emission distributions for all 11 architectures across five transition probabilities. No model achieved
$\mathrm{KL} < 0.05$ at low transition probabilities ($p = 0.10$ or $p = 0.30$), where states persist for long durations and switching events are rare within any given window. All passes occurred at $p \geq 0.50$,
where more frequent switching provides sufficient within-window evidence of alternation.

\begin{table}[ht]
\centering
\caption{Symmetric KL divergence across all models and transition probabilities. Lower values indicate closer distributional match. Bold values fall below the 0.05 threshold used for pass/fail classification in
the main text. Models are ordered by pass count (descending), then by mean KL.}
\label{tab:kl_full}
\begin{tabular}{lccccc}
\toprule
\textbf{Model} & $p = 0.10$ & $p = 0.30$ & $p = 0.50$ & $p = 0.70$ & $p = 0.90$ \\
\midrule
PatchTST     & 0.209 & 0.234 & 0.063          & \textbf{0.008} & \textbf{0.046} \\
ModernTCN    & 0.573 & 0.584 & 0.187          & \textbf{0.014} & 0.062 \\
MICN\_Mean   & 0.742 & 0.186 & \textbf{0.016} & 0.108          & 0.114 \\
MICN\_Regre  & 0.743 & 0.175 & \textbf{0.026} & 0.096          & 0.114 \\
FreMLP       & 0.399 & 1.154 & \textbf{0.028} & 0.061          & 0.073 \\
DLinear      & 0.618 & 1.061 & 0.261          & 0.278          & \textbf{0.022} \\
Linear       & 0.911 & 1.094 & 0.342          & 0.309          & \textbf{0.034} \\
NBeats       & 0.324 & 0.061 & 0.121          & 0.085          & 0.066 \\
Transformer  & 0.795 & 1.242 & 0.137          & 0.102          & 0.160 \\
FITS         & 0.574 & 0.291 & 0.276          & 0.375          & 0.202 \\
MLinear      & 0.319 & 1.362 & 1.180          & 1.347          & 0.094 \\
Autoformer   & 1.007 & 0.521 & 1.371          & 1.460          & 2.020 \\
\bottomrule
\end{tabular}
\end{table}

Several patterns in the full table merit attention. PatchTST achieved the lowest KL values at $p = 0.70$ (0.008) and $p = 0.90$ (0.046), with both successes occurring at higher transition probabilities where the model receives more within-window evidence of alternation. At lower probabilities ($p = 0.10$ and $p = 0.30$), where states persist for longer durations and switches are rare, PatchTST failed (KL $= 0.209$ and $0.234$), indicating that even the best-performing architecture requires sufficient switching events within its temporal processing window to encode the transition structure. ModernTCN passed only at $p = 0.70$ (KL $= 0.014$) but narrowly missed at $p = 0.90$ (KL $= 0.062$), suggesting that its convolutional temporal processing window is tuned to a narrow range of switching rates.
MICN\_Mean and MICN\_Regre both passed at $p = 0.50$ (KL $= 0.016$ and $0.026$) but failed elsewhere, and FreMLP passed at $p = 0.50$ (KL $= 0.028$) but showed highly inconsistent recovery across neighboring probabilities ($p = 0.30$: KL $= 1.154$; $p = 0.70$: KL $= 0.061$), revealing erratic rather than gradual degradation. DLinear and Linear passed only at $p = 0.90$ (KL $= 0.022$ and $0.034$), the highest switching
rate, where the signal approaches rapid alternation that can be captured as an averaged pattern.

NBeats is a notable case: despite strong performance on the deterministic state-transition paradigm (recovering phase within 15 timesteps in the main text), it showed no capacity to recover stochastic switching (KL
range: $0.061$ to $0.324$), with its closest approach at $p = 0.30$ (KL $= 0.061$) narrowly missing the threshold. This dissociation highlights that fast deterministic adaptation and probabilistic state recovery  are distinct capabilities. Autoformer showed the worst overall performance, with KL exceeding 0.5 at every probability level (range: $0.521$ to $2.020$), indicating systematic divergence from the true
switching distribution. MLinear failed broadly (KL range: $0.094$ to $1.362$), and Transformer showed moderate but consistently above-threshold divergence (KL range: $0.102$ to $1.242$).

\paragraph{Sensitivity to KL threshold.}
The pass/fail classification in the main text uses a symmetric KL threshold of 0.05. To assess whether the architectural ranking depends on this specific choice, we evaluated all models at four thresholds: 0.05,
0.10, 0.15, and 0.20 (Fig.~\ref{fig:kl_threshold_sensitivity}).

At the strictest threshold ($\mathrm{KL} < 0.05$), PatchTST passed at 2 of 5 probability levels, six models at 1/5, and five models at 0/5. Relaxing to $\mathrm{KL} < 0.10$ promoted FreMLP, NBeats, and PatchTST to 3/5, with NBeats' near-miss at $p = 0.30$ (KL $= 0.061$) now classified as a pass. Autoformer, FITS, and Transformer remained at 0/5. At $\mathrm{KL} < 0.15$, PatchTST reached 3/5 and ModernTCN reached 2/5. At the most lenient threshold ($\mathrm{KL} < 0.20$), PatchTST and NBeats reached 4/5, but Autoformer remained at 0/5 (KL range: 0.521 to 2.020). The rank ordering was preserved across all four thresholds: PatchTST consistently occupied the top position, the 1/5 models formed a stable middle tier, and Autoformer, FITS, and Transformer consistently occupied the bottom tier. This stability confirms that the main-text finding, that stochasticswitching dynamics remain largely unrecoverable by current forecasting architectures, is robust to the specific threshold used for classification.

\subsection{Multi-paradigm performance profiles}
\label{app:pareto-results}

\begin{figure}[ht]
\centering
\includegraphics[width=0.85\textwidth]{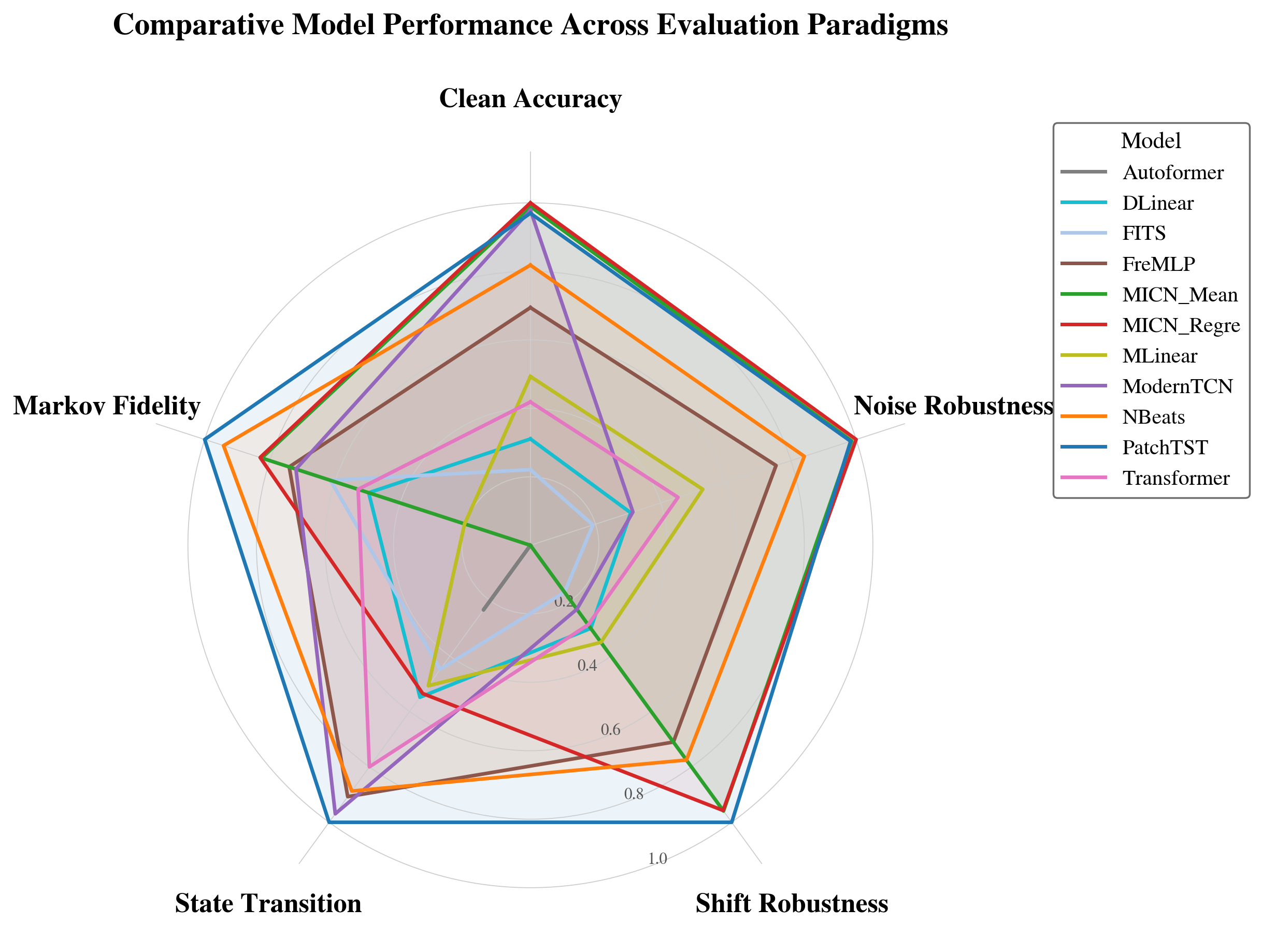}
\caption{\textbf{Comparative model performance across all five evaluation
paradigms.} Radar chart overlaying all 11 models. PatchTST (blue) is the most balanced model and leads on three of five axes.NBeats (orange) shows consistently high performance but is narrowly dominated by PatchTST. MICN variants (green) achieve the highest clean accuracy and noise robustness but collapse on state-transition adaptation. Linear-family models (cyan/light blue) and Autoformer (gray) occupy the interior, indicating weak performance across most paradigms. Scores are min-max normalized across models per paradigm (0 to 1 scale).}
\label{fig:supp_radar_all}
\end{figure}

The main text summarizes multi-paradigm performance via the Pareto frontier (main-text Fig.~\ref{fig:pareto}). To provide a more detailed view of each architecture's strengths and weaknesses, Fig.~\ref{fig:supp_radar_all} overlays all 11 models on a single radar chart, and Table~\ref{tab:pareto_scores} reports the full normalized scores and Pareto classifications.

Each axis represents one of the five evaluation paradigms (clean accuracy, noise robustness, shift robustness, state-transition adaptation, and Markov fidelity). Scores are min-max normalized across models (0 to 1
scale), where higher values indicate better performance relative to the other architectures.

\begin{table}[ht]
\centering
\caption{Normalized paradigm scores and Pareto classification. Scores are
min-max normalized (0 to 1) across models per paradigm. Frontier models
cannot be improved on any dimension without sacrificing another. PatchTST
achieves the highest mean score (0.99) and is the only model above 0.95
on all five dimensions. MICN variants score highest on clean and noise but
are weakened by state-transition performance. Autoformer scores 0.00 on
three of five dimensions.}
\label{tab:pareto_scores}
\begin{tabular}{lccccccl}
\toprule
\textbf{Model} & \textbf{Clean} & \textbf{Noise} & \textbf{Shift} & \textbf{State Tr.} & \textbf{Markov} & \textbf{Mean} & \textbf{Status} \\
\midrule
PatchTST    & 0.97 & 0.98 & 1.00 & 1.00 & 1.00 & 0.99 & Frontier  \\
MICN\_Regre & 1.00 & 1.00 & 0.96 & 0.54 & 0.83 & 0.86 & Frontier  \\
NBeats      & 0.82 & 0.84 & 0.78 & 0.89 & 0.94 & 0.85 & Dominated \\
MICN\_Mean  & 0.99 & 0.99 & 0.96 & 0.00 & 0.82 & 0.75 & Frontier  \\
FreMLP      & 0.69 & 0.75 & 0.71 & 0.91 & 0.74 & 0.76 & Dominated \\
ModernTCN   & 0.98 & 0.31 & 0.23 & 0.97 & 0.72 & 0.64 & Frontier  \\
Transformer & 0.42 & 0.45 & 0.29 & 0.80 & 0.53 & 0.50 & Dominated \\
MLinear     & 0.49 & 0.53 & 0.35 & 0.51 & 0.20 & 0.42 & Dominated \\
DLinear     & 0.31 & 0.31 & 0.30 & 0.55 & 0.50 & 0.39 & Dominated \\
FITS        & 0.22 & 0.19 & 0.17 & 0.45 & 0.61 & 0.33 & Dominated \\
Autoformer  & 0.00 & 0.00 & 0.00 & 0.23 & 0.00 & 0.05 & Dominated \\
\bottomrule
\end{tabular}
\end{table}

The normalized scores reveal several patterns not immediately visible in the Pareto frontier visualization. NBeats, despite being classified as dominated, achieved the third-highest mean score (0.85) with no dimension below 0.78, making it a strong general-purpose alternative when balanced performance is prioritized over frontier optimality. The gap between NBeats and the frontier is narrow: it is dominated by PatchTST across all five dimensions but only marginally so on clean (0.82 vs 0.97) and noise (0.84 vs 0.98). MICN\_Mean's state-transition score of 0.00 reflects its consistent degradation below linear baseline on that paradigm, a vulnerability that would be masked by its excellent clean and noise scores (both 0.99) if only aggregate performance were reported. ModernTCN's frontier status depends entirely on its state-transition strength (0.97), as its noise (0.31) and shift (0.23) scores are among the lowest, illustrating how Pareto optimality can reflect narrow specialization rather than balanced capability. Autoformer scored 0.00 on clean accuracy, noise robustness, and Markov fidelity, confirming its unsuitability for any digital twin application evaluated in this study.

\end{document}